\pdfoutput=1
\documentclass[10pt,twocolumn,letterpaper]{article}

\usepackage{cvpr}
\usepackage{times}

\usepackage{preamble}

\newif\iflongversion
\longversiontrue

\iflongversion
\newcommand{\vsum}{\sum}
\newcommand{\vneghspace}{}
\newcommand{\vnegvspace}{}
\else
\newcommand{\vsum}{\tsum}
\newcommand{\vneghspace}{\hspace{-.5em}}
\newcommand{\vnegvspace}{\vspace{-.5em}}
\fi

\newcommand{\Topk}{Top-\texorpdfstring{$k$}{k}}
\newcommand{\Sm}{\Delta}

\newcommand{\Ska}{\Delta^{\alpha}_{k}}
\newcommand{\Skb}{\Delta^{\beta}_{k}}
\newcommand{\kerr}[1]{\err_k\!\left(#1\right)}
\newcommand{\nerr}[2]{\err_{#1}\!\left(#2\right)}

\newcommand{\LrOva}{${\rm LR}^{\rm OVA}$}
\newcommand{\LrMulti}{${\rm LR}^{\rm Multi}$}
\newcommand{\LrTopK}[1]{${\rm top}\mhyphen {#1}~{\rm Ent}$}
\newcommand{\LrTopKn}[1]{${\rm top}\mhyphen {#1}~{\rm Ent_{tr}}$}
\newcommand{\SvmOva}{${\rm SVM}^{\rm OVA}$}

\newcommand{\SvmMulti}{${\rm SVM}^{\rm Multi}$}
\newcommand{\SvmTopK}[1]{${\rm top}\mhyphen {#1}~{\rm SVM}$}
\newcommand{\SvmTopKg}[2]{${\rm top}\mhyphen {#1}~{\rm SVM_{#2}}$}
\newcommand{\SvmTopKa}[1]{${\rm top}\mhyphen {#1}~{\rm SVM^\alpha}$}
\newcommand{\SvmTopKb}[1]{${\rm top}\mhyphen {#1}~{\rm SVM^\beta}$}
\newcommand{\SvmTopKag}[2]{${\rm top}\mhyphen {#1}~{\rm SVM^\alpha_{#2}}$}
\newcommand{\SvmTopKbg}[2]{${\rm top}\mhyphen {#1}~{\rm SVM^\beta_{#2}}$}

\def\R{\Rb}

\allowdisplaybreaks

\usepackage[pagebackref=true,breaklinks=true,colorlinks,bookmarks=false]{hyperref}

\cvprfinalcopy

\def\cvprPaperID{885}

\ifcvprfinal\fi
\begin{document}

\title{Loss Functions for Top-k Error: Analysis and Insights}

\author{
Maksim Lapin,$^1$
Matthias Hein$^2$
and
Bernt Schiele$^1$ \\
$^1$Max Planck Institute for Informatics, Saarbr\"ucken, Germany \\
$^2$Saarland University, Saarbr\"ucken, Germany
}

\maketitle

\begin{abstract}
In order to push the performance on realistic computer vision tasks,
the number of classes in modern benchmark datasets has significantly
increased in recent years.
This increase in the number of classes comes along
with increased ambiguity between the class labels,
raising the question if top-1 error is the right performance measure.
In this paper, we provide an extensive comparison and evaluation of 
established multiclass methods comparing their top-$k$ performance
both from a practical as well as from a theoretical perspective.
Moreover, we introduce novel top-$k$ loss functions
as modifications of the softmax and the multiclass SVM losses
and provide efficient optimization schemes for them.
In the experiments,
we compare on various datasets
all of the proposed and established methods for top-$k$ error optimization.
An interesting insight of this paper is that the softmax loss
yields competitive top-$k$ performance for all $k$ simultaneously.
For a specific top-$k$ error,
our new top-$k$ losses lead typically to further improvements
while being faster to train than the softmax.
\end{abstract}

\section{Introduction}
\label{sec:introduction}
The number of classes
is rapidly growing in modern computer vision benchmarks
\cite{ILSVRCarxiv14,zhou2014learning}.
Typically, this also leads to 
ambiguity in the labels as classes start to overlap.
Even for humans, the error rates in top-$1$ performance are often quite high
($\approx 30\%$ on SUN 397 \cite{xiao2010sun}).
While previous research focuses on minimizing the top-$1$ error,
we address top-$k$ error optimization in this paper.
We are interested in two cases:
a) achieving small top-$k$ error for \emph{all} reasonably small $k$;
and
b) minimization of a specific top-$k$ error.

While it is argued in \cite{akata2014good} that the one-versus-all (OVA)
SVM scheme performs on par in top-$1$ and top-$5$ accuracy
with the other SVM variations based on ranking losses,
we have recently shown in \cite{lapin2015topk}
that minimization of the \emph{top-$k$ hinge loss}
leads to improvements in top-$k$ performance compared to OVA SVM,
multiclass SVM, and other ranking-based formulations.
In this paper,
we study top-$k$ error optimization from a wider perspective.
On the one hand,
we compare OVA schemes and direct multiclass losses
in extensive experiments,
and on the other,
we present theoretical discussion
regarding their calibration for the top-$k$ error.
Based on these insights,
we suggest $4$ new families of loss functions for the top-$k$ error.
Two are smoothed versions of the top-$k$ hinge losses
\cite{lapin2015topk},
and the other two are top-$k$ versions
of the softmax loss.
We discuss their advantages and disadvantages,
and for the convex losses provide an efficient implementation
based on stochastic dual coordinate ascent (SDCA)
\cite{shalev2014accelerated}.

We evaluate a battery of loss functions on
$11$ datasets of different tasks ranging from text classification
to large scale vision benchmarks,
including fine-grained and scene classification.
We systematically optimize and report
results separately for each top-$k$ accuracy.
One interesting message that we would like to highlight
is that the softmax loss
is able to optimize \emph{all top-$k$ error measures simultaneously}.
This is in contrast to multiclass SVM
and is also reflected in our experiments.
Finally, we show that our new top-$k$ variants of
smooth multiclass SVM and the softmax loss 
can further improve top-$k$ performance for a specific $k$.

\textbf{Related work.}
Top-$k$ optimization has recently received revived attention
with the advent of large scale problems
\cite{Gupta2014,lapin2015topk,li2014top,liu2015transductive}.
The top-$k$ error in multiclass classification,
which promotes good ranking of class \emph{labels} for each example,
is closely related to the
precision@$k$ metric in information retrieval,
which counts the fraction of positive instances among the top-$k$
ranked \emph{examples}.
In essence, both approaches enforce a desirable
\emph{ranking} of items \cite{lapin2015topk}.

The classic approaches optimize pairwise ranking with
${\rm SVM^{struct}}$ \cite{joachims2005support,tsochantaridis2005large},
${\rm RankNet}$ \cite{burges2005learning},
and ${\rm LaRank}$ \cite{bordes2007solving}.
An alternative direction was proposed by
Usunier \etal \cite{usunier2009ranking}, who described a general family
of convex loss functions for ranking and classification.
One of the loss functions that we consider
(\SvmTopKb{k} \cite{lapin2015topk}) 
also falls into that family.
Weston \etal \cite{Weston2011} then introduced Wsabie,
which optimizes an approximation of a ranking-based loss from
\cite{usunier2009ranking}.
A Bayesian approach was suggested by \cite{swersky2012probabilistic}.

Recent works focus on the top of the ranked list
\cite{agarwal2011infinite,boyd2012accuracy,ICML2012Rakotomamonjy_664,rudin2009p},
scalability to large datasets
\cite{Gupta2014,lapin2015topk,li2014top},
explore transductive learning \cite{liu2015transductive}
and prediction of tuples \cite{ross2013learning}.

\bgroup
\def\arraystretch{1.3}
\begin{table*}[ht]\small\centering\setlength{\tabcolsep}{.45em}
\begin{tabular}{l|l|c|c|c|c|c}
\toprule
Method & Name & Loss function & Conjugate & SDCA update & Top-$k$ calibrated & Convex \\
\midrule
\midrule
\SvmOva & One-vs-all (OVA) SVM &
$\max\{0, \, 1 - a \}$ &
\multirow{2}{*}{\cite{shalev2014accelerated}} &
\multirow{2}{*}{\cite{shalev2014accelerated}} &
no$^1$ (Prop.~\ref{prop:calibrated-hinge}) &
\multirow{8}{*}{yes}
\\
\LrOva & OVA logistic regression &
$\log(1 + e^{- a} )$ &
&
&
yes (Prop.~\ref{prop:calibrated-lr}) &
\\
\cline{1-6}
\SvmMulti\ & Multiclass SVM &
$\max \big\{0, (a + c)_{\pi_1} \big\}$ &
\cite{lapin2015topk,shalev2014accelerated} &
\cite{lapin2015topk,shalev2014accelerated} &
no (Prop.~\ref{prop:calibrated-multi-hinge}) &
\\
\LrMulti\ & Softmax (maximum entropy) &
$\log\big( \sum_{j \in \Yc} \exp(a_j) \big)$ &
Prop.~\ref{prop:softmax-conj} &
Prop.~\ref{prop:topk-entropy-update} &
yes (Prop.~\ref{prop:calibrated-softmax}) &
\\
\cline{1-6}
\SvmTopKa{k} & Top-$k$ hinge ($\alpha$) &
$\max \big\{0, \frac{1}{k} \sum_{j=1}^k (a + c)_{\pi_j} \big\}$ &
\multirow{2}{*}{\cite{lapin2015topk}} &
\multirow{2}{*}{\cite{lapin2015topk}} &
\multirow{5}{*}{\multirowcell{open\\question\\for $k>1$}} &
\\
\SvmTopKb{k} & Top-$k$ hinge ($\beta$) &
$\frac{1}{k} \sum_{j=1}^k \max \big\{0, (a + c)_{\pi_j} \big\}$ &
&
&
\\
\cline{1-5}
\SvmTopKag{k}{\gamma} & Smooth top-$k$ hinge ($\alpha$) $^*$ &
Eq.~\eqref{eq:smooth-topk-alpha} w/ $\Ska$ &
\multirow{2}{*}{Prop.~\ref{prop:topk-smooth}} &
\multirow{2}{*}{Prop.~\ref{prop:topk-smooth-update}} &
&
\\
\SvmTopKbg{k}{\gamma} & Smooth top-$k$ hinge ($\beta$) $^*$ &
Eq.~\eqref{eq:smooth-topk-alpha} w/ $\Skb$ &
&
&
&
\\
\cline{1-5}
\LrTopK{k} & Top-$k$ entropy $^*$ &
Prop.~\ref{prop:topk-entropy-primal} &
Eq.~\eqref{eq:softmax-conj} &
Prop.~\ref{prop:topk-entropy-update} &
&
\\
\cline{6-7}
\LrTopKn{k} & Truncated top-$k$ entropy $^*$ &
Eq.~\eqref{eq:truncated-topk-entropy} &
- &
- &
yes (Prop.~\ref{prop:calibrated-truncated-topk}) &
no
\\
\midrule
\multicolumn{7}{c}{\multirowcell{
Note that
\SvmMulti\ $\equiv$ \SvmTopKa{1} $\equiv$ \SvmTopKb{1}
and
\LrMulti\ $\equiv$ \LrTopK{1} $\equiv$ \LrTopKn{1}.\\
We let
$a \bydef y f(x)$ (binary one-vs-all);
$a \bydef (f_j(x) - f_y(x))_{j \in \Yc}$,
$c \bydef \ones - e_y$ (multiclass);
$\pi:$ $a_{\pi_1} \geq \ldots \geq a_{\pi_m}$.
}}
\\\bottomrule
\end{tabular}
\caption{Overview of the methods we consider and our contributions.
$^*$Novel loss.
$^1$But \emph{smoothed} one is (Prop.~\ref{prop:calibrated-smooth-hinge}).}
\label{tbl:summary}
\end{table*}
\egroup

\textbf{Contributions.}
We study the problem of top-$k$ error optimization
on a diverse range of learning tasks.
We consider existing methods
as well as propose 4 novel loss functions for minimizing the top-$k$ error.
A brief overview of the methods is given in Table~\ref{tbl:summary}.
For the proposed convex top-$k$ losses,
we develop an efficient optimization scheme based on SDCA\footnote{
Code available at: \url{https://github.com/mlapin/libsdca}
},
which can also be used for training with the softmax loss.
All methods are evaluated empirically in terms of the top-$k$ error
and, whenever possible, in terms of classification calibration.
We discover that the softmax loss
and the proposed smooth top-$1$ SVM are astonishingly
competitive in all top-$k$ errors.
Further small improvements can be obtained with the new top-$k$ losses.

\section{Loss Functions for \Topk\ Error}
\label{sec:multiclass}
We consider multiclass problems with $m$ classes where the training set
$(x_i, y_i)_{i=1}^n$ consists of $n$ examples $x_i \in \R^d$
along with the corresponding labels $y_i \in \Yc \bydef \{ 1, \ldots, m \}$.
We use $\pi$ and $\tau$ to denote a permutation of (indexes) $\Yc$.
Unless stated otherwise, $a_\pi$ reorders components of a vector $a \in \R^m$
in descending order, \ie
$ a_{\pi_1} \geq a_{\pi_2} \geq \ldots \geq a_{\pi_m}. $
While we consider linear classifiers in our experiments,
all loss functions below are formulated  in the general setting
where a function $f: \Xc \rightarrow \R^m$ is learned
and prediction at test time is done via $\argmax_{y \in \Yc} f_y(x)$,
resp.\ the top-$k$ predictions.
For the linear case, all predictors $f_y$ have
the form $f_y(x) = \inner{w_y, x}$.
Let $W \in \R^{d \times m}$ be the stacked weight matrix,
$L : \Yc \times \R^m \rightarrow \R$ be a convex loss function,
and $\lambda > 0$ be a regularization parameter.
We consider the following multiclass optimization problem
$
\min_W \frac{1}{n}\sum_{i=1}^n L(y_i, \tra{W} x_i) + \lambda \norm{W}_F^2 .
$

We use the Iverson bracket notation $\iv{P}$,
defined as $\iv{P} = 1$ if $P$ is true, $0$ otherwise;
and introduce a shorthand $p_y(x) \bydef \Pr(Y=y \given X=x)$.
We generalize the standard zero-one error
and allow $k$ guesses instead of one.
Formally, the \textbf{top-$k$ zero-one loss} (\textbf{top-$k$ error}) is
\begin{align}\label{eq:topk-error} 
\kerr{y, f(x)} \bydef \iv{f_{\pi_k}(x) > f_y(x)} .
\end{align}
Note that for $k=1$ we recover the standard zero-one error.
\textbf{Top-$k$ accuracy} is defined as $1$ minus the top-$k$ error.

\subsection{Bayes Optimality and \Topk\ Calibration}
\label{sec:topk_calibration}

In this section, we establish the best achievable top-$k$ error,
determine when a classifier achieves it,
and define a notion of top-$k$ calibration.

\begin{lemma}\label{lem:bayes_kerr}
The Bayes optimal top-$k$ error at $x$ is
\begin{align*}
\min_{g \in \Rb^m} \Exp_{Y \given X}[\kerr{Y,g} \given X=x]
= 1 - \tsum_{j=1}^k p_{\tau_j}(x),
\end{align*}
where
$ p_{\tau_1}(x) \geq p_{\tau_2}(x) \geq \ldots \geq p_{\tau_m}(x)$.
A classifier $f$ is \textbf{top-$k$ Bayes optimal at $x$} if and only if
\begin{align*}
\big\{ y \given f_y(x) \geq f_{\pi_k}(x) \big\} \subset
\big\{ y \given p_y(x) \geq p_{\tau_k}(x) \big\} ,
\end{align*}
where $ f_{\pi_1}(x) \geq f_{\pi_2}(x) \geq \ldots \geq f_{\pi_m}(x)$.
\end{lemma}
\iflongversion
\begin{proof}
Let $g \in \R^m$ and $\pi$ be a permutation such that
$g_{\pi_1} \geq g_{\pi_2} \geq \ldots \geq g_{\pi_m}$.
The expected top-$k$ error at $x$ is
\begin{align*}
&\Exp_{Y \given X}[\kerr{Y,g} \given X=x] 
= \tsum_{y \in \Yc} \iv{g_{\pi_k} > g_{y}} p_y(x) \\
&= \tsum_{y \in \Yc} \iv{g_{\pi_k} > g_{\pi_y}} p_{\pi_y}(x)
= \tsum_{j=k+1}^m p_{\pi_j}(x) \\
&= 1 - \tsum_{j=1}^k p_{\pi_j}(x) .
\end{align*}
The error is minimal when $\tsum_{j=1}^k p_{\pi_j}(x)$ is maximal,
which corresponds to taking the $k$ largest conditional probabilities
$\tsum_{j=1}^k p_{\tau_j}(x)$ and yields the Bayes optimal top-$k$ error at $x$.

Since the relative order within $\{ p_{\tau_j}(x) \}_{j=1}^k$ is irrelevant
for the top-$k$ error, any classifier $f(x)$, for which the sets
$\{\pi_1, \ldots, \pi_k\}$ and $\{\tau_1, \ldots, \tau_k\}$
coincide, is Bayes optimal.

Note that we assumed \Wlog that there is a clear cut
$p_{\tau_k}(x) > p_{\tau_{k+1}}(x)$
between the $k$ most likely classes and the rest.
In general, ties can be resolved arbitrarily as long as we can guarantee
that the $k$ largest components of $f(x)$ correspond to the classes (indexes)
that yield the maximal sum $\tsum_{j=1}^k p_{\pi_j}(x)$ and lead to
top-$k$ Bayes optimality.
\end{proof}
\fi

Optimization of the zero-one loss (and, by extension, the top-$k$ error)
leads to hard combinatorial problems.
Instead, a standard approach is to use a convex surrogate loss
which upper bounds the zero-one error.
Under mild conditions on the loss function \cite{BarJorAuc2006,TewBar2007},
the optimal classifier \wrt the surrogate yields
a Bayes optimal solution for the zero-one loss.
Such loss is called \emph{classification calibrated},
which is known in statistical learning theory as a necessary condition
for a classifier to be universally Bayes consistent \cite{BarJorAuc2006}.
We introduce now the notion of calibration for the top-$k$ error.
\begin{definition}\label{def:calibration}
A loss function $L:\Yc \times \R^m \rightarrow \R$
(or a reduction scheme)
is called \textbf{top-$k$ calibrated} if 
for all possible data generating measures on $\R^d \times \Yc$
and all $x \in \R^d$
\begin{align*}
&\targmin_{g \in \R^m} \Exp_{Y \given X}[L(Y,g) \given X = x]  \\
\subseteq &\targmin_{g \in \R^m} \Exp_{Y \given X}[\kerr{Y,g} \given X = x] .
\end{align*}
\end{definition}
If a loss is \emph{not} top-$k$ calibrated,
it implies that even in the limit of infinite data,
one does not obtain a classifier with
the Bayes optimal top-$k$ error from Lemma~\ref{lem:bayes_kerr}.

\subsection{OVA and Direct Multiclass Approaches}
\label{sec:baselines}
The standard multiclass problem is often solved using the one-vs-all (OVA)
reduction into a set of $m$ binary classification problems.
Every class is trained versus the rest
which yields $m$ classifiers $\{f_y\}_{y \in \Yc}$.

Typically, the binary classification problems are formulated
with a convex margin-based loss function $L(yf(x))$, where $L:\R \rightarrow \R$
and $y = \pm 1$.
We consider in this paper:
\begin{align}
L(yf(x)) &= \max\{0, \, 1 - y f(x)\} , \label{eq:ova-hinge} \\
L(yf(x)) &= \log(1 + e^{- y f(x)} ) . \label{eq:ova-lr}
\end{align}
The \textbf{hinge} (\ref{eq:ova-hinge}) and
\textbf{logistic} (\ref{eq:ova-lr}) losses correspond to
the SVM and logistic regression respectively.
We now show when the OVA schemes are top-$k$ calibrated,
not only for $k=1$ (standard multiclass loss) but \emph{for all} $k$ simultaneously.

\begin{lemma}\label{lem:calibration-OVA}
The OVA reduction is top-$k$ calibrated for any $1\leq k\leq m$
if the Bayes optimal function of the
convex margin-based loss $L(yf(x))$ is a
strictly monotonically increasing function of $\Pr(Y=1 \given X=x)$.
\end{lemma}
\iflongversion
\begin{proof}
For every class $y \in \Yc$, the Bayes optimal classifier
for the corresponding binary problem has the form
\[
f_y(x) = g\big(\Pr(Y=y \given X=x)\big) ,
\]
where $g$ is a strictly monotonically increasing function.
The ranking of $f_y$ corresponds to the 
ranking of $\Pr(Y=y \given X=x)$ and hence the OVA reduction
is top-$k$ calibrated for any $k=1,\ldots,m$.
\end{proof}
\fi

Next, we check if the one-vs-all schemes employing hinge
and logistic regression losses are top-$k$ calibrated.
\begin{proposition}\label{prop:calibrated-hinge}
OVA SVM is not top-$k$ calibrated.
\end{proposition}
\iflongversion
\begin{proof}
First, we show that the Bayes optimal function for the binary hinge loss is
\begin{align*}
f^*(x) &= 2 \iv{\Pr(Y=1 \given X=x) > \tfrac{1}{2}} -1 .
\end{align*}
We decompose the expected loss as
\[ \Exp_{X,Y}[L(Y,f(X))]=\Exp_{X}[\Exp_{Y|X}[L(Y,f(x)) \given X = x]].\]
Thus, one can compute the Bayes optimal classifier $f^*$
pointwise by solving
\[ \argmin_{\alpha \in \R} \Exp_{Y|X}[L(Y,\alpha) \given X=x],\]
for every $x \in \R^d$, which leads to the following problem
\begin{align*}
\argmin_{\alpha \in \R} \;
\max\{0, 1-\alpha\} p_1(x) + \max\{0, 1+\alpha\} p_{-1}(x) ,
\end{align*}
where $p_y(x) \bydef \Pr(Y=y \given X=x)$.
It is obvious that the optimal $\alpha^*$ is contained in $[-1,1]$.
We get
\[
\argmin_{-1\leq \alpha\leq 1} \;
(1 - \alpha) p_1(x) + (1 + \alpha)p_{-1}(x).
\]
The minimum is attained at the boundary and we get
\[
f^*(x) =
\begin{cases} +1 & \text{if } p_1(x) > \frac{1}{2} , \\
-1 & \text{if } p_1(x) \leq \frac{1}{2} .
\end{cases}
\]
Therefore, the Bayes optimal classifier for the hinge loss is not
a strictly monotonically increasing function of $p_1(x)$.

To show that OVA hinge is not top-$k$ calibrated, we construct an example problem
with $3$ classes and $p_1(x) = 0.4$, $p_2(x) = p_3(x) = 0.3$.
Note that for every class $y = 1, 2, 3$, the Bayes optimal binary classifier is $-1$,
hence the predicted ranking of labels is arbitrary and
may not produce the Bayes optimal top-$k$ error.
\end{proof}
\fi

\noindent
In contrast, logistic regression is top-$k$ calibrated.

\begin{proposition}\label{prop:calibrated-lr}
OVA logistic regression is top-$k$ calibrated.
\end{proposition}
\iflongversion
\begin{proof}
First, we show that the Bayes optimal function for the binary logistic loss is
\begin{align*}
f^*(x) &= \log\Big(\frac{p_x(1)}{1-p_x(1)}\Big) .
\end{align*}
As above, the pointwise optimization problem is
\[ \argmin_{\alpha \in \R} \;
\log(1+\exp(-\alpha))p_1(x) + \log(1+\exp(\alpha))p_{-1}(x).\]
The logistic loss is known to be convex and differentiable and thus the optimum can be computed via
\[
\frac{-\exp(-\alpha)}{1+\exp(-\alpha)}p_1(x)
+ \frac{\exp(\alpha)}{1+\exp(\alpha)}p_{-1}(x)=0 .
\]
Re-writing the first fraction we get
\[
\frac{-1}{1+\exp(\alpha)}p_1(x)
+ \frac{\exp(\alpha)}{1+\exp(\alpha)}p_{-1}(x)=0 ,
\]
which can be solved as
$ \alpha^* = \log\Big(\frac{p_1(x)}{p_{-1}(x)}\Big)$
and leads to the formula for
the Bayes optimal classifier stated above.

We check now that the function $\phi: (0,1) \rightarrow \Rb$
defined as $\phi(x)=\log(\frac{x}{1-x})$ is strictly monotonically increasing.
\begin{align*}
\phi'(x) &= \frac{1-x}{x}\big( \frac{1}{1-x} + \frac{x}{(1-x)^2}\big) \\
&= \frac{1-x}{x}\frac{1}{(1-x)^2} = \frac{1}{x(1-x)} > 0, \quad \forall x \in (0,1).
\end{align*}
The derivative is strictly positive on $(0,1)$, which implies that
$\phi$ is strictly monotonically increasing.
The logistic loss, therefore, fulfills the conditions of
Lemma~\ref{lem:calibration-OVA} and is top-$k$ calibrated for
any $1 \leq k \leq m$.
\end{proof}
\fi

An alternative to the OVA scheme with binary losses is to use a \emph{multiclass} loss
$L:\Yc \times \R^m \rightarrow \R$ directly. 
We consider two generalizations of the hinge and logistic losses below:
\begin{align}
&L(y,f(x)) = \max_{j \in \Yc} \big\{ \iv{j\neq y} + f_{j}(x) - f_y(x) \big\} ,
\label{eq:multi-hinge} \\
&L(y,f(x)) = \log\Big( \tsum_{j \in \Yc} \exp(f_j(x) - f_y(x)) \Big) .
\label{eq:softmax}
\end{align}
Both the \textbf{multiclass hinge loss} (\ref{eq:multi-hinge}) of
Crammer \& Singer \cite{crammer2001algorithmic}
and the \textbf{softmax loss} (\ref{eq:softmax}) %
are popular losses for multiclass problems.
The latter is also known as the cross-entropy or multiclass logistic loss
and is often used as the last layer in deep architectures
\cite{bengio2009learning, krizhevsky2012imagenet, simonyan14c}.
The multiclass hinge loss has been shown to be competitive
in large-scale image classification \cite{akata2014good},
however, it is known to be not calibrated \cite{TewBar2007}
for the top-$1$ error.
Next, we show that it is not top-$k$ calibrated for any $k$.

\begin{proposition}\label{prop:calibrated-multi-hinge}
Multiclass SVM is not top-$k$ calibrated.
\end{proposition}
\iflongversion
\begin{proof}
First, we derive the Bayes optimal function.

Let $y \in \argmax_{j \in \Yc} p_j(x)$.
Given any $c \in \R$, a Bayes optimal function $f^*:\R^d \rightarrow \R^m$
for the loss \eqref{eq:multi-hinge} is
\begin{align*}
f^*_{y}(x) &= \begin{cases}
c + 1 & \text{if } \max_{j \in \Yc} p_j(x) \geq \frac{1}{2} , \\
c & \text{otherwise} ,
\end{cases} \\
f^*_{j}(x) &= c, \; j \in \Yc \setminus \{y\} .
\end{align*}
Let $g = f(x) \in \R^m$, then
\begin{align*}
\Exp_{Y|X}[L(Y,g) \given X] =
\sum_{l \in \Yc} \max_{j \in \Yc} \big\{\iv{j \neq l} + g_j - g_l \big\} p_l(x) .
\end{align*}
Suppose that the maximum of $(g_j)_{j \in \Yc}$ is not unique.
In this case, we have
\[
\max_{j \in \Yc} \big\{\iv{j\neq l} + g_j - g_l \big\} \geq 1 , \; \forall \, l \in \Yc
\]
as the term $\iv{j\neq l}$ is always active.
The best possible loss is obtained by setting $g_j = c$ for all $j \in \Yc$,
which yields an expected loss of $1$.
On the other hand, if the maximum is unique and is achieved by $g_y$, then
\begin{multline*}
\max_{j \in \Yc} \big\{\iv{j\neq l} + g_j - g_l \big\} \\
= \begin{cases}
1 + g_{y} - g_l & \textrm{ if } l \neq y, \\
\max\big\{0,\, \max_{j \neq y} \{ 1 + g_j - g_y \} \big\} & \textrm{ if } l=y.
\end{cases}
\end{multline*}
As the loss only depends on the gap $g_{y} - g_l$,
we can optimize this with $\beta_l = g_{y} - g_l$.
\begin{align*}
\Exp&_{Y|X}[L(Y,g) \given X=x] \\
&= \sum_{l \neq y} (1 + g_y - g_l) p_l(x) \\
&+ \max\big\{0,\, \max_{l \neq y} \{ 1 + g_l - g_y \} \big\} p_y(x) \\
&= \sum_{l \neq y} (1 + \beta_l) p_l(x)
+ \max\big\{0,\, \max_{l \neq y} \{ 1 - \beta_l \} \big\} p_y(x) \\
&= \sum_{l \neq y} (1 + \beta_l) p_l(x)
+ \max\{0, 1 - \min_{l \neq y} \beta_l \} p_y(x) .
\end{align*}
As only the minimal $\beta_l$ enters the last term,
the optimum is achieved if all $\beta_l$ are equal for $l \neq y$
(otherwise it is possible to reduce the first term without affecting the last term).
Let $\alpha \bydef \beta_l$ for all $l\neq y$.
The problem becomes
\begin{align*}
&\min_{\alpha \geq 0} \sum_{l \neq y} (1+\alpha) p_l(x) + \max\{0, 1-\alpha \} p_y(x) \\
&\equiv \min_{0 \leq \alpha \leq 1} \alpha (1 - 2 p_y(x) )
\end{align*}
Let $p \bydef p_y(x) = \Pr(Y=y \given X=x)$.
The solution is
\begin{align*}
\alpha^* = \begin{cases}
0 & \textrm{if } p < \frac{1}{2} , \\
1 & \textrm{if } p \geq \frac{1}{2} ,
\end{cases}
\end{align*}
and the associated risk is
\begin{align*}
\Exp_{Y|X}[L(Y,g) \given X=x] =
\begin{cases}
1 & \text{if } p < \frac{1}{2}, \\
2 (1 - p) & \text{if } p \geq \frac{1}{2} .
\end{cases}
\end{align*}
If $p < \frac{1}{2}$, then the Bayes optimal classifier
$f^*_j(x) = c$ for all $j \in \Yc$ and any $c \in \Rb$.
Otherwise, $p \geq \frac{1}{2}$ and
$$
f^*_j(x) = \begin{cases}
c+1 & \text{if } j=y , \\
c & \text{if } j \in \Yc \setminus \{y\} .
\end{cases}
$$
Moreover, we have that the Bayes risk at $x$ is
$$
\Exp_{Y|X}[L(Y,f^*(x)) \given X=x] = \min\{1,2(1 - p)\} \leq 1 .
$$

It follows, that the multiclass hinge loss is not
(top-$1$) classification calibrated
at any $x$ where $\max_{y \in \Yc} p_y(x) < \frac{1}{2}$
as its Bayes optimal classifier reduces to a constant.
Moreover, even if $p_y(x) \geq \frac{1}{2}$ for some $y$,
the loss is not top-$k$ calibrated for $k \geq 2$
as the predicted order of the remaining classes need not be optimal.
\end{proof}
\fi

\noindent
Again, a contrast between the hinge and logistic losses.

\begin{proposition}\label{prop:calibrated-softmax}
The softmax loss is top-$k$ calibrated.
\end{proposition}
\iflongversion
\begin{proof}
The multiclass logistic loss is (top-$1$) calibrated for the zero-one error
in the following sense.
If
\[
f^*(x) \in \argmin_{ g \in \R^m} \Exp_{Y|X}[L(Y,g) \given X=x] ,
\]
then for some $\alpha>0$ and all $y \in \Yc$
\[
f^*_y(x) = \begin{cases}
\log(\alpha \, p_y(x) ) & \text{if } p_y(x) > 0, \\
-\infty & \text{otherwise},
\end{cases}
\]
which implies
\[
\argmax_{y \in \Yc} f^*_y(x) = \argmax_{y \in \Yc} \Pr(Y=y \given X=x) .
\]
We now prove this result and show that it also generalizes
to top-$k$ calibration for $k>1$.
Using the identity
$$
L(y,g) = \log\big(\tsum_{j \in \Yc} e^{g_j - g_y}\big)
= \log\big(\tsum_{j \in \Yc} e^{g_j}\big) - g_y
$$
and the fact that $\sum_{y \in \Yc} p_y(x) = 1$,
we write for a $g \in \R^m$
\begin{align*}
\Exp&_{Y|X}[L(Y,g) \given X=x] \\
&= \sum_{y \in \Yc} L(y,g) p_y(x)
= \log\big( \sum_{y \in \Yc} e^{g_y} \big) - \sum_{y \in \Yc} g_y p_x(y) .
\end{align*}
As the loss is convex and differentiable,
we get the global optimum by computing a critical point.
We have
\begin{align*}
\frac{\partial}{\partial g_j}\Exp_{Y|X}[L(Y,g) \given X=x] =
\frac{e^{g_j}}{\sum_{y \in \Yc} e^{g_y}} - p_j(x) = 0
\end{align*}
for $j \in \Yc$.
We note that the critical point is not unique as multiplication
$g\rightarrow \kappa g$ leaves the equation invariant for any $\kappa > 0$.
One can verify that $e^{g_j} = \alpha p_j(x)$
satisfies the equations for any $\alpha > 0$.
This yields a solution
\begin{align*}
f^*_y(x) &= \begin{cases}
\log(\alpha p_y(x) ) & \text{if } p_y(x) > 0 , \\
-\infty & \text{otherwise},
\end{cases}
\end{align*}
for any fixed $\alpha>0$.
We note that $f^*_y$ is a strictly monotonically increasing function
of the conditional class probabilities.
Therefore, it preserves the ranking of $p_y(x)$
and implies that $f^*$ is top-$k$ calibrated for any $1 \leq k \leq m$.
\end{proof}
\fi

The implicit reason for top-$k$ calibration of the OVA schemes
and the softmax loss is that one can estimate the probabilities $p_y(x)$
from the Bayes optimal classifier.
Loss functions which allow this are called \emph{proper}.
We refer to \cite{ReiWil2010} and references therein for a detailed discussion.

We have established that the OVA logistic regression and the softmax loss
are top-$k$ calibrated for any $k$, so why should we be interested
in defining new loss functions for the top-$k$ error?
The reason is that calibration is an asymptotic property
as the Bayes optimal functions are obtained pointwise.
The picture changes if we use linear classifiers, 
since they obviously cannot be minimized independently at each point.
Indeed, most of the Bayes optimal classifiers
cannot be realized by linear functions. 

In particular, convexity of the softmax and multiclass hinge losses
leads to phenomena where $\kerr{y, f(x)} = 0$, but $L(y, f(x)) \gg 0$.
This happens if
$f_{\pi_1}(x) \gg f_y(x) \geq f_{\pi_k}(x)$
and adds a bias when working with
``rigid'' function classes such as linear ones.
The loss functions which we introduce in the following
are modifications of the above losses
with the goal of alleviating that phenomenon.

\subsection{Smooth \Topk\ Hinge Loss}
\label{sec:topk-smooth}

Recently, we introduced two top-$k$ versions of
the multiclass hinge loss \eqref{eq:multi-hinge}
in \cite{lapin2015topk},
where the second version is based on
the family of ranking losses introduced earlier by \cite{usunier2009ranking}.
We use our notation from \cite{lapin2015topk} for direct comparison
and refer to the first version as $\alpha$ and the second one as $\beta$.
Let $c = \ones - e_y$, where $\ones$ is the all ones vector,
$e_y$ is the $y$-th basis vector,
and let $a \in \Rb^m$ be defined componentwise as
$a_j \bydef \inner{w_j, x} - \inner{w_y, x}$.
The two \textbf{top-$k$ hinge losses} are
\begin{align}
L(a) &=  \max \big\{0, \tfrac{1}{k} \tsum_{j=1}^k (a + c)_{\pi_j} \big\}
\; \big(\text{\SvmTopKa{k}}\big), \label{eq:topk-hinge-alpha} \\
L(a) &=  \tfrac{1}{k} \tsum_{j=1}^k \!\max \big\{0, (a + c)_{\pi_j} \big\}
\; \big(\text{\SvmTopKb{k}}\big), \label{eq:topk-hinge-beta}
\end{align}
where $(a)_{\pi_j}$ is the $j$-th largest component of $a$.
It was shown in \cite{lapin2015topk} that (\ref{eq:topk-hinge-alpha})
is a tighter upper bound on the top-$k$ error than (\ref{eq:topk-hinge-beta}),
however, both losses performed similarly in our experiments.
In the following, we simply refer to them as the top-$k$ hinge
or the top-$k$ SVM loss.

Both losses reduce to the multiclass hinge loss
\eqref{eq:multi-hinge} for $k=1$.
Therefore, they are unlikely to be top-$k$ calibrated,
even though we can currently neither
prove nor disprove this for $k>1$.
The multiclass hinge loss is not calibrated as
it is non-smooth and does not allow to estimate
the class conditional probabilities $p_y(x)$.
Our new family of \emph{smooth} top-$k$ hinge losses
is based on the Moreau-Yosida regularization
\cite{BecTeb2012, nesterov2005smooth}.
This technique has been used in \cite{shalev2014accelerated}
to smooth the binary hinge loss (\ref{eq:ova-hinge}).
Interestingly, smooth binary hinge loss fulfills
the conditions of Lemma~\ref{lem:calibration-OVA}
and leads to a top-$k$ calibrated OVA scheme.
The hope is that the smooth top-$k$ hinge loss
becomes top-$k$ calibrated as well.

Smoothing works by adding a quadratic term to the conjugate function\footnote{
The \textbf{convex conjugate} of $f$ is
$f^*(x^*) = \sup_x \{ \inner{x^*, x} - f(x) \}$.
},
which then becomes strongly convex.
Smoothness of the loss, among other things,
typically leads to much faster optimization as
we discuss in Section \ref{sec:optimization}.

\begin{proposition}\label{prop:calibrated-smooth-hinge}
OVA smooth hinge is top-$k$ calibrated.
\end{proposition}
\iflongversion
\begin{proof}
In order to derive the smooth hinge loss,
we first compute the \textbf{conjugate} of the standard \textbf{binary hinge loss},
\begin{align}
L(\alpha) &= \max\{0,1-\alpha\} , \nonumber\\
L^*(\beta) &=
\sup_{\alpha \in \R} \big\{ \alpha \beta - \max\{0,1-\alpha\} \big\} \nonumber\\
&= \begin{cases}
\beta & \textrm{if } -1 \leq \beta \leq 0, \\
\infty & \textrm{otherwise} .
\end{cases}
\label{eq:ova-hinge-conj}
\end{align}
The smoothed conjugate is
$$
L^*_\gamma(\beta) = L^*(\beta)+\frac{\gamma}{2}\beta^2 .
$$
The corresponding primal \textbf{smooth hinge loss} is given by
\begin{align}
L_\gamma(\alpha) &=
\sup_{-1 \leq \beta \leq 0}
\big\{ \alpha\beta - \beta-\tfrac{\gamma}{2}\beta^2 \big\} \nonumber\\
&= \begin{cases}
1 - \alpha - \frac{\gamma}{2} & \text{if } \alpha < 1 - \gamma, \\
\frac{(\alpha-1)^2}{2 \gamma} & \text{if } 1 - \gamma \leq \alpha \leq 1, \\
0, &  \text{if } \alpha > 1 .
\end{cases}
\label{eq:ova-hinge-smooth}
\end{align}
$L_\gamma(\alpha)$ is convex and differentiable with the derivative
\[
L'_\gamma(\alpha) =  \begin{cases}
-1 & \text{if } \alpha < 1-\gamma, \\
\frac{\alpha - 1}{\gamma} & \text{if } 1 - \gamma \leq \alpha \leq 1, \\
0, & \text{if } \alpha > 1.
\end{cases}
\]
We compute the Bayes optimal classifier pointwise.
\[
f^*(x) = \argmin_{\alpha \in \R} \; L(\alpha) p_1(x) + L(-\alpha) p_{-1}(x).
\]
Let $p \bydef p_1(x)$, the optimal $\alpha^*$ is found by solving
\[
L'(\alpha) p - L'(-\alpha) (1 - p) = 0 .
\]

\textbf{Case $0 < \gamma \leq 1$.}
Consider the case $1 - \gamma \leq \alpha \leq 1$,
\begin{align*}
\frac{\alpha-1}{\gamma} p + (1-p) = 0 \quad \Longrightarrow \quad
\alpha^* = 1 - \gamma \frac{1-p}{p}.
\end{align*}
This case corresponds to $p \geq \frac{1}{2}$,
which follows from the constraint $\alpha^* \geq 1 - \gamma$.
Next, consider $\gamma - 1 \leq \alpha \leq 1 - \gamma$,
\[ -p + (1-p) = 1-2p \neq 0 , \]
unless $p = \frac{1}{2}$, which is already captured by the first case.
Finally, consider $-1 \leq \alpha \leq \gamma - 1 \leq 1 - \gamma$.
Then
\[ -p - \frac{-\alpha-1}{\gamma}(1-p) = 0 \quad \Longrightarrow \quad
\alpha^* = -1 + \gamma\frac{p}{1-p},\]
where we have $-1 \leq \alpha^* \leq \gamma - 1$ if $p \leq \frac{1}{2}$.
We obtain the Bayes optimal classifier for $0 < \gamma \leq 1$ as follows:
\[
f^*(x) =
\begin{cases}
1 - \gamma \frac{1-p}{p} & \text{if } p \geq \frac{1}{2}, \\
-1 + \gamma \frac{p}{1-p} & \text{if } p < \frac{1}{2}.
\end{cases}
\]
Note that while $f^*(x)$ is not a continuous function of $p = p_1(x)$
for $\gamma < 1$,
it is still a strictly monotonically increasing function of $p$
for any $0 < \gamma \leq 1$.

\textbf{Case $\gamma>1$.}
First, consider $\gamma - 1 \leq \alpha \leq 1$,
\begin{align*}
\frac{\alpha - 1}{\gamma} p + (1-p) = 0 \quad \Longrightarrow \quad
\alpha^* = 1 - \gamma\frac{1-p}{p}.
\end{align*}
From $\alpha^* \geq \gamma - 1$, we get the condition $p \geq \frac{\gamma}{2}$.
Next, consider $1 - \gamma \leq \alpha \leq \gamma - 1$,
\[ \frac{\alpha-1}{\gamma}p - \frac{-\alpha-1}{\gamma} (1-p) =0
\quad \Longrightarrow \quad \alpha^* = 2p - 1, \]
which is in the range $[1 - \gamma, \gamma - 1]$
if $1 - \frac{\gamma}{2} \leq p \leq \frac{\gamma}{2}$.
Finally, consider $-1 \leq \alpha \leq 1 - \gamma$,
\[ -p - \frac{-\alpha-1}{\gamma} (1-p) = 0 \quad \Longrightarrow \quad
\alpha^* = -1 + \gamma\frac{p}{1-p}, \]
where we have $-1 \leq \alpha^* \leq 1 - \gamma$
if $p \leq 1 - \frac{\gamma}{2}$.
Overall, the Bayes optimal classifier for $\gamma>1$ is
\begin{align*}
f^*(x) &=
\begin{cases}
1 - \gamma \frac{1 - p}{p} &
\text{if } p \geq \frac{\gamma}{2}, \\
2 p - 1 &
\text{if } 1 - \frac{\gamma}{2} \leq p \leq \frac{\gamma}{2}, \\
-1 + \gamma \frac{p}{1 - p} &
\text{if } p <  1 - \frac{\gamma}{2}.
\end{cases}
\end{align*}
Note that $f^*$ is again a strictly monotonically increasing function
of $p = p_1(x)$.
Therefore, for any $\gamma > 0$,
the one-vs-all scheme with the smooth hinge loss (\ref{eq:ova-hinge-smooth})
is top-$k$ calibrated for all $1 \leq k \leq m$ by Lemma~\ref{lem:calibration-OVA}.
\end{proof}
\fi

Next, we introduce the \emph{multiclass} smooth top-$k$ hinge losses,
which extend the top-$k$ hinge losses (\ref{eq:topk-hinge-alpha})
and (\ref{eq:topk-hinge-beta}).
We define the \textbf{top-$k$ simplex} ($\alpha$ and $\beta$) of radius $r$ as
\begin{align}
\Ska(r) &\bydef
\big\{ x \given \inner{\ones, x} \leq r, \;
0 \leq x_i \leq \tfrac{1}{k} \inner{\ones, x}, \; \forall i \big\} ,
\label{eq:topk-simplex-alpha} \\
\Skb(r) &\bydef
\big\{ x \given \inner{\ones, x} \leq r, \;
0 \leq x_i \leq \tfrac{1}{k} r , \; \forall i \big\} .
\label{eq:topk-simplex-beta}
\end{align}
We also let $\Ska \bydef \Ska(1)$ and $\Skb \bydef \Skb(1)$.
\iflongversion
\begin{proposition}[\cite{lapin2015topk}]\label{prop:topk-conj}
The convex conjugates of (\ref{eq:topk-hinge-alpha}) and (\ref{eq:topk-hinge-beta})
are respectively
$L^*(b) = - \inner{c, b}$, if $b \in \Ska$, $+\infty$ otherwise;
and
$L^*(b) = - \inner{c, b}$, if $b \in \Skb$, $+\infty$ otherwise.
\end{proposition}
\fi

Smoothing applied to the top-$k$ hinge loss (\ref{eq:topk-hinge-alpha})
yields the following \textbf{smooth top-$k$ hinge loss} ($\alpha$).
Smoothing of (\ref{eq:topk-hinge-beta}) is done similarly,
but the set $\Ska(r)$ is replaced with $\Skb(r)$.

\begin{proposition}\label{prop:topk-smooth}
Let $\gamma > 0$ be the smoothing parameter.
The smooth top-$k$ hinge loss ($\alpha$) and its conjugate are
\begin{align}
L_{\gamma}(a) &= \tfrac{1}{\gamma} \big(
\inner{a + c, p} - \tfrac{1}{2} \inner{p, p} \big) ,
\label{eq:smooth-topk-alpha} \\
L_{\gamma}^*(b) &= \tfrac{\gamma}{2} \inner{b, b} - \inner{c, b} ,
\text{ if } b \in \Ska,\,
+\infty \text{ o/w},
\label{eq:smooth-topk-alpha-conj}
\end{align}
where $p = \proj_{\Ska(\gamma)}(a + c)$
is the Euclidean projection of $(a + c)$ on $\Ska(\gamma)$.
Moreover,
$L_{\gamma}(a)$ is $1/\gamma$-smooth.
\end{proposition}
\iflongversion
\begin{proof}
We take the convex conjugate of the top-$k$ hinge loss,
which was derived in \cite[Proposition~2]{lapin2015topk},
$$
L^*(b) = 
\begin{cases}
- \inner{c, b} & \text{if } b \in \Ska(1) , \\
+\infty & \text{otherwise} ,
\end{cases}
$$
and add the regularizer $\frac{\gamma}{2} \inner{b,b}$
to obtain the $\gamma$-strongly convex conjugate loss
$L_{\gamma}^*(b)$ as stated in the proposition.
As mentioned above \cite{HirLem2001}
(see also \cite[Lemma~2]{shalev2014accelerated}),
the primal smooth top-$k$ hinge loss $L_{\gamma}(a)$,
obtained as the convex conjugate of
$L_{\gamma}^*(b)$,
is $1/\gamma$-smooth.
We now obtain a formula to compute it based on the Euclidean
projection onto the top-$k$ simplex.
By definition,
\begin{align*}
L_{\gamma}(a) &=
\sup_{b \in \Rb^m} \{ \inner{a,b} - L_{\gamma}^*(b) \} \\
&=
\max_{b \in \Ska(1)}
\Big\{ \inner{a,b} - \tfrac{\gamma}{2} \inner{b,b} + \inner{c, b} \Big\} \\
&=
-\min_{b \in \Ska(1)}
\Big\{ \tfrac{\gamma}{2} \inner{b,b} - \inner{a + c,b} \Big\} \\
&=
- \tfrac{1}{\gamma} \min_{b \in \Ska(1)}
\Big\{ \tfrac{1}{2} \inner{\gamma b, \gamma b}
- \inner{a + c, \gamma b} \Big\} \\
&=
- \tfrac{1}{\gamma} \min_{\frac{b}{\gamma} \in \Ska(1)}
\Big\{ \tfrac{1}{2} \inner{b, b} - \inner{a + c, b} \Big\} .
\end{align*}
For the constraint $\frac{b}{\gamma} \in \Ska(1)$, we have
\begin{align*}
\inner{\ones, b / \gamma} &\leq 1, &
0 &\leq b_i / \gamma \leq
\tfrac{1}{k} \inner{\ones, b / \gamma} \;
\Longleftrightarrow \\
\inner{\ones, b } &\leq \gamma, &
0 &\leq b_i \leq
\tfrac{1}{k} \inner{\ones, b} \;
\Longleftrightarrow 
\; b \in \Ska(\gamma) .
\end{align*}
The final expression follows from the fact that
\begin{align*}
&\argmin_{b \in \Ska(\gamma)}
\big\{ \tfrac{1}{2} \inner{b, b} - \inner{a + c, b} \big\} \\
&\equiv \argmin_{b \in \Ska(\gamma)} \norms{(a + c) - b} 
\equiv \proj_{\Ska(\gamma)}(a + c) .
\end{align*}
\end{proof}
\fi

There is no analytic expression for (\ref{eq:smooth-topk-alpha})
and evaluation requires computing a projection
onto the top-$k$ simplex $\Ska(\gamma)$,
which can be done in $O(m \log m)$ time as shown in \cite{lapin2015topk}.
The non-analytic nature of smooth top-$k$ hinge losses
currently prevents us from proving their top-$k$ calibration.

\subsection{\Topk\ Entropy Loss}
\label{sec:topk-softmax}

As shown in \S~\ref{sec:synthetic} on synthetic data,
top-$1$ and top-$2$ error optimization,
when limited to linear classifiers,
lead to completely different solutions.
The softmax loss, primarily aiming at top-$1$ performance,
produces a solution that is reasonably good in top-$1$ error,
but is far from what can be achieved in top-$2$ error.
That reasoning motivated us to adapt the softmax loss
to top-$k$ error optimization.
Inspired by the conjugate of the top-$k$ hinge loss,
we introduce in this section the top-$k$ entropy loss.

Recall that the conjugate functions of
multiclass SVM \cite{crammer2001algorithmic}
and the top-$k$ SVM \cite{lapin2015topk}
differ only in their effective domain\footnote{
The \textbf{effective domain} of $f$ is
$\dom f = \{x \in X \given f(x) < +\infty \}$.
} while the conjugate function is the same.
Instead of the standard simplex, the conjugate of the
top-$k$ hinge loss is defined on a subset,
the top-$k$ simplex.

This suggests a way to \emph{construct novel losses}
with specific properties by taking the conjugate of an existing loss function,
and modifying its essential domain in a way
that enforces the desired properties.
The motivation for doing so comes from the interpretation of the dual
variables as forces with which every training example pushes
the decision surface in the direction given by the ground truth label.
The absolute value of the dual variables determines the magnitude of these forces
and the optimal values are often attained at the boundary of the feasible set
(which coincides with the essential domain of the loss).
Therefore, by reducing the feasible set we can limit the maximal contribution
of a given training example.

We begin with the \textbf{conjugate} of the \textbf{softmax loss}.
Let $\wo{a}{y}$ be obtained by removing the $y$-th coordinate from $a$.

\begin{proposition}\label{prop:softmax-conj}
The convex conjugate of \eqref{eq:softmax} is
\begin{align}\label{eq:softmax-conj}
\hspace{-5pt}
L^*(v) &= \begin{cases}
\sum_{j \neq y} v_j \log v_j + (1 + v_y) \log(1 + v_y) , \\
\quad\quad\quad\!\text{if }
\inner{\ones, v} = 0 \text{ and } \wo{v}{y} \in \Sm , \\
+ \infty \quad \text{otherwise},
\end{cases}\hspace{-5pt}
\end{align}
where $\Sm \bydef
\big\{ x \given \inner{\ones, x} \leq 1, \;
0 \leq x_j \leq 1, \; \forall j \big\}$.
\end{proposition}
\iflongversion
\begin{proof}
We provide a derivation for the convex conjugate of the softmax loss
which was already given in \cite[Appendix~D.2.3]{mairal2010sparse}
without a proof.
We also highlight the constraint
$\inner{\ones, v} = 0$
which can be easily missed when computing the conjugate
and is re-stated explicitly in Lemma~\ref{lem:conjugate}.

Let $u \bydef f(x) \in \Rb^m$.
The softmax loss on example $x$ is
\begin{align*}
L(u) = \log\Big( \sum_{j \in \Yc} \exp(u_j - u_y) \Big)
= \log\Big( \sum_{j \in \Yc} \exp(u'_j) \Big) ,
\end{align*}
where we let $u' \bydef H_y u$ and $H_y \bydef \Id - \ones \tra{e_y}$.
Let
$$
\phi(u) \bydef \log\Big( \tsum_{j \in \Yc} \exp(u_j) \Big) ,
$$
then $L(u) = \phi(H_y u)$ and
the convex conjugate is computed similar to
\cite[Lemma~2]{lapin2015topk} as follows.
\begin{align*}
L^*(v)
= \sup\{ &\inner{u,v} - L(u) \given u \in \Rb^m \} \\
= \sup\{ &\inner{u,v} - \phi(H_y u) \given u \in \Rb^m \} \\
= \sup\{ &\innern{\para{u},v} + \innern{\ort{u},v} - \phi(H_y \ort{u})
\given \\
& \para{u} \in \Ker H_y, \ort{u} \in \ort{\Ker} H_y \} ,
\end{align*}
where
$\Ker H_y = \{ u \given H_y u = 0 \} = \{ t\ones \given t \in \Rb \}$
and
$\ort{\Ker} H_y = \{ u \given \inner{\ones, u} = 0 \}$.
It follows that $L^*(v)$ can only be finite if
$\innern{\para{u},v} = 0$, which implies
$v \in \ort{\Ker} H_y \Longleftrightarrow \inner{\ones, v} = 0$.
Let $\pinv{H}_y$ be the Moore-Penrose pseudoinverse of $H_y$.
For a $v \in \ort{\Ker} H_y$, we write
\begin{align*}
L^*(v) &=
\sup\{ \innern{\pinv{H}_y H_y \ort{u},v} - \phi(H_y \ort{u}) \given \ort{u} \} \\
&= \sup\{ \innern{z, \tra{(\pinv{H}_y)} v} - \phi(z) \given z \in \Img H_y \} ,
\end{align*}
where
$\Img H_y = \{ H_y u \given u \in \Rb^m \} = \{ u \given u_y = 0 \}$.
Using rank-$1$ update of the pseudoinverse
\cite[\S~3.2.7]{petersen2008matrix}, we have
$$
\tra{(\pinv{H}_y)} = \Id - e_y \tra{e}_y
- \frac{1}{m}(\ones - e_y) \tra{\ones} ,
$$
which together with $\inner{\ones, v} = 0$ implies
$$
\tra{(\pinv{H}_y)} v = v - v_y e_y .
$$
Therefore,
\begin{align*}
L^*(v)
&= \sup\{ \innern{u, v - v_y e_y} - \phi(u) \given u_y = 0 \} \\
&= \sup\Big\{ \innern{\wo{u}{y}, \wo{v}{y}} -
\log\Big( 1 + \sum_{j \neq y} \exp(u_j) \Big) \Big\} .
\end{align*}
The function inside $\sup$ is concave and differentiable,
hence the global optimum is at the critical point \cite{boyd2004convex}.
Setting the partial derivatives to zero yields
\begin{align*}
v_j = \exp(u_j) / \big(1 + \tsum_{j \neq y} \exp(u_j) \big)
\end{align*}
for $j \neq y$, from which we conclude,
similar to \cite[\S~5.1]{shalev2014accelerated},
that
$\inner{\ones, v} \leq 1$ and
$0 \leq v_j \leq 1$ for all $j \neq y$, \ie
$\wo{v}{y} \in \Sm$.
Let $Z \bydef \sum_{j \neq y} \exp(u_j)$, we have at the optimum
\begin{align*}
u_j &= \log(v_j) + \log(1 + Z), \quad
\forall j \neq y .
\end{align*}
Since $\inner{\ones, v} = 0$, we also have that
$v_y = - \sum_{j \neq y} v_j$, hence
\begin{align*}
L^*(v)
&= \tsum_{j \neq y} u_j v_j - \log(1 + Z) \\
&= \tsum_{j \neq y} v_j \log(v_j) + \log(1 + Z)
\Big( \sum_{j \neq y} v_j - 1 \Big) \\
&= \tsum_{j \neq y} v_j \log(v_j) - \log(1 + Z)(1 + v_y) .
\end{align*}
Summing $v_j$ and using the definition of $Z$,
\begin{align*}
\sum_{j \neq y} v_j
= \sum_{j \neq y} \exp(u_j) / \big(1 + \sum_{j \neq y} \exp(u_j) \big)
= Z / (1 + Z) .
\end{align*}
Therefore,
\begin{align*}
1 + Z = 1 / \big(1 - \tsum_{j \neq y} v_j \big) = 1 / (1 + v_y) ,
\end{align*}
which finally yields
\begin{align*}
L^*(v)
&= \tsum_{j \neq y} v_j \log(v_j) + \log(1 + v_y)(1 + v_y) ,
\end{align*}
if $\inner{\ones, v} = 0$ and $\wo{v}{y} \in \Sm$
as stated in the proposition.
\end{proof}
\fi

The \textbf{conjugate} of the \textbf{top-$k$ entropy loss} is obtained by replacing
$\Sm$ in (\ref{eq:softmax-conj}) with $\Ska$.
A $\beta$ version could be obtained using the $\Skb$ instead, which defer to future work.
There is no closed-form solution for the primal
top-$k$ entropy loss for $k > 1$,
but we can evaluate it as follows.

\begin{proposition}\label{prop:topk-entropy-primal}
Let $u_j \bydef f_j(x) - f_y(x)$ for all $j \in \Yc$.
The \textbf{top-$k$ entropy loss} is defined as
\begin{equation}\label{eq:topk-entropy-primal}
\begin{aligned}
L(u) = \max&\big\{
\innern{\wo{u}{y},x} - (1-s) \log(1-s) \\
&- \inner{x, \log x}
\given x \in \Ska, \; \inner{\ones, x} = s \big\} .
\end{aligned}
\end{equation}
Moreover, we recover the softmax loss \eqref{eq:softmax} if $k=1$.
\end{proposition}
\iflongversion
\begin{proof}
The convex conjugate of the top-$k$ entropy loss is
\begin{align*}
L^*(v) &\bydef \begin{cases}
\sum_{j \neq y} v_j \log v_j + (1 + v_y) \log(1 + v_y) , \\
\quad\quad\quad\!\text{if }
\inner{\ones, v} = 0 \text{ and } \wo{v}{y} \in \Ska , \\
+ \infty \quad \text{otherwise},
\end{cases}
\end{align*}
where the setting is the same as in
Proposition~\ref{prop:softmax-conj}.
The (primal) top-$k$ entropy loss is defined as the convex conjugate
of the $L^*(v)$ above.
We have
\begin{align*}
L(u)
= \sup\{ &\inner{u,v} - L^*(v) \given v \in \Rb^m \} \\
= \sup\{ &\inner{u,v} - \sum_{j \neq y} v_j \log v_j 
- (1 + v_y) \log(1 + v_y) \\
&\given
\inner{\ones, v} = 0, \; \wo{v}{y} \in \Ska \} \\
= \sup\{ &\innern{\wo{u}{y},\wo{v}{y}} - u_y \sum_{j \neq y} v_j
- \sum_{j \neq y} v_j \log v_j \\
& - (1 - \sum_{j \neq y} v_j) \log(1 - \sum_{j \neq y} v_j)
\given \wo{v}{y} \in \Ska \} .
\end{align*}
Note that $u_y = 0$, and hence the corresponding term vanishes.
Finally, we let $x \bydef \wo{v}{y}$
and $s \bydef \sum_{j \neq y} v_j = \inner{\ones, x}$
and obtain (\ref{eq:topk-entropy-primal}).

Next, we discuss how this problem can be solved and show that it
reduces to the softmax loss for $k=1$.
Let $a \bydef \wo{u}{y}$ and consider an equivalent problem below.
\begin{equation}\label{eq:topk-entropy-primal-min}
\begin{aligned}
L(u) = -\min\big\{&
\inner{x, \log x} + (1-s) \log(1-s) \\
&- \innern{a,x} \given x \in \Ska, \; \inner{\ones, x} = s \big\} .
\end{aligned}
\end{equation}
The Lagrangian for (\ref{eq:topk-entropy-primal-min}) is
\begin{align*}
\Lc
&= \inner{x, \log x} + (1-s) \log(1-s) - \innern{a,x} \\
&+ t(\inner{\ones, x} - s) + \lambda (s - 1)
- \inner{\mu, x} + \inner{\nu, x - \tfrac{s}{k} \ones} ,
\end{align*}
where $t \in \Rb$ and $\lambda, \mu, \nu \geq 0$ are the dual variables.
Computing the partial derivatives of $\Lc$ \wrt $x_j$ and $s$,
and setting them to zero, we obtain
\begin{align*}
\log x_j &= a_j - 1 - t + \mu_j - \nu_j , \quad \forall j \\
\log(1-s) &= -1 - t - \tfrac{1}{k} \inner{\ones, \nu} + \lambda .
\end{align*}
Note that $x_j = 0$ and $s = 1$ cannot satisfy the above conditions
for any choice of the dual variables in $\Rb$.
Therefore, $x_j > 0$ and $s < 1$, which implies
$\mu_j = 0$ and $\lambda = 0$.
The only constraint that might be active is $x_j \leq \frac{s}{k}$.
Note, however, that in view of $x_j > 0$
it can only be active if either $k > 1$ or
we have a one dimensional problem.
We consider the case when this constraint is active below.

Consider $x_j$'s for which
$0 < x_j < \frac{s}{k}$ holds at the optimum.
The complementary slackness conditions imply that
the corresponding $\mu_j = \nu_j = 0$.
Let $p \bydef \inner{\ones, \nu}$ and re-define $t$
as $t \leftarrow 1 + t$.
We obtain the simplified equations
\begin{align*}
\log x_j &= a_j - t , \\
\log(1-s) &= - t - \tfrac{p}{k} .
\end{align*}
If $k=1$, then $0 < x_j < s$ for all $j$
in a multiclass problem as discussed above,
hence also $p=0$.
We have
\begin{align*}
x_j &= e^{a_j - t} , &
1-s &= e^{-t} ,
\end{align*}
where $t \in \Rb$ is to be found.
Plugging that into the objective,
\begin{align*}
&\tsum_j (a_j - t) e^{a_j - t} - t e^{-t} - \tsum_j a_j e^{a_j - t} \\
&= e^{-t} \Big[
\tsum_j (a_j - t) e^{a_j} - t - \tsum_j a_j e^{a_j} \Big] \\
&= -t e^{-t} \big[ 1 + \tsum_j e^{a_j} \big] 
= -t \big[ e^{-t} + \tsum_j e^{a_j - t} \big] \\
&= -t \big[ 1 - s + s \big] = -t .
\end{align*}
To compute $t$, we note that
$$
\tsum_j e^{a_j - t} = \inner{\ones, x} = s = 1 - e^{-t} ,
$$
from which we conclude
\begin{align*}
1 &= \big( 1 + \sum_j e^{a_j} \big) e^{-t} \; \Longrightarrow \;
-t = - \log(1 + \tsum_j e^{a_j}) .
\end{align*}
Taking into account the minus in front of the $\min$ in
(\ref{eq:topk-entropy-primal-min})
and the definition of $a$, we finally recover the softmax loss
\begin{align*}
L(y,f(x)) = \log\big(1 + \tsum_{j \neq y} \exp(f_j(x) - f_y(x)) \big) .
\end{align*}
\end{proof}
\fi

\iflongversion
The non-analytic nature of the loss for $k>1$ does not allow us to check
if it is top-$k$ calibrated.
We now show how this problem can be solved efficiently.

\textbf{How to solve (\ref{eq:topk-entropy-primal}).}
We continue the derivation started in the proof of
Propostion~\ref{prop:topk-entropy-primal}.
First, we write the system that follows directly
from the KKT \cite{boyd2004convex} optimality conditions.
\begin{equation}\label{eq:topk-entropy-kkt}
\begin{aligned}
x_j &= \min\{ \exp(a_j - t), \tfrac{s}{k} \} , \quad \forall j, \\
\nu_j &= \max\{ 0, a_j - t - \log(\tfrac{s}{k}) \} , \quad \forall j, \\
1-s &= \exp(-t - \tfrac{p}{k}) , \\
s &= \inner{\ones, x} , \quad
p = \inner{\ones, \nu} .
\end{aligned}
\end{equation}
Next, we define the two index sets $U$ and $M$ as follows
\begin{align*}
U &\bydef \{ j \given x_j = \tfrac{s}{k} \}, &
M &\bydef \{ j \given x_j < \tfrac{s}{k} \} .
\end{align*}
Note that the set $U$ contains at most $k$ indexes corresponding
to the largest components of $a_j$.
Now, we proceed with finding a $t$ that solves (\ref{eq:topk-entropy-kkt}).
Let $\rho \bydef \frac{\abs{U}}{k}$.
We eliminate $p$ as
\begin{align*}
p &= \sum_j \nu_j
= \sum_U a_j - \abs{U} \big(t + \log(\tfrac{s}{k}) \big)
\quad \Longrightarrow \\
\tfrac{p}{k}
&= \tfrac{1}{k} \sum_U a_j - \rho \big(t + \log(\tfrac{s}{k})\big) .
\end{align*}
Let $Z \bydef \sum_M \exp a_j$, we write for $s$
\begin{align*}
s &= \sum_j x_j
= \sum_U \tfrac{s}{k} + \sum_M \exp(a_j - t) \\
&= \rho s + \exp(-t) \sum_M \exp a_j
= \rho s + \exp(-t) Z .
\end{align*}
We conclude that
\begin{align*}
(1 - \rho) s &= \exp(-t) Z \quad \Longrightarrow \\
t &= \log Z - \log\big( (1-\rho)s \big) .
\end{align*}
Let $\alpha \bydef \tfrac{1}{k} \sum_U a_j$.
We further write
\begin{align*}
\log(1-s) =& -t - \tfrac{p}{k} \\
=& -t - \alpha + \rho \big(t + \log(\tfrac{s}{k})\big) \\
=& \rho \log(\tfrac{s}{k}) - (1 - \rho) t - \alpha \\
=& \rho \log(\tfrac{s}{k}) - \alpha \\
&- (1 - \rho) \big[ \log Z - \log\big( (1-\rho)s \big) \big] ,
\end{align*}
which yields the following equation for $s$
\begin{align*}
&\log(1-s) - \rho ( \log s - \log k)  + \alpha \\
&+ (1 - \rho) \big[ \log Z - \log(1-\rho) - \log s \big] = 0.
\end{align*}
Therefore,
\begin{align*}
&\log(1-s) - \log s + \rho \log k + \alpha \\
&+ (1-\rho) \log Z
- (1-\rho)\log(1-\rho) = 0 , \\
&\log\left( \frac{1-s}{s} \right) =
\log\left(
\frac{(1-\rho)^{(1-\rho)} \exp(-\alpha)}{k^\rho Z^{(1-\rho)}}
\right) .
\end{align*}
We finally get
\begin{equation}\label{eq:topk-entropy-sQ}
\begin{aligned}
s &= 1 / (1 + Q), \\
Q &\bydef (1-\rho)^{(1-\rho)} / (k^\rho Z^{(1-\rho)} e^\alpha) .
\end{aligned}
\end{equation}
We note that:
\emph{a)} $Q$ is readily computable once the sets $U$ and $M$ are fixed;
and
\emph{b)} $Q = 1/Z$ if $k=1$ since $\rho = \alpha = 0$ in that case.
This yields the formula for $t$ as
\begin{align}\label{eq:topk-entropy-t}
t = \log Z + \log(1 + Q) - \log(1-\rho).
\end{align}
As a sanity check, we note that we again recover the softmax loss
for $k=1$, since
$t = \log Z + \log(1 + 1/Z) = \log(1 + Z) = \log(1 + \sum_j \exp a_j)$.

To verify that the computed $s$ and $t$ are compatible with the choice
of the sets $U$ and $M$, we check if this holds:
\begin{align*}
\exp(a_j - t) &\geq \tfrac{s}{k}, \quad \forall j \in U , \\
\exp(a_j - t) &\leq \tfrac{s}{k}, \quad \forall j \in M ,
\end{align*}
which is equivalent to
\begin{align}\label{eq:topk-entropy-feasible}
\max_M a_j \leq \log(\tfrac{s}{k}) + t \leq \min_U a_j .
\end{align}

\textbf{Computation of the top-$k$ entropy loss (\ref{eq:topk-entropy-primal}).}
The above derivation suggests a simple and efficient algorithm
to compute $s$ and $t$ that solve the KKT system (\ref{eq:topk-entropy-kkt})
and, therefore, the original problem (\ref{eq:topk-entropy-primal}).
\begin{enumerate}
\item Initialization: $U \leftarrow \{\}$, $M \leftarrow \{1, \ldots, m\}$.

\item\label{alg:topk-entropy-loop}
$Z \leftarrow \sum_M \exp a_j$,
$\alpha \leftarrow \frac{1}{k} \sum_U a_j$,
$\rho \leftarrow \frac{\abs{U}}{k}$.

\item Compute $s$ and $t$ using \eqref{eq:topk-entropy-sQ} and
\eqref{eq:topk-entropy-t}.

\item If \eqref{eq:topk-entropy-feasible} holds, stop;
otherwise, $j \leftarrow \argmax_M a_j$.

\item $U \leftarrow U \cup \{j\}$,
$M \leftarrow M \setminus \{j\}$
and go to step~\ref{alg:topk-entropy-loop}.
\end{enumerate}
Note that the algorithm terminates after at most $k$ iterations
since $\abs{U} \leq k$.
The overall complexity is therefore $O(km)$.

To compute the actual loss (\ref{eq:topk-entropy-primal}),
we note that if $U$ is empty, \ie there were no violated constraints,
then the top-$k$ entropy loss coincides with the softmax loss
and is directly given by $t$.
Otherwise, we have
\begin{align*}
&\inner{a,x} - \inner{x, \log x} - (1-s)\log(1-s) \\
&= \sum_U a_j \tfrac{s}{k} + \sum_M a_j \exp(a_j - t)
- \sum_U \tfrac{s}{k} \log(\tfrac{s}{k}) \\
&- \sum_M (a_j - t) \exp(a_j - t)
- (1-s)\log(1-s) \\
&= \alpha s - \rho s \log(\tfrac{s}{k})
+ t \exp(-t) Z - (1-s)\log(1-s) \\
&= \alpha s - \rho s \log(\tfrac{s}{k})
+ (1-\rho) s t - (1-s)\log(1-s) .
\end{align*}
Therefore, the top-$k$ entropy loss is readily computed
once the optimal $s$ and $t$ are found.

\fi

\subsection{Truncated \Topk\ Entropy Loss}
\label{sec:topk-nonconvex}
A major limitation of the softmax loss for top-$k$ error optimization
is that it cannot ignore the highest scoring predictions,
which yields a high loss even if the top-$k$ error is zero.
This can be seen by rewriting \eqref{eq:softmax} as
\begin{equation}\label{eq:softmax-2}
L(y,f(x)) = \log\big( 1 + \vsum_{j\neq y} \exp(f_{j}(x) - f_y(x)) \big).
\end{equation}
If there is only a \emph{single} $j$ such that $f_j(x) - f_y(x) \gg 0$,
then $L(y,f(x)) \gg 0$ even though $\nerr{2}{y, f(x)} = 0$.

This problem is is also present in all top-$k$ hinge losses
considered above and is an inherent limitation due to their convexity.
The origin of the problem is the fact that ranking
based losses \cite{usunier2009ranking} are based on functions such as
\[ \phi(f(x)) = \tfrac{1}{m} \tsum_{j \in \Yc} \alpha_j f_{\pi_j}(x) - f_y(x) .\]
The function $\phi$ is convex if the sequence $(\alpha_j)$
is monotonically non-increasing \cite{boyd2004convex}.
This implies that convex ranking based losses have to put \emph{more} weight
on the highest scoring classifiers,
while we would like to put \emph{less} weight on them.
To that end, we drop the first $(k-1)$ highest scoring
predictions from the sum in \eqref{eq:softmax-2},
sacrificing convexity of the loss,
and define the \textbf{truncated top-$k$ entropy loss} as follows
\begin{align}\label{eq:truncated-topk-entropy}
\vneghspace
L(y,f(x)) = \log\big( 1 + \vsum_{j \in \Jc_y} \exp(f_j(x) - f_y(x)) \big),
\vneghspace
\end{align}
where $\Jc_y$
are the indexes corresponding to the $(m - k)$ \emph{smallest}
components of $(f_j(x))_{j \neq y}$.
This loss can be seen as a smooth version
of the top-$k$ error \eqref{eq:topk-error},
as it is small whenever the top-$k$ error is zero.
Below, we show that this loss is top-$k$ calibrated.

\begin{proposition}\label{prop:calibrated-truncated-topk}
The truncated top-$k$ entropy loss is top-$s$ calibrated
for any $k \leq s \leq m$.
\end{proposition}
\iflongversion
\begin{proof}
Given a $g \in \Rb^m$, let $\pi$ be a permutation such that
$g_{\pi_1} \geq g_{\pi_2} \geq \ldots \geq g_{\pi_m}$.
Then, we have
\begin{align*}
\Jc_y = \begin{cases}
\{ \pi_{k+1}, \ldots, \pi_m \} &\text{if } y \in \{ \pi_1, \ldots, \pi_{k-1} \}, \\
\{ \pi_{k}, \ldots, \pi_m \} \setminus \{y\} &\text{if } y \in \{ \pi_{k}, \ldots, \pi_m \}.
\end{cases}
\end{align*}
Therefore, the expected loss at $x$ can be written as
\begin{align*}
\Exp_{Y|X}&[L(Y,g) \given X = x] = \tsum_{y \in \Yc} L(y,g) \, p_y(x) \\
&= \tsum_{r=1}^{k-1}
\log\big(1 + \tsum_{j=k+1}^m e^{g_{\pi_j} - g_{\pi_r}} \big) \, p_{\pi_r}(x) \\
&+
\tsum_{r=k}^{m}
\log\big(\tsum_{j=k}^m e^{g_{\pi_j} - g_{\pi_r}} \big) \, p_{\pi_r}(x) .
\end{align*}
Note that the sum inside the logarithm does not depend on $g_{\pi_r}$ for $r<k$.
Therefore, a Bayes optimal classifier will have $g_{\pi_r} = +\infty$ for all $r<k$
as then the first sum vanishes.

Let $p \bydef (p_y(x))_{y \in \Yc}$ and $q \bydef (L(y, g))_{y \in \Yc}$,
then
\begin{align*}
q_{\pi_1} = \ldots = q_{\pi_{k-1}} = 0 \leq q_{\pi_k} \leq \ldots \leq q_{\pi_m}
\end{align*}
and we can re-write the expected loss as
\begin{align*}
\Exp_{Y|X}&[L(Y,g) \given X = x] = \inner{p, q} 
= \inner{p_{\pi}, q_{\pi}} \geq \inner{p_{\tau}, q_{\pi}} ,
\end{align*}
where $p_{\tau_1} \geq p_{\tau_2} \geq \ldots \geq p_{\tau_m}$
and we used the rearrangement inequality.
Therefore, the expected loss is minimized when $\pi$ and $\tau$ coincide
(up to a permutation of the first $k-1$ elements,
since they correspond to zero loss).

We can also derive a Bayes optimal classifier following the proof of
Proposition~\ref{prop:calibrated-softmax}.
We have
\begin{align*}
\Exp_{Y|X}&[L(Y,g) \given X = x] \\
&=\tsum_{r=k}^{m}
\log\big(\tsum_{j=k}^m e^{g_{\tau_j} - g_{\tau_r}} \big) \, p_{\tau_r}(x) \\
&=\tsum_{r=k}^{m} \Big(
\log\big(\tsum_{j=k}^m e^{g_{\tau_j}} \big) - g_{\tau_r} \Big) \, p_{\tau_r}(x) .
\end{align*}
A critical point is found by setting partial derivatives to zero
for all $y \in \{\tau_k, \ldots, \tau_m\}$, which leads to
\begin{align*}
\frac{e^{g_y}}{\tsum_{j=k}^m e^{g_{\tau_j}}}
\tsum_{r=k}^{m} p_{\tau_r}(x) = p_y(x) .
\end{align*}
We let $g_y = -\infty$ if $p_y(x) = 0$, and obtain finally
\begin{align*}
g^*_{\tau_j} =
\begin{cases}
+\infty &\text{if } j < k, \\
\log\big(\alpha p_{\tau_j}(x)\big)
&\text{if } j \geq k \text{ and } p_{\tau_j}(x) > 0, \\
-\infty &\text{if } j \geq k \text{ and } p_{\tau_j}(x) = 0,
\end{cases}
\end{align*}
as a Bayes optimal classifier for any $\alpha > 0$.

Note that $g^*$ preserves the ranking of
$p_y(x)$ for all $y$ in $\{\tau_k, \ldots, \tau_m\}$,
hence, it is top-$s$ calibrated for all $s \geq k$.
\end{proof}
\fi

As the loss \eqref{eq:truncated-topk-entropy} is nonconvex,
we use solutions obtained with the softmax loss \eqref{eq:softmax}
as initial points and optimize them further via gradient descent.
However, the resulting optimization problem seems to be ``mildly nonconvex''
as the same-quality solutions are obtained from different initializations.
In Section \ref{sec:synthetic}, we show a synthetic experiment,
where the advantage of discarding the highest scoring classifier in
the loss becomes apparent.

\section{Optimization Method}
\label{sec:optimization}
In this section, we briefly discuss how the proposed
smooth top-$k$ hinge losses and the top-$k$ entropy loss
can be optimized efficiently
within the SDCA framework of \cite{shalev2014accelerated}.

\textbf{The primal and dual problems.}
Let $X \in \Rb^{d \times n}$ be the matrix of training examples $x_i \in \Rb^d$,
$K = \tra{X}\!X$ the corresponding Gram matrix,
$W \in \Rb^{d \times m}$ the matrix of primal variables,
$A \in \Rb^{m \times n}$ the matrix of dual variables,
and
$\lambda > 0$ the regularization parameter.
The primal and Fenchel dual \cite{borwein2000convex}
objective functions are given as
\begin{equation}\label{eq:primal-dual}
\hspace{-0.5em}
\begin{aligned}
P(W) &=
+\frac{1}{n} \sum_{i=1}^n
L \left( y_i, \tra{W}x_i \right)
+ \frac{\lambda}{2} \tr\left(\tra{W} W \right) ,
\\
D(A) &= 
-\frac{1}{n} \sum_{i=1}^n
L^* \left( y_i, - \lambda n a_i \right)
- \frac{\lambda}{2} \tr\left( A K \tra{A} \right) ,
\end{aligned}
\hspace{-0.5em}
\end{equation}
where $L^*$ is the convex conjugate of $L$.
SDCA proceeds by randomly picking a variable $a_i$
(which in our case is a vector of dual variables over all $m$ classes
for a sample $x_i$) and modifying it to achieve maximal increase
in the dual objective $D(A)$.
It turns out that this update step is equivalent to a proximal problem,
which can be seen as a regularized projection
onto the essential domain of $L^*$.

\iflongversion
\textbf{The convex conjugate.}
An important ingredient in the SDCA framework is the
convex conjugate $L^*$.
We show that for all multiclass loss functions that we consider
the fact that they depend on the differences
$f_j(x) - f_y(x)$ enforces a certain constraint on the conjugate function.

\begin{lemma}\label{lem:conjugate}
Let $H_y = \Id - \ones e_y^{\top}$ and
let $\Phi(u) = \phi(H_y u)$.
$\Phi^*(v) = +\infty$ unless $\inner{\ones,v} = 0$.
\end{lemma}
\iflongversion
\begin{proof}
The proof follows directly from \cite[Lemma~2]{lapin2015topk}
and was already reproduced in the proof
of Proposition~\ref{prop:softmax-conj} for the softmax loss.
We have formulated this simplified lemma
since \cite[Lemma~2]{lapin2015topk} additionally required
$y$-compatibility to show that if $\inner{\ones,v} = 0$,
then $\Phi^*(v) = \phi^*(v - v_y e_y)$,
which does not hold \eg for the softmax loss.
\end{proof}
\fi

Lemma~\ref{lem:conjugate} tells us that we need to enforce
$\inner{\ones, a_i} = 0$ at all times, which translates into
$a_{y_i} = - \sum_{j \neq y_i} a_{j}$.
The update steps are performed on the $(m-1)$-dimensional vector
obtained by removing the coordinate $a_{y_i}$.
\fi

\textbf{The update step for \SvmTopKag{k}{\gamma}.}
Let $\wo{a}{y}$ be obtained by removing the $y$-th coordinate from vector $a$.
We show that performing an update step for
the smooth top-$k$ hinge loss is equivalent to
projecting a certain vector $b$,
computed from the prediction scores $\tra{W}x_i$,
onto the essential domain of $L^*$,
the top-$k$ simplex,
with an added regularization $\rho \inner{\ones, x}^2$,
which biases the solution to be orthogonal to $\ones$.

\begin{proposition}\label{prop:topk-smooth-update}
Let $L$ and $L^*$ in (\ref{eq:primal-dual}) be
respectively
the \SvmTopKag{k}{\gamma} loss and its conjugate as in
Proposition~\ref{prop:topk-smooth}.
The update
$\max_{a_i} \{ D(A) \given \inner{\ones, a_i} = 0 \}$
is equivalent with the change of variables
$x \leftrightarrow - \wo{a_i}{y_i}$ to solving
\begin{align}\label{eq:smooth-hinge-update}
\min_{x} \{ \norms{x - b} + \rho \inner{\ones, x}^2
\given x \in \Ska(\tfrac{1}{\lambda n}) \} ,
\end{align}
where
$b = \frac{1}{\inner{x_i,x_i} + \gamma n \lambda}
\left( \wo{q}{y_i} + (1 - q_{y_i})\ones \right)$,\\
$q = \tra{W} x_i - \inner{x_i,x_i} a_i$,
and
$\rho = \frac{\inner{x_i,x_i}}{\inner{x_i,x_i} + \gamma n \lambda}$.
\end{proposition}
\iflongversion
\begin{proof}
We follow the proof of \cite[Proposition~4]{lapin2015topk}.
We choose $i \in \{1, \ldots, n\}$ and, having all other variables fixed,
update $a_i$ to maximize
$$
-\tfrac{1}{n} L^* \left(y_i, - \lambda n a_i \right)
- \tfrac{\lambda}{2} \tr\left( A K \tra{A} \right) .
$$
For the nonsmooth top-$k$ hinge loss,
it was shown \cite{lapin2015topk} that
$$
L^* \left(y_i, - \lambda n a_i \right) =
\inner{c, \lambda n ( a_i - a_{y_i,i} e_{y_i}) }
$$
if $- \lambda n ( a_i - a_{y_i,i} e_{y_i}) \in \Ska$ and
$+ \infty$ otherwise.
Now, for the smoothed loss,
we add regularization and obtain
\begin{align*}
-\tfrac{1}{n} \left(
\tfrac{\gamma}{2} \norms{- \lambda n ( a_i - a_{y_i,i} e_{y_i})}
+ \inner{c, \lambda n ( a_i - a_{y_i,i} e_{y_i}) } \right)
\end{align*}
with $- \lambda n ( a_i - a_{y_i,i} e_{y_i}) \in \Ska$.
Using $c = \ones - e_{y_i}$ and $\inner{\ones, a_i} = 0$,
one can simplify it to
\begin{align*}
- \frac{\gamma n \lambda^2}{2} \normsb{\wo{a_i}{y_i}}
+ \lambda a_{y_i,i} ,
\end{align*}
and the feasibility constraint can be re-written as
\begin{align*}
-\wo{a_i}{y_i} &\in \Ska(\tfrac{1}{\lambda n}) , &
a_{y_i,i} &= \innern{\ones, -\wo{a_i}{y_i}} .
\end{align*}
For the regularization term $\tr\left( A K \tra{A} \right)$, we have
\begin{align*}
\tr\left( A K \tra{A} \right) =
K_{ii} \inner{a_i,a_i} + 2 \sum_{j \neq i} K_{ij} \inner{a_i,a_j}
+ {\rm const} .
\end{align*}
We let
$q = \sum_{j \neq i} K_{ij} a_j = A K_i - K_{ii} a_i$
and $x = -\wo{a_i}{y_i}$:
\begin{align*}
\inner{a_i,a_i} &= \inner{\ones,x}^2 + \inner{x,x}, \\
\inner{q,a_i} &= q_{y_i} \inner{\ones,x} - \innern{\wo{q}{y_i},x}.
\end{align*}
Now, we plug everything together and multiply with $-2/\lambda$.
\begin{align*}
\min_{x \in \Ska(\frac{1}{\lambda n})}
& \gamma n \lambda \norms{x}
- 2 \inner{\ones, x}
+ 2 \big(
q_{y_i} \inner{\ones,x} - \innern{\wo{q}{y_i},x} \big) \\
&+ K_{ii} \big(\inner{\ones,x}^2 + \inner{x,x} \big) .
\end{align*}
Collecting the corresponding terms finishes the proof.
\end{proof}
\fi

Note that setting $\gamma = 0$, we recover the update step
for the non-smooth top-$k$ hinge loss \cite{lapin2015topk}.
It turns out that we can employ their projection
procedure for solving (\ref{eq:smooth-hinge-update})
with only a minor modification of $b$ and $\rho$.

The update step for the \SvmTopKbg{k}{\gamma} loss is derived similarly
using the set $\Skb$ in \eqref{eq:smooth-hinge-update}
instead of $\Ska$.
The resulting projection problem is
a biased continuous quadratic knapsack problem,
which is discussed in the supplement of \cite{lapin2015topk}.

Smooth top-$k$ hinge losses converge significantly faster 
than their nonsmooth variants as we show in the scaling experiments below. 
This can be explained by the theoretical results of  \cite{shalev2014accelerated} 
on the convergence rate of SDCA.
They also had similar observations for the smoothed binary hinge loss.

\textbf{The update step for \LrTopK{k}.}
We now discuss the optimization of the proposed top-$k$ entropy loss
in the SDCA framework.
Note that the top-$k$ entropy loss reduces to the softmax loss for $k=1$.
Thus, our SDCA approach can be used for
\emph{gradient-free} optimization of the softmax loss without having to
tune step sizes or learning rates.

\begin{proposition}\label{prop:topk-entropy-update}
Let $L$ in (\ref{eq:primal-dual}) be the
\LrTopK{k} loss (\ref{eq:topk-entropy-primal})
and $L^*$ be its convex conjugate as in (\ref{eq:softmax-conj})
with $\Sm$ replaced by $\Ska$.
The update
$\max_{a_i} \{ D(A) \given \inner{\ones, a_i} = 0 \}$
is equivalent with the change of variables
$x \leftrightarrow - \lambda n \wo{a_i}{y_i}$ to solving
\begin{equation}\label{eq:topk-ent-update}
\begin{aligned}
\min_{x \in \Ska} \; &
\tfrac{\alpha}{2} ( \inner{x,x} + \inner{\ones,x}^2 )
- \inner{b, x} + \\
&
\inner{x, \log x}
+ (1 - \inner{\ones,x}) \log(1 - \inner{\ones,x}) \\
\end{aligned}
\end{equation}
where
$\alpha = \frac{\inner{x_i, x_i}}{\lambda n}$,
$b = \wo{q}{y_i} - q_{y_i}\ones$,
$q = \tra{W} x_i - \inner{x_i,x_i} a_i$.
\end{proposition}
\iflongversion
\begin{proof}
Let $v \bydef - \lambda n a_i$ and $y = y_i$.
Using Proposition~\ref{prop:softmax-conj},
\begin{align*}
L^*(y, v) =
\tsum_{j \neq y} v_j \log v_j + (1 + v_y) \log(1 + v_y) ,
\end{align*}
where $\inner{\ones, v} = 0$ and $\wo{v}{y} \in \Ska$.
Let $x \bydef \wo{v}{y}$ and $s \bydef - v_y$.
It follows that $s = \inner{\ones, x}$ and from $\tr\left( A K \tra{A} \right)$ we get
\begin{align*}
K_{ii} (\inner{x,x} + s^2) / (\lambda n)^2 
- 2 \innerb{\wo{q}{y} - q_{y} \ones, x} / (\lambda n) ,
\end{align*}
where
$q = \sum_{j \neq i} K_{ij} a_j = A K_i - K_{ii} a_i$
as before.
Finally, we plug everything together as in Proposition~\ref{prop:topk-smooth-update}.
\end{proof}
\fi

Note that this optimization problem is similar
to (\ref{eq:smooth-hinge-update}),
but is more difficult to solve due to
the presence of logarithms in the objective.
We propose to tackle this problem 
using the Lambert $W$ function introduced below.

\textbf{Lambert $W$ function.}
The Lambert $W$ function is defined to be the inverse of the function
$w \mapsto w e^w$ and is widely used in many fields
\cite{corless1996lambertw, Fukushima201377, Veberic20122622}.
Taking logarithms on both sides of the defining equation $z = W e^W$,
we obtain $\log z = W(z) + \log W(z)$.
Therefore, if we are given an equation of the form
$x + \log x = t$ for some $t \in \Rb$,
we can directly ``solve'' it in closed-form as
$x = W(e^t)$.
The crux of the problem is that the function
$V(t) \bydef W(e^t)$ is transcendental \cite{Fukushima201377}
just like the logarithm and the exponent.
There exist highly optimized implementations for the latter 
and we argue that the same can be done for the Lambert $W$ function.
In fact, there is already some work on this topic
\cite{Fukushima201377, Veberic20122622}, which we also employ in our
implementation.

\iflongversion
\begin{figure}[ht]\small\centering%
\begin{subfigure}[t]{0.49\columnwidth}\centering
\includegraphics[width=\columnwidth]{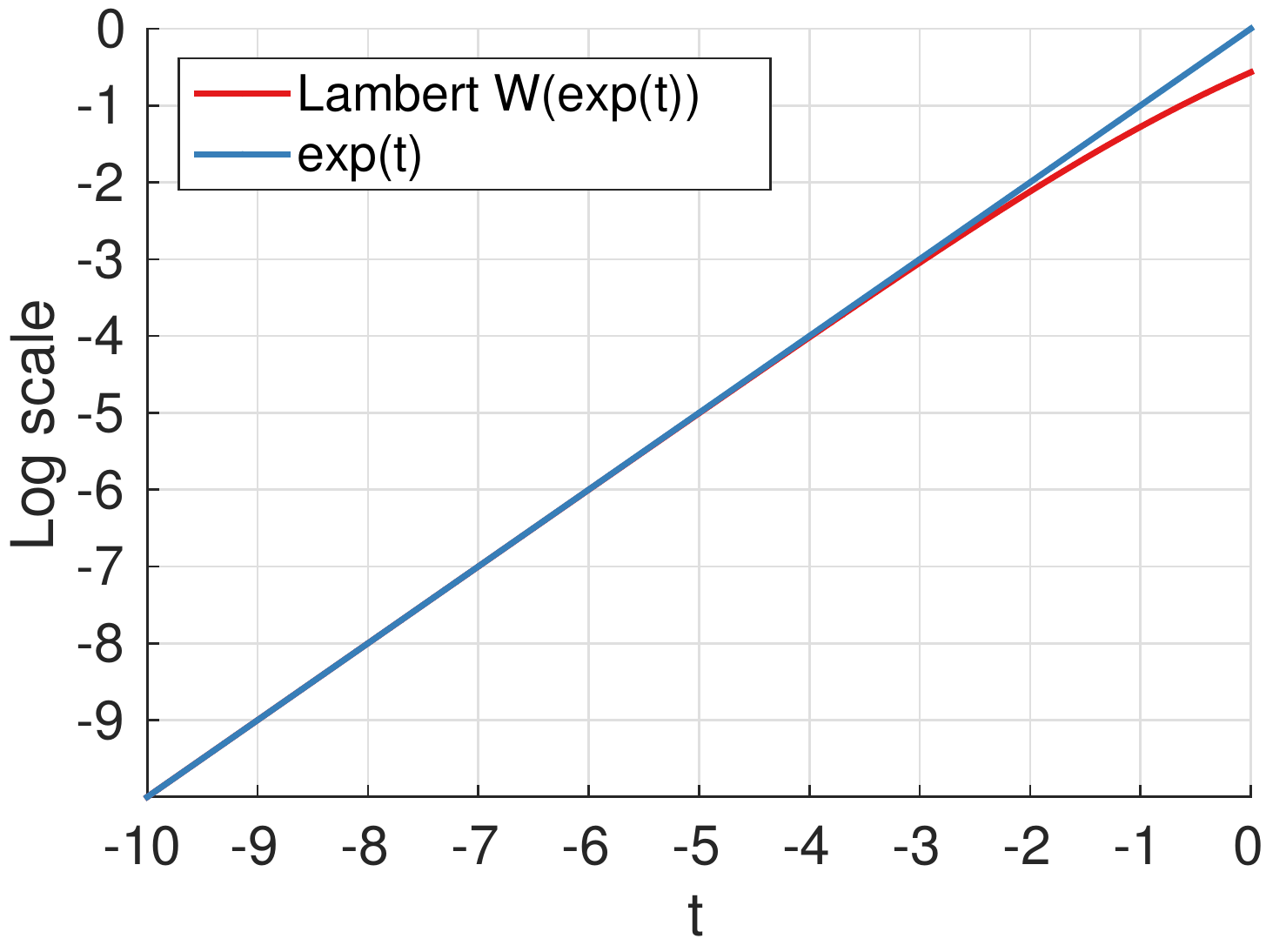}%
\caption{$V(t) \approx e^t$ for $t \ll 0$.}\label{fig:lambert:a}
\end{subfigure}
\begin{subfigure}[t]{0.49\columnwidth}\centering
\includegraphics[width=\columnwidth]{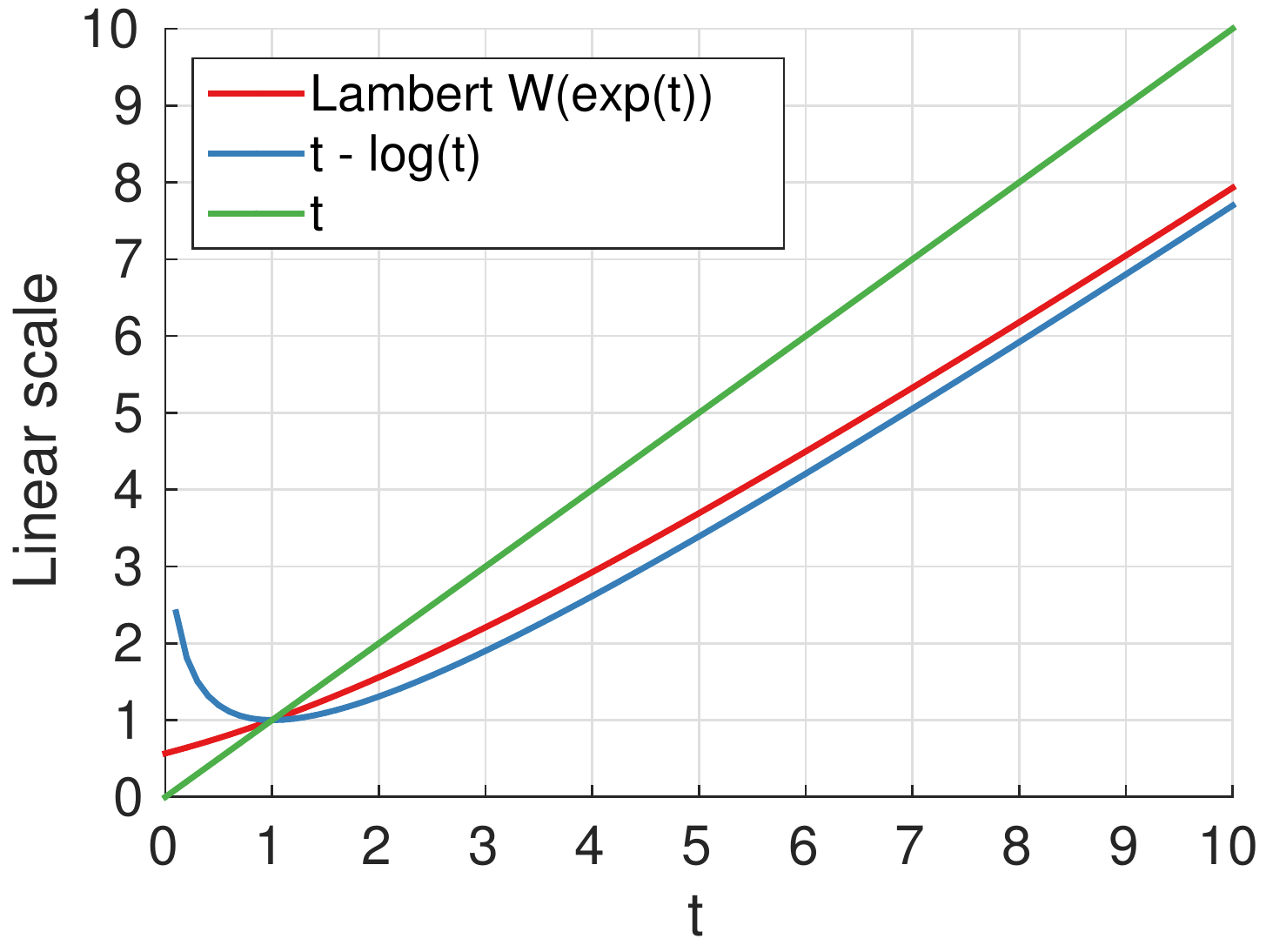}%
\caption{$V(t) \approx t - \log t$ for $t \gg 0$.}\label{fig:lambert:b}
\end{subfigure}
\caption{%
Behavior of the Lambert $W$ function of the exponent ($V(t) = W(e^t)$).
{\bfseries (a)} Log scale plot with $t \in (-10,0)$.
{\bfseries (b)} Linear scale plot with $t \in (0,10)$.
}\label{fig:lambert}\vspace*{-1em}%
\end{figure}

To develop intuition concerning the Lambert $W$ function of the exponent,
we now briefly discuss how the function $V(t) = W(e^t)$
behaves for different values of $t$.
An illustration is provided in Figure~\ref{fig:lambert}.
One can see directly from the equation
$x + \log x = t$ that the behavior of $x=V(t)$
changes dramatically depending on whether
$t$ is a large positive or a large negative number.
In the first case, the linear part dominates the logarithm
and the function is approximately linear;
a better approximation is $x(t) \approx t - \log t$,
when $t \gg 1$.
In the second case, the function behaves like an exponent $e^t$.
To see this, we write $x = e^t e^{-x}$ and note that
$e^{-x} \approx 1$ when $t \ll 0$,
therefore, $x(t) \approx e^t$, if $t \ll 0$.
We use these approximations as initial points for a $5$-th order
Householder method \cite{householder1970numerical},
which was also used in \cite{Fukushima201377}.
A \emph{single} iteration is already sufficient
to get full \texttt{float} precision
and at most two iterations are needed for \texttt{double}.

\textbf{How to solve (\ref{eq:topk-ent-update}).}
We present a similar derivation as was already done for the
problem (\ref{eq:topk-entropy-primal}) above.
The main difference is that we now encounter
the Lambert $W$ function in the optimality conditions.
We re-write the problem as
\begin{align*}
\min \big\{ &\tfrac{\alpha}{2} ( \inner{x,x} + s^2 )
- \inner{a, x} + \inner{x, \log x} \big. \\
\big. &+ (1-s) \log(1-s) \given s = \inner{\ones, x}, \; x \in \Ska \big\} .
\end{align*}
The Lagrangian is given by
\begin{align*}
\Lc =
& \tfrac{\alpha}{2} ( \inner{x,x} + s^2 )
- \inner{a, x} + \inner{x, \log x} \\
& + (1-s) \log(1-s) + t(\inner{\ones, x} - s) \\
& + \lambda (s-1) - \inner{\mu, x}
+ \inner{\nu, x - \tfrac{s}{k}\ones} ,
\end{align*}
where $t \in \Rb$, $\lambda, \mu, \nu \geq 0$ are the dual variables.
Computing partial derivatives of $\Lc$ \wrt $x_i$ and $s$,
and setting them to zero, we obtain
\begin{align*}
\alpha x_i + \log x_i &= a_i - 1 - t + \mu_i - \nu_i, \quad \forall i, \\
\alpha (1-s) + \log(1-s) &= \alpha - 1 - t
- \lambda - \tfrac{1}{k}\inner{\ones, \nu},
\quad \forall i .
\end{align*}
Note that $x_i > 0$ and $s < 1$ as before, which implies
$\mu_i = 0$ and $\lambda = 0$.
We re-write the above as
\begin{align*}
&\alpha x_i + \log(\alpha x_i) =
a_i - 1 - t + \log \alpha - \nu_i , \\
&\alpha (1-s) + \log\big(\alpha (1-s)\big) =
\alpha - 1 - t + \log \alpha - \tfrac{\inner{\ones, \nu}}{k} .
\end{align*}
Note that these equations correspond to the Lambert $W$ function
of the exponent, \ie $V(t) = W(e^t)$ discussed above.
Let $p \bydef \inner{\ones, \nu}$ and
re-define $t \leftarrow 1 + t - \log \alpha$.
\begin{align*}
\alpha x_i &= W\big( \exp(a_i - t - \nu_i) \big) , \\
\alpha (1-s) &= W\big( \exp(\alpha - t - \tfrac{p}{k}) \big) .
\end{align*}
Finally, we obtain the following system:
\begin{align*}
x_i &= \min\{ \tfrac{1}{\alpha} V(a_i - t) , \tfrac{s}{k} \} ,
\quad \forall i, \\
\alpha x_i &= V(a_i - t - \nu_i) ,
\quad \forall i, \\
\alpha(1 -  s) &= V(\alpha - t - \tfrac{p}{k}) , \\
s &= \inner{\ones, x} , \quad
p = \inner{\ones, \nu} .
\end{align*}
Note that $V(t)$ is a strictly monotonically increasing function,
therefore, it is invertible and we can write
\begin{align*}
a_i -t - \nu_i &= V^{-1}(\alpha x_i) , \\
\alpha - t - \tfrac{p}{k} &= V^{-1}\big(\alpha (1-s) \big) .
\end{align*}
Next, we defined the sets $U$ and $M$ as before and write
\begin{align*}
s &= \inner{\ones, x}
= \tsum_U \tfrac{s}{k} + \tsum_M \tfrac{1}{\alpha} V(a_i - t) , \\
p &= \inner{\ones, \nu}
= \tsum_U a_i - \abs{U}\big( t + V^{-1}(\tfrac{\alpha s}{k}) \big) .
\end{align*}
Let $\rho \bydef \tfrac{\abs{U}}{k}$ as before
and $A \bydef \tfrac{1}{k} \tsum_U a_i$, we get
\begin{align*}
(1-\rho)s &= \tfrac{1}{\alpha} \tsum_M V(a_i - t) , \\
\tfrac{p}{k} &= A - \rho\big( t + V^{-1}(\tfrac{\alpha s}{k}) \big) .
\end{align*}
Finally, we eliminate $p$ and obtain a system in \emph{two} variables,
\begin{align*}
&\alpha (1-\rho)s - \tsum_M V(a_i - t) = 0 , \\
&(1-\rho)t + V^{-1}\big(\alpha (1-s) \big)
- \rho V^{-1}(\tfrac{\alpha s}{k}) + A - \alpha = 0 ,
\end{align*}
which can be solved using the Newton's method \cite{nocedal2006numerical}.
Moreover, when $U$ is empty, the system above simplifies into a single
equation in \emph{one} variable $t$
\begin{align*}
V(\alpha - t) + \tsum_M V(a_i - t) = \alpha ,
\end{align*}
which can be solved efficiently using the Householder's method \cite{householder1970numerical}.
As both methods require derivatives of $V(t)$, we note that
$\partial_t V(t) = V(t) / (1 + V(t))$ \cite{corless1996lambertw}.
Therefore, $V(a_i - t)$ is only computed \emph{once} for each $a_i - t$
and then re-used to also compute the derivatives.

The efficiency of this approach
crucially depends on fast computation of $V(t)$.
Our implementation was able to scale the training procedure
to large datasets as we show next.
\fi

\begin{figure}[ht]\small\centering%
\begin{subfigure}[t]{0.49\columnwidth}\centering
\includegraphics[width=\columnwidth]{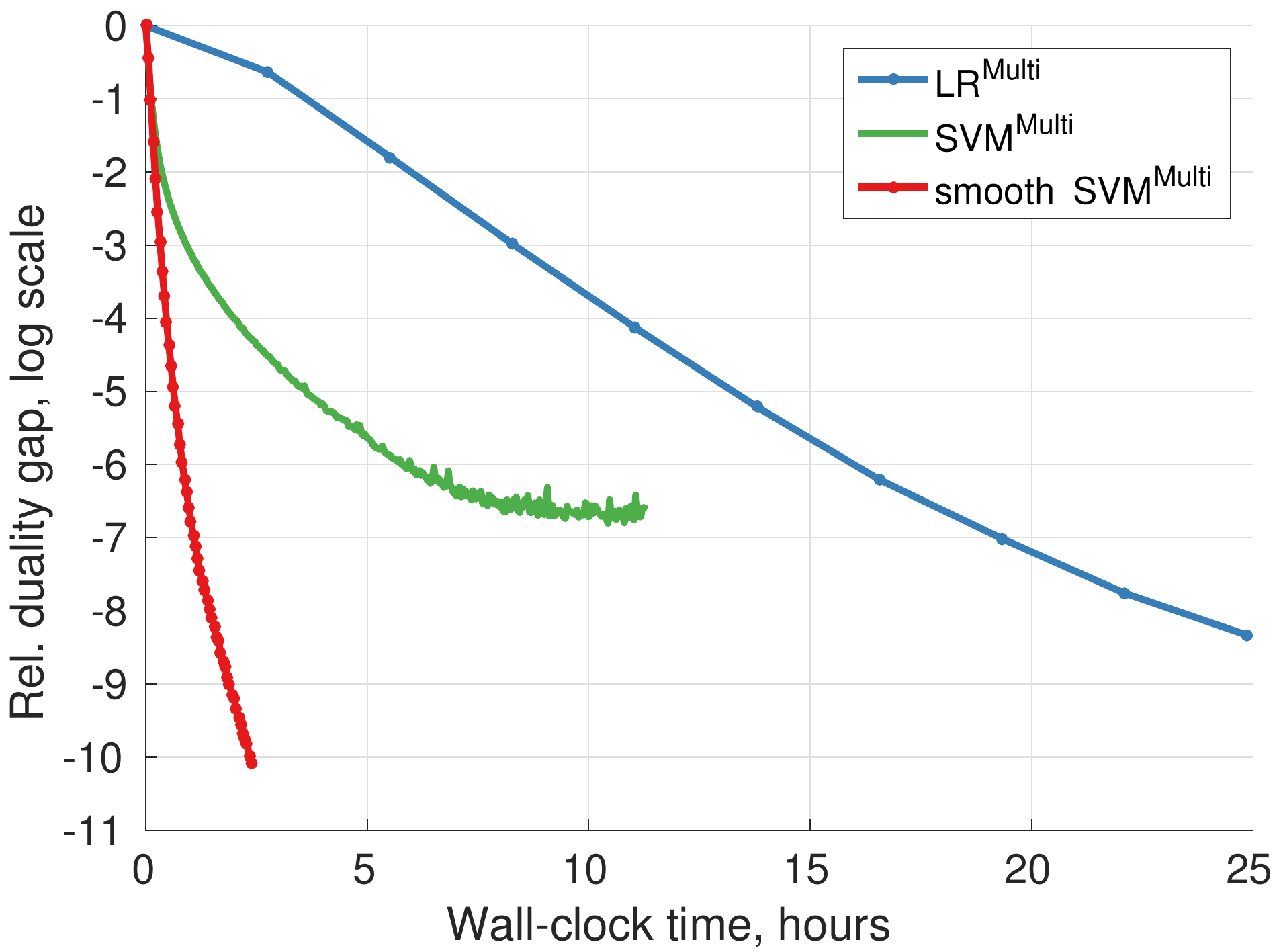}%
\caption{Relative duality gap \V time}\label{fig:time:a}
\end{subfigure}
\begin{subfigure}[t]{0.49\columnwidth}\centering
\includegraphics[width=\columnwidth]{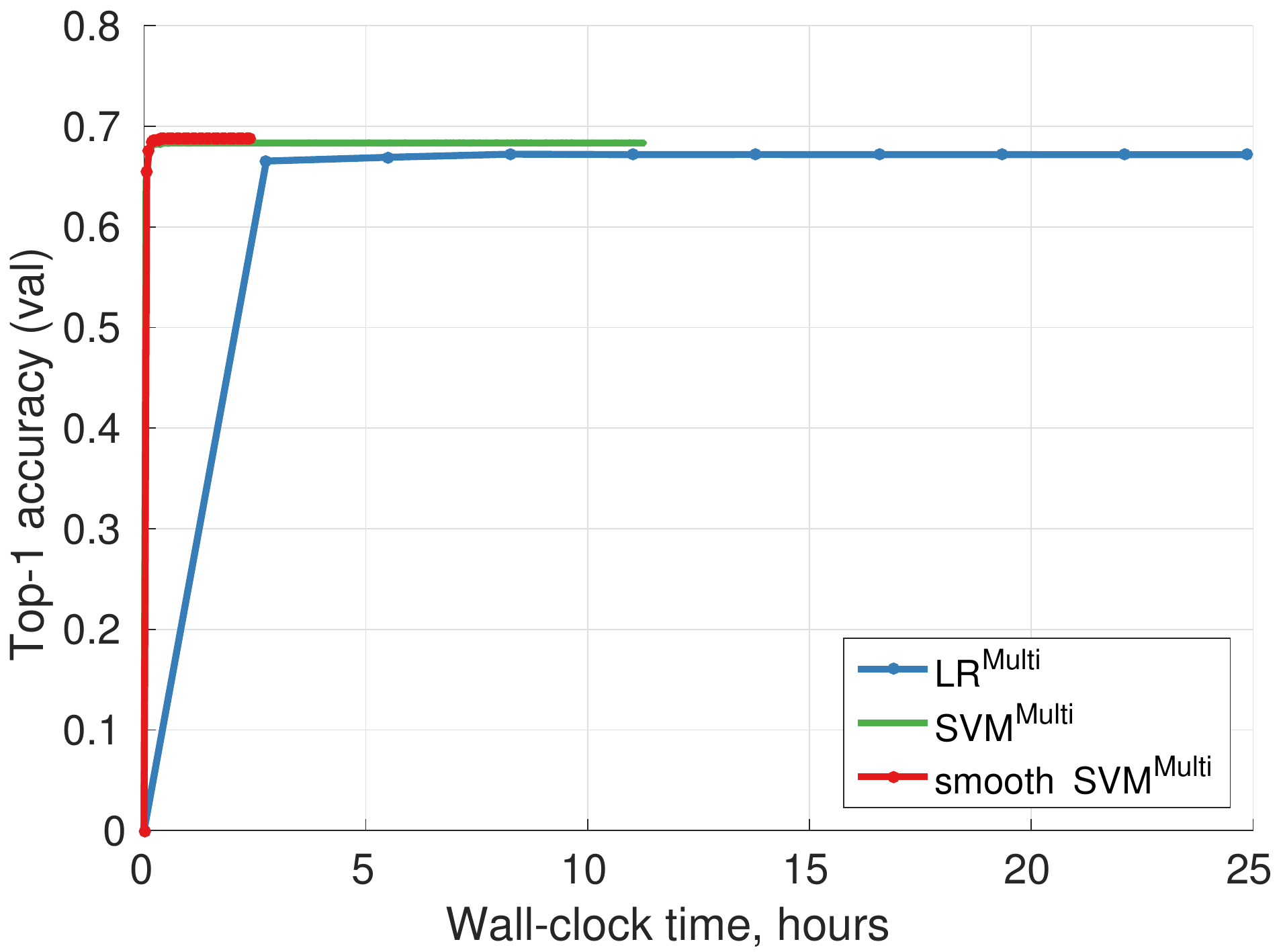}%
\caption{Top-$1$ accuracy \V time}\label{fig:time:b}
\end{subfigure}
\caption{%
SDCA convergence with \LrMulti, \SvmMulti, and \SvmTopKag{1}{1}
objectives on ILSVRC 2012.
}\label{fig:time}
\end{figure}

\iflongversion
\newlength{\oldcolumnsep}
\setlength{\oldcolumnsep}{\columnsep}
\setlength{\columnsep}{1em}
\begin{wrapfigure}[12]{l}{0.48\columnwidth}\centering
\includegraphics[width=0.48\columnwidth]{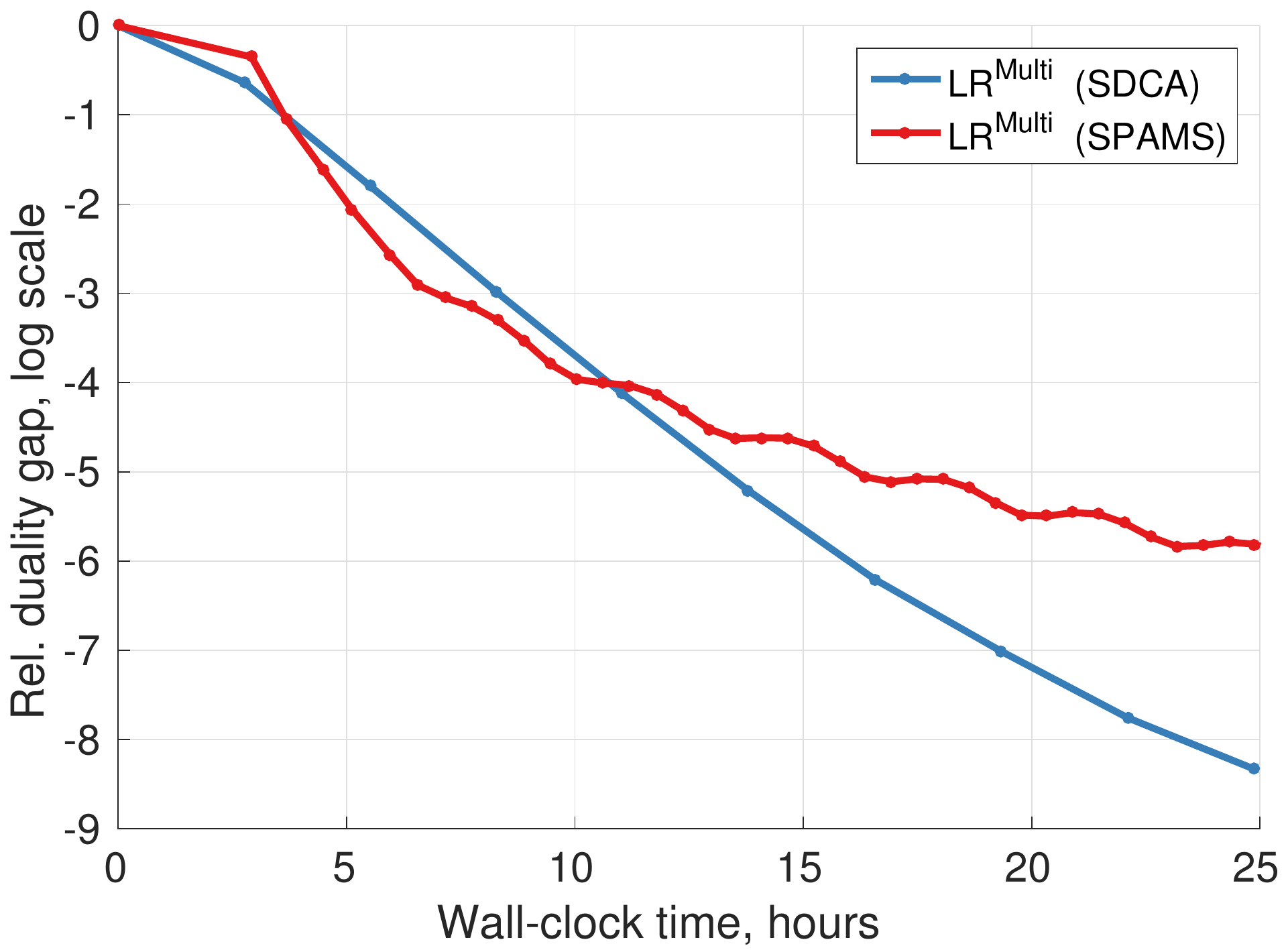}
\caption{%
Convergence rate of SDCA (ours)
and the SPAMS toolbox \cite{mairal2010network}.
}\label{fig:time-spams}%
\end{wrapfigure}

\fi

\textbf{Runtime.}
We compare the wall-clock runtime
of the top-$1$ multiclass SVM \cite{lapin2015topk} (\SvmMulti)
with our smooth multiclass SVM (smooth \SvmMulti)
and the softmax loss (\LrMulti) objectives in Figure~\ref{fig:time}.
We plot the relative duality gap $(P(W) - D(A)) / P(W)$
and the validation accuracy versus time for the best performing models
on ILSVRC 2012.
We obtain substantial improvement
of the convergence rate for smooth top-$1$ SVM compared to the non-smooth
baseline.
Moreover, top-$1$ accuracy saturates after a few passes over the training data,
which justifies the use of a fairly loose stopping criterion
(we used $10^{-3}$).
For \LrMulti, the cost of each epoch is significantly higher
compared to the top-$1$ SVMs, which is due to the difficulty of
solving (\ref{eq:topk-ent-update}).
This suggests that one can use the smooth \SvmTopKag{1}{1}
and obtain competitive performance (see \S~\ref{sec:experiments})
at a lower training cost.

\iflongversion
We also compare our implementation \LrMulti\ (SDCA)
with the SPAMS optimization toolbox \cite{mairal2010network},
denoted \LrMulti\ (SPAMS),
which provides an efficient implementation of FISTA \cite{beck2009fast}.
We note that the rate of convergence of SDCA is competitive
with FISTA for $\epsilon \geq 10^{-4}$
and is noticeably better for $\epsilon < 10^{-4}$.
We conclude that our approach is competitive with the state-of-the-art,
and faster computation of $V(t)$ would lead to a further speedup.
\setlength{\columnsep}{\oldcolumnsep}
\fi

\textbf{Gradient-based optimization.}
Finally, we note that the proposed smooth top-$k$ hinge
and the truncated top-$k$ entropy losses
are easily amenable to gradient-based optimization,
\iflongversion
in particular, for training deep architectures (see \S~\ref{sec:experiments}).
\fi
The computation of the gradient of \eqref{eq:truncated-topk-entropy}
is straightforward,
while for the smooth top-$k$ hinge loss
\eqref{eq:smooth-topk-alpha} we have
\iflongversion
\begin{align*}
\nabla L_{\gamma}(a) &= \tfrac{1}{\gamma} \proj_{\Ska(\gamma)}(a + c),
\end{align*}
which follows from the fact that $L_{\gamma}(a)$
can be written as $\tfrac{1}{2 \gamma}(\norms{x} - \norms{x - p})$
for $x = a+c$ and $p = \proj_{\Ska(\gamma)}(x)$,
and a known result from convex analysis
\cite[\S~3, Ex.~12.d]{borwein2000convex}
which postulates that $\nabla_x \tfrac{1}{2}\norms{x - P_C(x)} = x - P_C(x)$.
\fi

\section{Synthetic Example}
\label{sec:synthetic}
\begin{figure}[t]\small\centering%
\begin{subfigure}[t]{0.49\columnwidth}\centering
\includegraphics[width=\columnwidth]{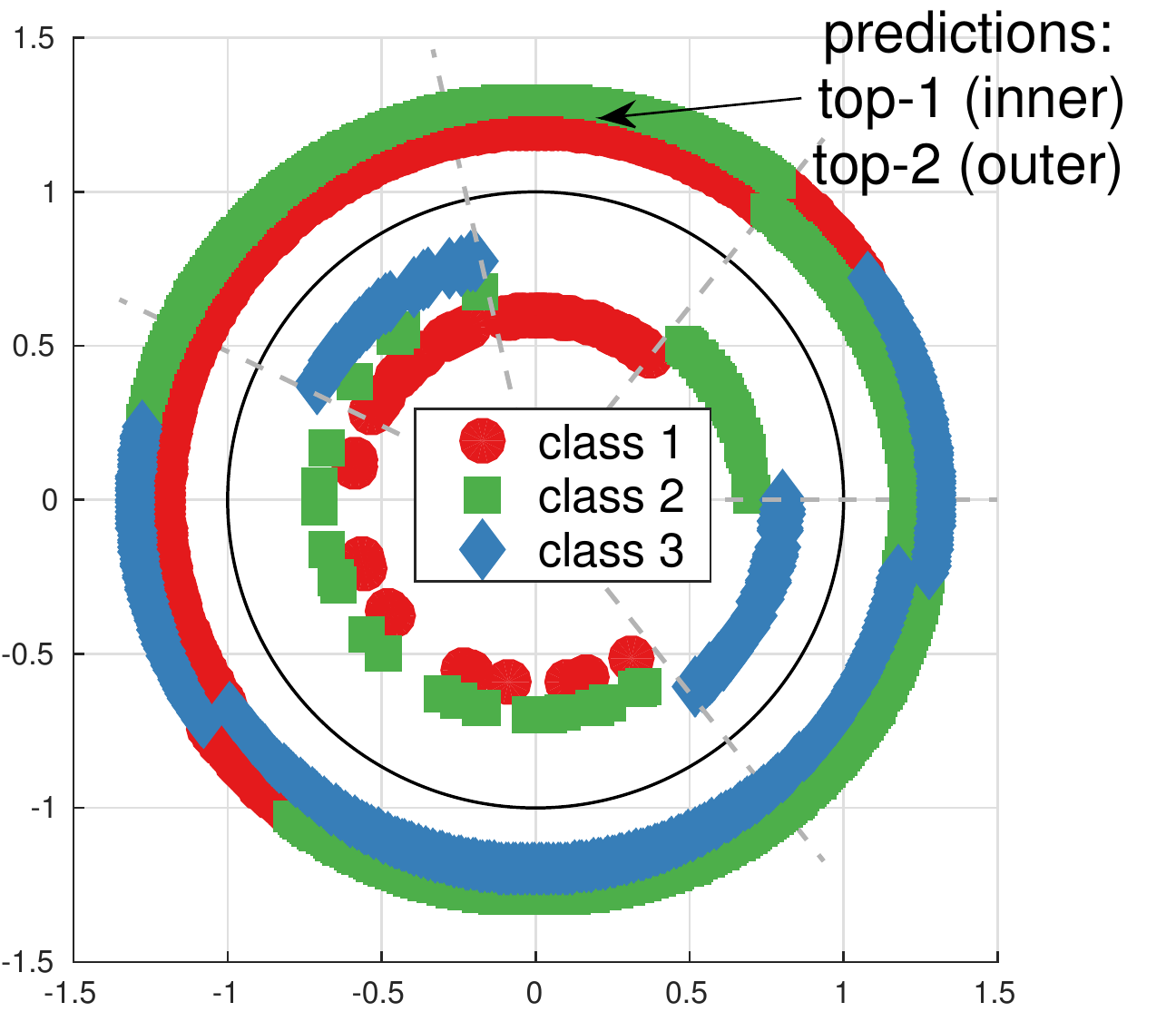}%
\caption{\SvmTopKg{1}{1} test accuracy\\
(top-$1$ / top-$2$): $65.7\%$ / $81.3\%$}\label{fig:toy:a}
\end{subfigure}
\begin{subfigure}[t]{0.49\columnwidth}\centering
\includegraphics[width=\columnwidth]{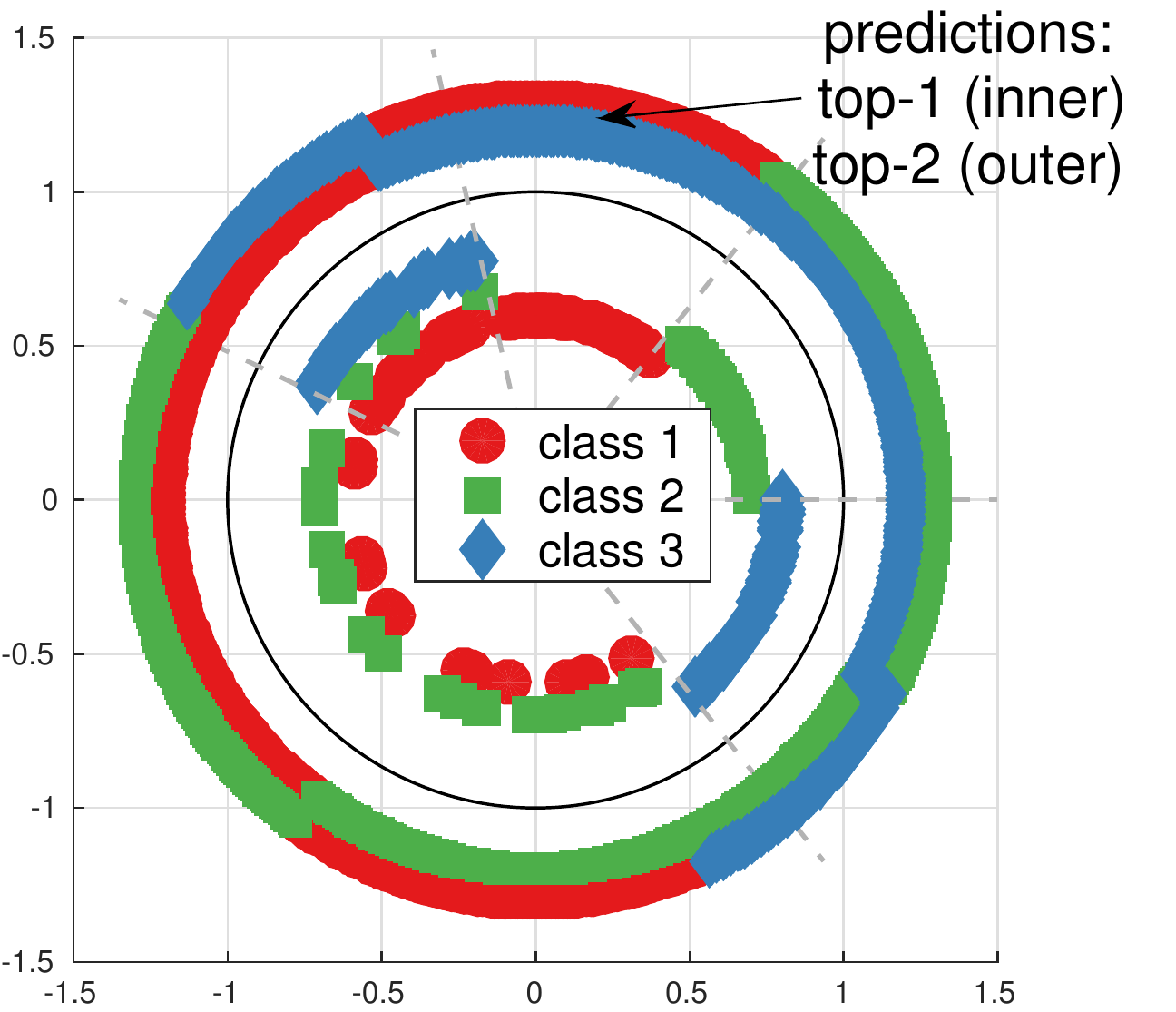}%
\caption{\LrTopKn{2}\ test accuracy\\
(top-$1$ / top-$2$): $29.4\%$, $96.1\%$}\label{fig:toy:b}
\end{subfigure}
\caption{%
Synthetic data on the unit circle in $\Rb^2$ (inside black circle)
and visualization of top-$1$ and top-$2$ predictions (outside black circle).
{\bfseries (a)} Smooth \SvmTopKg{1}{1} optimizes
top-$1$ error which impedes its top-$2$ error.
{\bfseries (b)} Trunc.\ top-$2$ entropy loss
ignores top-$1$ scores and optimizes directly top-$2$ errors
leading to a much better top-$2$ result.
}\label{fig:toy}
\end{figure}

In this section, we demonstrate in a synthetic experiment 
that our proposed top-$2$ losses outperform
the top-$1$ losses when one aims at optimal top-$2$ performance.
The dataset with three classes is shown in the inner circle of
Figure~\ref{fig:toy}.
\iflongversion

\textbf{Sampling.}
First, we generate samples in $[0,7]$ which is subdivided
into $5$ segments. All segments have unit length,
except for the $4$-th segment which has length $3$.
We sample uniformly at random in each of the $5$ segments
according to the following
class-conditional probabilities:
$(0, 1, .4, .3, 0)$ for class $1$,
$(1, 0, .1, .7, 0)$ for class $2$, and
$(0, 0, .5, 0, 1)$ for class $3$.
Finally, the data is rescaled to $[0,1]$ and mapped onto the unit circle.

\begin{table}[ht]\scriptsize\centering\setlength{\tabcolsep}{.4em}
\begin{tabular}{l|cc||l|cc}
\multicolumn{6}{c}{\small\textbf{Circle} (synthetic)} \\\toprule
Method & Top-1 & Top-2 & Method & Top-1 & Top-2\\
\midrule
\midrule
\SvmOva & $54.3$ & $85.8$ &
\SvmTopKg{1}{1} & $\mathbf{65.7}$ & $83.9$ \\
\LrOva & $54.7$ & $81.7$ &
\SvmTopKg{2}{0/1} & $54.4$ / $54.5$ & $87.1$ / $87.0$ \\
\SvmMulti & $58.9$ & $89.3$ &
\LrTopK{2} & $54.6$ & $87.6$ \\
\LrMulti & $54.7$ & $81.7$ &
\LrTopKn{2} & $58.4$ & $\mathbf{96.1}$ \\
\bottomrule
\end{tabular}
\caption{Top-$k$ accuracy (\%) on synthetic data.
{\bfseries Left:} Baselines methods.
{\bfseries Right:} Top-$k$ SVM (nonsmooth / smooth)
and top-$k$ softmax losses (convex and nonconvex).
}
\label{tbl:toy}\vnegvspace
\end{table}

\fi
Samples of different classes are plotted next to each other for better visibility
as there is significant class overlap. We visualize %
top-$1/2$ predictions with two colored circles (outside the black circle).
We sample $200$/$200$/$200\K$ points for
training/validation/testing
and tune the $C = \tfrac{1}{\lambda n}$ parameter in the range
$2^{-18}$ to $2^{18}$.
Results are in Table~\ref{tbl:toy}.

\iflongversion
\begin{table*}[ht]\scriptsize\centering\setlength{\tabcolsep}{.5em}
\begin{tabular}{l|cccc|cccc|cccc|cccc}
\multicolumn{17}{c}{\small\textbf{
\qquad\qquad\qquad\qquad\quad
ALOI \qquad\qquad\qquad\qquad\;
Letter \qquad\qquad\qquad\qquad
News 20 \qquad\qquad
Caltech 101 Silhouettes
}} \\
\toprule
State-of-the-art &
\multicolumn{4}{c|}{$93 \pm 1.2$ \cite{rocha2014multiclass}} &
\multicolumn{4}{c|}{$97.98$ \cite{hsu2002comparison} (RBF kernel)} &
\multicolumn{4}{c|}{$86.9$ \cite{rennie2001improving}} &
$62.1$ & $79.6$ & $83.4$ & \cite{swersky2012probabilistic} \\
\midrule
\midrule
Method &
Top-$1$ & Top-$3$ & Top-$5$ & Top-$10$ &
Top-$1$ & Top-$3$ & Top-$5$ & Top-$10$ &
Top-$1$ & Top-$3$ & Top-$5$ & Top-$10$ &
Top-$1$ & Top-$3$ & Top-$5$ & Top-$10$ \\
\midrule
\midrule
\SvmOva &
$82.4$ & $89.5$ & $91.5$ & $93.7$ &
$63.0$ & $82.0$ & $88.1$ & $94.6$ &
$84.3$ & $95.4$ & $97.9$ & $\mathbf{99.5}$ &
$61.8$ & $76.5$ & $80.8$ & $86.6$ \\
\LrOva &
$86.1$ & $93.0$ & $94.8$ & $96.6$ &
$68.1$ & $86.1$ & $90.6$ & $96.2$ &
$84.9$ & $96.3$ & $97.8$ & $99.3$ &
$63.2$ & $80.4$ & $84.4$ & $89.4$ \\
\midrule
\SvmMulti &
$90.0$ & $95.1$ & $96.7$ & $98.1$ &
$76.5$ & $89.2$ & $93.1$ & $97.7$ &
$85.4$ & $94.9$ & $97.2$ & $99.1$ &
$62.8$ & $77.8$ & $82.0$ & $86.9$ \\
\LrMulti &
$89.8$ & $95.7$ & $97.1$ & $98.4$ &
$75.3$ & $90.3$ & $94.3$ & $98.0$ &
$84.5$ & $96.4$ & $98.1$ & $\mathbf{99.5}$ &
$63.2$ & $\mathbf{81.2}$ & $85.1$ & $89.7$ \\
\midrule
\SvmTopK{3} &
$89.2$ & $95.5$ & $97.2$ & $98.4$ &
$74.0$ & $91.0$ & $94.4$ & $97.8$ &
$85.1$ & $96.6$ & $98.2$ & $99.3$ &
$63.4$ & $79.7$ & $83.6$ & $88.3$ \\
\SvmTopK{5} & $87.3$ & $95.6$ & $97.4$ & $98.6$ &
$70.8$ & $\mathbf{91.5}$ & $95.1$ & $98.4$ &
$84.3$ & $96.7$ & $98.4$ & $99.3$ &
$63.3$ & $80.0$ & $84.3$ & $88.7$ \\
\SvmTopK{10} &
$85.0$ & $95.5$ & $97.3$ & $\mathbf{98.7}$ &
$61.6$ & $88.9$ & $96.0$ & $99.6$ &
$82.7$ & $96.5$ & $98.4$ & $99.3$ &
$63.0$ & $80.5$ & $84.6$ & $89.1$ \\
\midrule
\SvmTopKg{1}{1} &
$\mathbf{90.6}$ & $95.5$ & $96.7$ & $98.2$ &
$\mathbf{76.8}$ & $89.9$ & $93.6$ & $97.6$ &
$\mathbf{85.6}$ & $96.3$ & $98.0$ & $99.3$ &
$\mathbf{63.9}$ & $80.3$ & $84.0$ & $89.0$ \\
\SvmTopKg{3}{1} &
$89.6$ & $95.7$ & $97.3$ & $98.4$ &
$74.1$ & $90.9$ & $94.5$ & $97.9$ &
$85.1$ & $96.6$ & $98.4$ & $99.4$ &
$63.3$ & $80.1$ & $84.0$ & $89.2$ \\
\SvmTopKg{5}{1} &
$87.6$ & $95.7$ & $\mathbf{97.5}$ & $98.6$ &
$70.8$ & $\mathbf{91.5}$ & $95.2$ & $98.6$ &
$84.5$ & $96.7$ & $98.4$ & $99.4$ &
$63.3$ & $80.5$ & $84.5$ & $89.1$ \\
\SvmTopKg{10}{1} &
$85.2$ & $95.6$ & $97.4$ & $\mathbf{98.7}$ &
$61.7$ & $89.1$ & $95.9$ & $\mathbf{99.7}$ &
$82.9$ & $96.5$ & $98.4$ & $\mathbf{99.5}$ &
$63.1$ & $80.5$ & $84.8$ & $89.1$ \\
\midrule
\LrTopK{3} &
$89.0$ & $95.8$ & $97.2$ & $98.4$ &
$73.0$ & $90.8$ & $94.9$ & $98.5$ &
$84.7$ & $96.6$ & $98.3$ & $99.4$ &
$63.3$ & $81.1$ & $85.0$ & $89.9$ \\
\LrTopK{5} &
$87.9$ & $95.8$ & $97.2$ & $98.4$ &
$69.7$ & $90.9$ & $95.1$ & $98.8$ &
$84.3$ & $\mathbf{96.8}$ & $\mathbf{98.6}$ & $99.4$ &
$63.2$ & $80.9$ & $85.2$ & $89.9$ \\
\LrTopK{10} &
$86.0$ & $95.6$ & $97.3$ & $98.5$ &
$65.0$ & $89.7$ & $\mathbf{96.2}$ & $99.6$ &
$82.7$ & $96.4$ & $98.5$ & $99.4$ &
$62.5$ & $80.8$ & $\mathbf{85.4}$ & $90.1$ \\
\midrule
\LrTopKn{3} &
$89.3$ & $\mathbf{95.9}$ & $97.3$ & $98.5$ &
$63.6$ & $91.1$ & $95.6$ & $98.8$ &
$83.4$ & $96.4$ & $98.3$ & $99.4$ &
$60.7$ & $81.1$ & $85.2$ & $\mathbf{90.2}$ \\
\LrTopKn{5} &
$87.9$ & $95.7$ & $97.3$ & $98.6$ &
$50.3$ & $87.7$ & $96.1$ & $99.4$ &
$83.2$ & $96.0$ & $98.2$ & $99.4$ &
$58.3$ & $79.8$ & $85.2$ & $\mathbf{90.2}$ \\
\LrTopKn{10} &
$85.2$ & $94.8$ & $97.1$ & $98.5$ &
$46.5$ & $80.9$ & $93.7$ & $99.6$ &
$82.9$ & $95.7$ & $97.9$ & $99.4$ &
$51.9$ & $78.4$ & $84.6$ & $\mathbf{90.2}$ \\
\bottomrule
\end{tabular}
\\[.4em]\setlength{\tabcolsep}{.55em}
\begin{tabular}{l|cccc|cccc|cccc|ccc}
\multicolumn{16}{c}{\small\textbf{
\qquad\qquad\qquad\quad
Indoor 67 \qquad\qquad\qquad\quad
CUB \qquad\qquad\qquad\qquad\;\;
Flowers \qquad\qquad\qquad\quad
FMD
}} \\
\toprule
State-of-the-art &
\multicolumn{4}{c|}{$82.0$ \cite{wang2015places}} &
\multicolumn{4}{c|}{$62.8$ \cite{cimpoi15deep} / $76.37$ \cite{zhang2014part}} &
\multicolumn{4}{c|}{$86.8$ \cite{razavian2014cnn}} &
\multicolumn{3}{c}{$77.4$ \cite{cimpoi15deep} / $82.4$ \cite{cimpoi15deep}} \\
\midrule
\midrule
Method &
Top-$1$ & Top-$3$ & Top-$5$ & Top-$10$ &
Top-$1$ & Top-$3$ & Top-$5$ & Top-$10$ &
Top-$1$ & Top-$3$ & Top-$5$ & Top-$10$ &
Top-$1$ & Top-$3$ & Top-$5$ \\
\midrule
\midrule
\SvmOva &
$81.9$ & $94.3$ & $96.5$ & $98.0$ &
$60.6$ & $77.1$ & $83.4$ & $89.9$ &
$82.0$ & $91.7$ & $94.3$ & $96.8$ &
$77.4$ & $92.4$ & $96.4$ \\
\LrOva &
$82.0$ & $94.9$ & $97.2$ & $98.7$ &
$62.3$ & $80.5$ & $87.4$ & $93.5$ &
$82.6$ & $92.2$ & $94.8$ & $97.6$ &
$79.6$ & $94.2$ & $\mathbf{98.2}$ \\
\midrule
\SvmMulti &
$82.5$ & $\mathbf{95.4}$ & $97.3$ & $99.1$ &
$61.0$ & $79.2$ & $85.7$ & $92.3$ &
$82.5$ & $92.2$ & $94.8$ & $96.4$ &
$77.6$ & $93.8$ & $97.2$ \\
\LrMulti &
$82.4$ & $95.2$ & $\mathbf{98.0}$ & $99.1$ &
$62.3$ & $81.7$ & $87.9$ & $\mathbf{93.9}$ &
$82.9$ & $92.4$ & $95.1$ & $97.8$ &
$79.0$ & $94.6$ & $97.8$ \\
\midrule
\SvmTopK{3} &
$81.6$ & $95.1$ & $97.7$ & $99.0$ &
$61.3$ & $80.4$ & $86.3$ & $92.5$ &
$81.9$ & $92.2$ & $95.0$ & $96.1$ &
$78.8$ & $94.6$ & $97.8$ \\
\SvmTopK{5} &
$79.9$ & $95.0$ & $97.7$ & $99.0$ &
$60.9$ & $81.2$ & $87.2$ & $92.9$ &
$81.7$ & $92.4$ & $95.1$ & $97.8$ &
$78.4$ & $94.4$ & $97.6$ \\
\SvmTopK{10} &
$78.4$ & $95.1$ & $97.4$ & $99.0$ &
$59.6$ & $81.3$ & $87.7$ & $93.4$ &
$80.5$ & $91.9$ & $95.1$ & $97.7$ \\
\midrule
\SvmTopKg{1}{1} &
$\mathbf{82.6}$ & $95.2$ & $97.6$ & $99.0$ &
$61.9$ & $80.2$ & $86.9$ & $93.1$ &
$\mathbf{83.0}$ & $92.4$ & $95.1$ & $97.6$ &
$78.6$ & $93.8$ & $98.0$ \\
\SvmTopKg{3}{1} &
$81.6$ & $95.1$ & $97.8$ & $99.0$ &
$61.9$ & $81.1$ & $86.6$ & $93.2$ &
$82.5$ & $92.3$ & $95.2$ & $97.7$ &
$79.0$ & $94.4$ & $98.0$ \\
\SvmTopKg{5}{1} &
$80.4$ & $95.1$ & $97.8$ & $99.1$ &
$61.3$ & $81.3$ & $87.4$ & $92.9$ &
$82.0$ & $\mathbf{92.5}$ & $95.1$ & $97.8$ &
$79.4$ & $94.4$ & $97.6$ \\
\SvmTopKg{10}{1} &
$78.3$ & $95.1$ & $97.5$ & $99.0$ &
$59.8$ & $81.4$ & $87.8$ & $93.4$ &
$80.6$ & $91.9$ & $95.1$ & $97.7$ \\
\midrule
\LrTopK{3} &
$81.4$ & $\mathbf{95.4}$ & $97.6$ & $\mathbf{99.2}$ &
$\mathbf{62.5}$ & $81.8$ & $87.9$ & $\mathbf{93.9}$ &
$82.5$ & $92.0$ & $\mathbf{95.3}$ & $97.8$ &
$\mathbf{79.8}$ & $94.8$ & $98.0$ \\
\LrTopK{5} &
$80.3$ & $95.0$ & $97.7$ & $99.0$ &
$62.0$ & $\mathbf{81.9}$ & $88.1$ & $93.8$ &
$82.1$ & $92.2$ & $95.1$ & $\mathbf{97.9}$ &
$79.4$ & $94.4$ & $98.0$ \\
\LrTopK{10} &
$79.2$ & $95.1$ & $97.6$ & $99.0$ &
$61.2$ & $81.6$ & $\mathbf{88.2}$ & $93.8$ &
$80.9$ & $92.1$ & $95.0$ & $97.7$ \\
\midrule
\LrTopKn{3} &
$79.8$ & $95.0$ & $97.5$ & $99.1$ &
$62.0$ & $81.4$ & $87.6$ & $93.4$ &
$82.1$ & $92.2$ & $95.2$ & $97.6$ &
$78.4$ & $\mathbf{95.4}$ & $\mathbf{98.2}$ \\
\LrTopKn{5} &
$76.4$ & $94.3$ & $97.3$ & $99.0$ &
$61.4$ & $81.2$ & $87.7$ & $93.7$ &
$81.4$ & $92.0$ & $95.0$ & $97.7$ &
$77.2$ & $94.0$ & $97.8$ \\
\LrTopKn{10} &
$72.6$ & $92.8$ & $97.1$ & $98.9$ &
$59.7$ & $80.7$ & $87.2$ & $93.4$ &
$77.9$ & $91.1$ & $94.3$ & $97.3$ \\
\bottomrule
\end{tabular}
\\[.5em]\setlength{\tabcolsep}{.7em}
\begin{tabular}{l|cccc|cccc|cccc}
\multicolumn{13}{c}{\small{
\qquad\qquad\qquad\qquad\qquad
\textbf{SUN 397} (10 splits) \qquad\qquad\qquad\qquad\quad
\textbf{Places 205} (val) \qquad\qquad
\textbf{ILSVRC 2012} (val)
}} \\
\toprule
State-of-the-art &
\multicolumn{4}{c|}{$66.9$ \cite{wang2015places}} &
$60.6$ & & $88.5$ & \cite{wang2015places} &
$76.3$ & & $93.2$ & \cite{simonyan14c} \\
\midrule
\midrule
Method &
Top-$1$ & Top-$3$ & Top-$5$ & Top-$10$ &
Top-$1$ & Top-$3$ & Top-$5$ & Top-$10$ &
Top-$1$ & Top-$3$ & Top-$5$ & Top-$10$ \\
\midrule
\midrule
\SvmMulti &
$65.8 \pm 0.1$ & $85.1 \pm 0.2$ & $90.8 \pm 0.1$ & $95.3 \pm 0.1$ &
$58.4$ & $78.7$ & $84.7$ & $89.9$ &
$68.3$ & $82.9$ & $87.0$ & $91.1$ \\
\LrMulti &
$\mathbf{67.5 \pm 0.1}$ & $\mathbf{87.7 \pm 0.2}$ & $\mathbf{92.9 \pm 0.1}$ & $\mathbf{96.8 \pm 0.1}$ &
$59.0$ & $\mathbf{80.6}$ & $\mathbf{87.6}$ & $\mathbf{94.3}$ &
$67.2$ & $83.2$ & $87.7$ & $92.2$ \\
\midrule
\SvmTopK{3} &
$66.5 \pm 0.2$ & $86.5 \pm 0.1$ & $91.8 \pm 0.1$ & $95.9 \pm 0.1$ &
$58.6$ & $80.3$ & $87.3$ & $93.3$ &
$68.2$ & $84.0$ & $88.1$ & $92.1$ \\
\SvmTopK{5} &
$66.3 \pm 0.2$ & $87.0 \pm 0.2$ & $92.2 \pm 0.2$ & $96.3 \pm 0.1$ &
$58.4$ & $80.5$ & $87.4$ & $94.0$ &
$67.8$ & $\mathbf{84.1}$ & $88.2$ & $92.4$ \\
\SvmTopK{10} &
$64.8 \pm 0.3$ & $87.2 \pm 0.2$ & $92.6 \pm 0.1$ & $96.6 \pm 0.1$ &
$58.0$ & $80.4$ & $87.4$ & $\mathbf{94.3}$ &
$67.0$ & $83.8$ & $88.3$ & $\mathbf{92.6}$ \\
\midrule
\SvmTopKg{1}{1} &
$67.4 \pm 0.2$ & $86.8 \pm 0.1$ & $92.0 \pm 0.1$ & $96.1 \pm 0.1$ &
$\mathbf{59.2}$ & $80.5$ & $87.3$ & $93.8$ &
$\mathbf{68.7}$ & $83.9$ & $88.0$ & $92.1$ \\
\SvmTopKg{3}{1} &
$67.0 \pm 0.2$ & $87.0 \pm 0.1$ & $92.2 \pm 0.1$ & $96.2 \pm 0.0$ &
$58.9$ & $80.5$ & $\mathbf{87.6}$ & $93.9$ &
$68.2$ & $\mathbf{84.1}$ & $88.2$ & $92.3$ \\
\SvmTopKg{5}{1} &
$66.5 \pm 0.2$ & $87.2 \pm 0.1$ & $92.4 \pm 0.2$ & $96.3 \pm 0.0$ &
$58.5$ & $80.5$ & $87.5$ & $94.1$ &
$67.9$ & $\mathbf{84.1}$ & $\mathbf{88.4}$ & $92.5$ \\
\SvmTopKg{10}{1} &
$64.9 \pm 0.3$ & $87.3 \pm 0.2$ & $92.6 \pm 0.2$ & $96.6 \pm 0.1$ &
$58.0$ & $80.4$ & $87.5$ & $\mathbf{94.3}$ &
$67.1$ & $83.8$ & $88.3$ & $\mathbf{92.6}$ \\
\midrule
\LrTopK{3} &
$67.2 \pm 0.2$ & $\mathbf{87.7 \pm 0.2}$ & $\mathbf{92.9 \pm 0.1}$ & $\mathbf{96.8 \pm 0.1}$ &
$58.7$ & $\mathbf{80.6}$ & $\mathbf{87.6}$ & $94.2$ &
$66.8$ & $83.1$ & $87.8$ & $92.2$ \\
\LrTopK{5} &
$66.6 \pm 0.3$ & $\mathbf{87.7 \pm 0.2}$ & $\mathbf{92.9 \pm 0.1}$ & $\mathbf{96.8 \pm 0.1}$ &
$58.1$ & $80.4$ & $87.4$ & $94.2$ &
$66.5$ & $83.0$ & $87.7$ & $92.2$ \\
\LrTopK{10} &
$65.2 \pm 0.3$ & $87.4 \pm 0.1$ & $92.8 \pm 0.1$ & $\mathbf{96.8 \pm 0.1}$ &
$57.0$ & $80.0$ & $87.2$ & $94.1$ &
$65.8$ & $82.8$ & $87.6$ & $92.1$ \\
\bottomrule
\end{tabular}
\caption{Top-$k$ accuracy (\%) on various datasets.
The first line is a reference to the state-of-the-art on each dataset
and reports top-$1$ accuracy except when the numbers are aligned with Top-$k$.
We compare the one-vs-all
and multiclass baselines
with the \SvmTopKa{k} \cite{lapin2015topk}
as well as the proposed
smooth \SvmTopKag{k}{\gamma}, \LrTopK{k}, and the nonconvex
\LrTopKn{k}.
}
\label{tbl:all}
\end{table*}

\fi

In each column we provide the results for the model that optimizes
the corresponding top-$k$ accuracy, which is in general different
for top-$1$ and top-$2$.
First, we note that all top-$1$ baselines perform similar in top-$1$
performance, except for \SvmMulti\ and \SvmTopKg{1}{1}
which show better results.
Next, we see that our top-$2$ losses improve
the top-$2$ accuracy and the improvement is most significant
for the nonconvex \LrTopKn{2} loss,
which is close to the optimal solution for this dataset.
This is because \LrTopKn{2} is a tight bound on the top-$2$ error
and ignores top-$1$ errors in the loss.
Unfortunately, similar significant improvements were not observed
on the real-world data sets that we tried.

\iflongversion\else

\fi

\section{Experimental Results}
\label{sec:experiments}
The goal of this section is to provide an extensive empirical evaluation
of the top-$k$ performance of different losses in multiclass classification.
To this end, we evaluate the loss functions introduced in \S~\ref{sec:multiclass}
on $11$ datasets ($500$ to $2.4\M$ training examples, $10$ to $1000$ classes),
from various problem domains (vision and non-vision;
fine-grained, scene and general object classification).
The detailed statistics of the datasets is given in Table~\ref{tbl:stats}.

\begin{table}[ht]\scriptsize\centering\setlength{\tabcolsep}{.4em}
\begin{tabular}{l|ccc||l|ccc}
\toprule
Dataset & $m$ & $n$ & $d$ & Dataset & $m$ & $n$ & $d$ \\
\midrule
\midrule
ALOI \cite{rocha2014multiclass} &
$1\K$ & $54\K$ & $128$ &
Indoor 67 \cite{quattoni2009recognizing} &
$67$ & $5354$ & $4\K$ \\
Caltech 101 Sil \cite{swersky2012probabilistic} &
$101$ & $4100$ & $784$ &
Letter \cite{hsu2002comparison} &
$26$ & $10.5\K$ & $16$ \\
CUB \cite{wah2011caltech} &
$202$ & $5994$ & $4\K$ &
News 20 \cite{lang1995newsweeder} &
$20$ & $15.9\K$ & $16\K$ \\
Flowers \cite{nilsback2008automated} &
$102$ & $2040$ & $4\K$ &
Places 205 \cite{zhou2014learning} &
$205$ & $2.4\M$ & $4\K$ \\
FMD \cite{sharan2009material} &
$10$ & $500$ & $4\K$ &
SUN 397 \cite{xiao2010sun} &
$397$ & $19.9\K$ & $4\K$ \\
ILSVRC 2012 \cite{ILSVRCarxiv14} &
$1\K$ & $1.3\M$ & $4\K$ &
& & & \\
\bottomrule
\end{tabular}
\caption{Statistics of the datasets used in the experiments
($m$ -- \# classes, $n$ -- \# training examples, $d$ -- \# features).
}
\label{tbl:stats}
\end{table}

Please refer to Table~\ref{tbl:summary} for an overview of the methods
and our naming convention.
\iflongversion
A broad selection of results is also reported at the end of the paper.
\fi
As other ranking based losses did not perform well in \cite{lapin2015topk},
we do no further comparison here.

\textbf{Solvers.}
We use \texttt{LibLinear} \cite{REF08a} for the one-vs-all baselines
\SvmOva\ and \LrOva; and our code from \cite{lapin2015topk}
for \SvmTopK{k}.
We extended the latter to support the smooth \SvmTopKg{k}{\gamma} and \LrTopK{k}.
The multiclass loss baselines \SvmMulti\ and \LrMulti\
correspond respectively to \SvmTopK{1} and \LrTopK{1}.
For the nonconvex \LrTopKn{k},
we use the \LrMulti\ solution as an initial point
and perform gradient descent with line search.
We cross-validate hyper-parameters in the range
$10^{-5}$ to $10^3$, extending it when the optimal value is
at the boundary.

\textbf{Features.}
For ALOI, Letter, and News20 datasets,
we use the features provided by the \texttt{LibSVM} \cite{chang2011libsvm}
datasets.
For ALOI, we randomly split the data into equally sized 
training and test sets preserving class distributions.
The Letter dataset comes with a separate validation set,
which we used for model selection only.
For News20, we use PCA to reduce dimensionality
of sparse features
from $62060$ to $15478$
preserving all non-singular PCA components\footnote{
The \SvmTopK{k} solvers that we used were designed for dense inputs.}.

For Caltech101 Silhouettes, we use the features and the
train/val/test splits provided by \cite{swersky2012probabilistic}.

For CUB, Flowers, FMD, and ILSVRC 2012,
we use \texttt{MatConvNet}
\cite{vedaldi15matconvnet} to extract
the outputs of the last fully connected layer of
the \texttt{imagenet-vgg-verydeep-16} model
which is pre-trained on ImageNet \cite{deng2009imagenet}
and achieves state-of-the-art results
in image classification \cite{simonyan14c}.

For Indoor 67, SUN 397, and Places 205,
we use the \texttt{Places205-VGGNet-16} model by \cite{wang2015places}
which is pre-trained on Places 205 \cite{zhou2014learning}
and outperforms the ImageNet pre-trained model
on scene classification tasks \cite{wang2015places}.
\iflongversion
Further results can be found at the end of the paper.
\fi
In all cases we obtain a similar behavior 
in terms of the ranking of the considered losses as discussed below.

\textbf{Discussion.}
The experimental results are given in Table~\ref{tbl:all}.
There are several interesting observations that one can make.
While the OVA schemes perform quite similar to 
the multiclass approaches
(logistic OVA \V softmax,
hinge OVA \V multiclass SVM),
which confirms earlier observations in \cite{akata2014good,rifkin2004defense},
the OVA schemes performed worse on ALOI and Letter.
Therefore it seems safe to recommend to use multiclass losses instead of the OVA schemes.

Comparing the softmax \V multiclass SVM losses,
we see that there is no clear winner in top-$1$ performance,
but softmax consistently outperforms multiclass SVM
in top-$k$ performance for $k>1$.
This might be due
to the strong property of softmax being top-$k$ calibrated for all $k$.
Please note that this
trend is uniform across all datasets,
in particular, also for the ones where the features are not coming from a convnet.
Both the smooth top-$k$ hinge and the top-$k$ entropy losses 
perform slightly better than softmax if one
compares the corresponding top-$k$ errors.
However, the good performance of the truncated top-$k$ loss
on synthetic data does not transfer to the real world datasets.
This might be due to a relatively high dimension of the feature spaces,
but requires further investigation.

\iflongversion
\begin{table}[t]\scriptsize\centering\setlength{\tabcolsep}{.75em}
\begin{tabular}{l|cccc}
\multicolumn{5}{c}{\small\textbf{Places 205} (val)} \\
\toprule
Method &
Top-$1$ & Top-$3$ & Top-$5$ & Top-$10$ \\
\midrule
\midrule
\LrMulti &
$59.97$ & $81.39$ & $88.17$ & $94.59$ \\
\midrule
\SvmTopKg{3}{1}\ (FT) &
$60.73$ & $82.09$ & $88.58$ & $94.56$ \\
\SvmTopKg{5}{1}\ (FT) &
$\mathbf{60.88}$ & $\mathbf{82.18}$ & $\mathbf{88.78}$ & $94.75$ \\
\midrule
\LrTopKn{3}\ (FT) &
$60.51$ & $81.86$ & $88.69$ & $94.78$ \\
\LrTopKn{5}\ (FT) &
$60.48$ & $81.66$ & $88.66$ & $94.80$ \\
\midrule
\LrMulti\ (FT) &
$60.73$ & $82.07$ & $88.71$ & $\mathbf{94.82}$ \\
\bottomrule
\end{tabular}

\vspace{1em}
\begin{tabular}{l|cccc}
\multicolumn{5}{c}{\small\textbf{ILSVRC 2012} (val)} \\
\toprule
Method &
Top-$1$ & Top-$3$ & Top-$5$ & Top-$10$ \\
\midrule
\midrule
\LrMulti &
$68.60$ & $84.29$ & $88.66$ & $92.83$ \\
\midrule
\SvmTopKg{3}{1}\ (FT) &
$71.66$ & $86.63$ & $90.55$ & $94.17$ \\
\SvmTopKg{5}{1}\ (FT) &
$71.60$ & $86.67$ & $90.56$ & $94.23$ \\
\midrule
\LrTopKn{3}\ (FT) &
$71.41$ & $86.80$ & $90.77$ & $94.35$ \\
\LrTopKn{5}\ (FT) &
$71.20$ & $86.57$ & $90.75$ & $\mathbf{94.38}$ \\
\midrule
\LrMulti\ (FT) &
$\mathbf{72.11}$ & $\mathbf{87.08}$ & $\mathbf{90.88}$ & $\mathbf{94.38}$ \\
\bottomrule
\end{tabular}
\caption{Top-$k$ accuracy (\%), as reported by Caffe \cite{jia2014caffe},
on large scale datasets after fine-tuning (FT)
for approximately one epoch on Places and 3 epochs on ILSVRC.
The first line (\LrMulti) is the reference performance w/o fine-tuning.
}
\label{tbl:finetune}
\end{table}

\fi

\iflongversion
\textbf{Fine-tuning experiments.}
We also performed a number of fine-tuning experiments
where the original network was trained further for $1$-$3$ epochs
with the smooth top-$k$ hinge and
the truncated top-$k$ entropy losses\footnote{
Code: \url{https://github.com/mlapin/caffe/tree/topk}}.
The motivation was to see if the full end-to-end training would be more
beneficial compared to training just the classifier.
Results are reported in Table~\ref{tbl:finetune}.
First, we note that the setting is now slightly different:
there is no feature extraction step with the \texttt{MatConvNet}
and there is a non-regularized bias term in Caffe \cite{jia2014caffe}.
Next, we see that the top-$k$ specific losses are able to improve
the performance compared to the reference model,
and that the \SvmTopKg{5}{1}\ loss achieves the best top-$1..5$ performance
on Places 205.
However, in this set of experiments,
we also observed similar improvements
when fine-tuning with the standard softmax loss,
which achieves best performance on ILSVRC 2012.
\fi

We conclude that a safe choice for multiclass problems
seems to be the softmax loss
as it yields competitive results in all top-$k$ errors.
An interesting alternative is the smooth top-$k$ hinge loss
which is faster to train (see Section~\ref{sec:optimization})
and achieves competitive performance.
If one wants to optimize directly for a top-$k$ error
(at the cost of a higher top-$1$ error),
then further improvements are possible using either
the smooth top-$k$ SVM or the top-$k$ entropy losses.

\section{Conclusion}
\label{sec:conclusion}
We have done an extensive experimental study of
top-$k$ performance optimization.
We observed that the softmax loss and the smooth top-$1$ hinge loss
are competitive across all top-$k$ errors
and should be considered the primary candidates in practice.
Our new top-$k$ loss functions
can further improve these results slightly, especially
if one is targeting a particular top-$k$ error as the performance measure.
Finally, we would like to highlight our new optimization scheme based on SDCA
for the top-$k$ entropy loss
which also includes the softmax loss and is of an independent interest.

{\small
\bibliographystyle{ieee}
\bibliography{main}

\begin{thebibliography}{10}\itemsep=-1pt

\bibitem{agarwal2011infinite}
S.~Agarwal.
\newblock The infinite push: A new support vector ranking algorithm that
  directly optimizes accuracy at the absolute top of the list.
\newblock In {\em SDM}, pages 839--850, 2011.

\bibitem{akata2014good}
Z.~Akata, F.~Perronnin, Z.~Harchaoui, and C.~Schmid.
\newblock Good practice in large-scale learning for image classification.
\newblock {\em PAMI}, 36(3):507--520, 2014.

\bibitem{BarJorAuc2006}
P.~L. Bartlett, M.~I. Jordan, and J.~D. McAuliffe.
\newblock Convexity, classification and risk bounds.
\newblock {\em Journal of the American Statistical Association}, 101:138--156,
  2006.

\bibitem{beck2009fast}
A.~Beck and M.~Teboulle.
\newblock A fast iterative shrinkage-thresholding algorithm for linear inverse
  problems.
\newblock {\em SIAM Journal on Imaging Sciences}, 2(1):183--202, 2009.

\bibitem{BecTeb2012}
A.~Beck and M.~Teboulle.
\newblock Smoothing and first order methods, a unified framework.
\newblock {\em SIAM Journal on Optimization}, 22:557--580, 2012.

\bibitem{bengio2009learning}
Y.~Bengio.
\newblock Learning deep architectures for {AI}.
\newblock {\em Foundations and Trends in Machine Learning}, 2(1):1--127, 2009.

\bibitem{bordes2007solving}
A.~Bordes, L.~Bottou, P.~Gallinari, and J.~Weston.
\newblock Solving multiclass support vector machines with {LaRank}.
\newblock In {\em ICML}, pages 89--96, 2007.

\bibitem{borwein2000convex}
J.~M. Borwein and A.~S. Lewis.
\newblock {\em Convex Analysis and Nonlinear Optimization: Theory and
  Examples}.
\newblock Cms Books in Mathematics Series. Springer Verlag, 2000.

\bibitem{boyd2012accuracy}
S.~Boyd, C.~Cortes, M.~Mohri, and A.~Radovanovic.
\newblock Accuracy at the top.
\newblock In {\em NIPS}, pages 953--961, 2012.

\bibitem{boyd2004convex}
S.~Boyd and L.~Vandenberghe.
\newblock {\em Convex Optimization}.
\newblock Cambridge University Press, 2004.

\bibitem{burges2005learning}
C.~Burges, T.~Shaked, E.~Renshaw, A.~Lazier, M.~Deeds, N.~Hamilton, and
  G.~Hullender.
\newblock Learning to rank using gradient descent.
\newblock In {\em ICML}, pages 89--96, 2005.

\bibitem{chang2011libsvm}
C.-C. Chang and C.-J. Lin.
\newblock {LIBSVM}: A library for support vector machines.
\newblock {\em ACM Transactions on Intelligent Systems and Technology},
  2:1--27, 2011.

\bibitem{cimpoi15deep}
M.~Cimpoi, S.~Maji, and A.~Vedaldi.
\newblock Deep filter banks for texture recognition and segmentation.
\newblock In {\em CVPR}, 2015.

\bibitem{corless1996lambertw}
R.~M. Corless, G.~H. Gonnet, D.~E. Hare, D.~J. Jeffrey, and D.~E. Knuth.
\newblock On the lambert {W} function.
\newblock {\em Advances in Computational Mathematics}, 5(1):329--359, 1996.

\bibitem{crammer2001algorithmic}
K.~Crammer and Y.~Singer.
\newblock On the algorithmic implementation of multiclass kernel-based vector
  machines.
\newblock {\em JMLR}, 2:265--292, 2001.

\bibitem{deng2009imagenet}
J.~Deng, W.~Dong, R.~Socher, L.-J. Li, K.~Li, and L.~Fei-Fei.
\newblock Imagenet: A large-scale hierarchical image database.
\newblock In {\em CVPR}, pages 248--255, 2009.

\bibitem{REF08a}
R.-E. Fan, K.-W. Chang, C.-J. Hsieh, X.-R. Wang, and C.-J. Lin.
\newblock {LIBLINEAR}: A library for large linear classification.
\newblock {\em JMLR}, 9:1871--1874, 2008.

\bibitem{Fukushima201377}
T.~Fukushima.
\newblock Precise and fast computation of {Lambert} {W}-functions without
  transcendental function evaluations.
\newblock {\em Journal of Computational and Applied Mathematics}, 244:77 -- 89,
  2013.

\bibitem{gehler2009feature}
P.~Gehler and S.~Nowozin.
\newblock On feature combination for multiclass object classification.
\newblock In {\em ICCV}, pages 221--228, 2009.

\bibitem{Gupta2014}
M.~R. Gupta, S.~Bengio, and J.~Weston.
\newblock Training highly multiclass classifiers.
\newblock {\em JMLR}, 15:1461--1492, 2014.

\bibitem{HirLem2001}
J.-B. Hiriart-Urruty and C.~Lemar{\'e}chal.
\newblock {\em Fundamentals of Convex Analysis}.
\newblock Springer, Berlin, 2001.

\bibitem{householder1970numerical}
A.~S. Householder.
\newblock {\em The Numerical Treatment of a Single Nonlinear Equation}.
\newblock McGraw-Hill, 1970.

\bibitem{hsu2002comparison}
C.-W. Hsu and C.-J. Lin.
\newblock A comparison of methods for multiclass support vector machines.
\newblock {\em Neural Networks}, 13(2):415--425, 2002.

\bibitem{jia2014caffe}
Y.~Jia, E.~Shelhamer, J.~Donahue, S.~Karayev, J.~Long, R.~Girshick,
  S.~Guadarrama, and T.~Darrell.
\newblock Caffe: Convolutional architecture for fast feature embedding.
\newblock {\em arXiv preprint arXiv:1408.5093}, 2014.

\bibitem{joachims2005support}
T.~Joachims.
\newblock A support vector method for multivariate performance measures.
\newblock In {\em ICML}, pages 377--384, 2005.

\bibitem{krizhevsky2012imagenet}
A.~Krizhevsky, I.~Sutskever, and G.~Hinton.
\newblock Imagenet classification with deep convolutional neural networks.
\newblock In {\em NIPS}, pages 1106--1114, 2012.

\bibitem{lang1995newsweeder}
K.~Lang.
\newblock Newsweeder: Learning to filter netnews.
\newblock In {\em ICML}, pages 331--339, 1995.

\bibitem{lapin2015topk}
M.~Lapin, M.~Hein, and B.~Schiele.
\newblock Top-k multiclass {SVM}.
\newblock In {\em NIPS}, 2015.

\bibitem{lapin2014scalable}
M.~Lapin, B.~Schiele, and M.~Hein.
\newblock Scalable multitask representation learning for scene classification.
\newblock In {\em CVPR}, 2014.

\bibitem{li2014top}
N.~Li, R.~Jin, and Z.-H. Zhou.
\newblock Top rank optimization in linear time.
\newblock In {\em NIPS}, pages 1502--1510, 2014.

\bibitem{liu2015transductive}
L.~Liu, T.~G. Dietterich, N.~Li, and Z.~Zhou.
\newblock Transductive optimization of top k precision.
\newblock {\em CoRR}, abs/1510.05976, 2015.

\bibitem{mairal2010sparse}
J.~Mairal.
\newblock {\em Sparse Coding for Machine Learning, Image Processing and
  Computer Vision}.
\newblock PhD thesis, Ecole Normale Superieure de Cachan, 2010.

\bibitem{mairal2010network}
J.~Mairal, R.~Jenatton, F.~R. Bach, and G.~R. Obozinski.
\newblock Network flow algorithms for structured sparsity.
\newblock In {\em NIPS}, pages 1558--1566, 2010.

\bibitem{nesterov2005smooth}
Y.~Nesterov.
\newblock Smooth minimization of non-smooth functions.
\newblock {\em Mathematical Programming}, 103(1):127--152, 2005.

\bibitem{nilsback2008automated}
M.-E. Nilsback and A.~Zisserman.
\newblock Automated flower classification over a large number of classes.
\newblock In {\em ICVGIP}, pages 722--729, 2008.

\bibitem{nocedal2006numerical}
J.~Nocedal and S.~J. Wright.
\newblock {\em Numerical Optimization}.
\newblock Springer Science+ Business Media, 2006.

\bibitem{petersen2008matrix}
K.~B. Petersen, M.~S. Pedersen, et~al.
\newblock The matrix cookbook.
\newblock {\em Technical University of Denmark}, 450:7--15, 2008.

\bibitem{quattoni2009recognizing}
A.~Quattoni and A.~Torralba.
\newblock Recognizing indoor scenes.
\newblock In {\em CVPR}, 2009.

\bibitem{ICML2012Rakotomamonjy_664}
A.~Rakotomamonjy.
\newblock Sparse support vector infinite push.
\newblock In {\em ICML}, pages 1335--1342. ACM, 2012.

\bibitem{razavian2014cnn}
A.~S. Razavian, H.~Azizpour, J.~Sullivan, and S.~Carlsson.
\newblock {CNN} features off-the-shelf: an astounding baseline for recognition.
\newblock In {\em CVPRW, DeepVision workshop}, 2014.

\bibitem{ReiWil2010}
M.~Reid and B.~Williamson.
\newblock Composite binary losses.
\newblock {\em JMLR}, 11:2387--2422, 2010.

\bibitem{rennie2001improving}
J.~D. Rennie.
\newblock Improving multi-class text classification with naive bayes.
\newblock Technical report, Massachusetts Institute of Technology, 2001.

\bibitem{rifkin2004defense}
R.~Rifkin and A.~Klautau.
\newblock In defense of one-vs-all classification.
\newblock {\em JMLR}, 5:101--141, 2004.

\bibitem{rocha2014multiclass}
A.~Rocha and S.~Klein~Goldenstein.
\newblock Multiclass from binary: Expanding one-versus-all, one-versus-one and
  ecoc-based approaches.
\newblock {\em Neural Networks and Learning Systems, IEEE Transactions on},
  25(2):289--302, 2014.

\bibitem{ross2013learning}
S.~Ross, J.~Zhou, Y.~Yue, D.~Dey, and D.~Bagnell.
\newblock Learning policies for contextual submodular prediction.
\newblock In {\em ICML}, pages 1364--1372, 2013.

\bibitem{rudin2009p}
C.~Rudin.
\newblock The p-norm push: A simple convex ranking algorithm that concentrates
  at the top of the list.
\newblock {\em JMLR}, 10:2233--2271, 2009.

\bibitem{ILSVRCarxiv14}
O.~Russakovsky, J.~Deng, H.~Su, J.~Krause, S.~Satheesh, S.~Ma, Z.~Huang,
  A.~Karpathy, A.~Khosla, M.~Bernstein, A.~C. Berg, and L.~Fei-Fei.
\newblock {ImageNet Large Scale Visual Recognition Challenge}, 2014.

\bibitem{shalev2014accelerated}
S.~Shalev-Shwartz and T.~Zhang.
\newblock Accelerated proximal stochastic dual coordinate ascent for
  regularized loss minimization.
\newblock {\em Mathematical Programming}, pages 1--41, 2014.

\bibitem{sharan2009material}
L.~Sharan, R.~Rosenholtz, and E.~Adelson.
\newblock Material perception: What can you see in a brief glance?
\newblock {\em Journal of Vision}, 9(8):784--784, 2009.

\bibitem{simonyan14c}
K.~Simonyan and A.~Zisserman.
\newblock Very deep convolutional networks for large-scale image recognition.
\newblock {\em CoRR}, abs/1409.1556, 2014.

\bibitem{swersky2012probabilistic}
K.~Swersky, B.~J. Frey, D.~Tarlow, R.~S. Zemel, and R.~P. Adams.
\newblock Probabilistic $n$-choose-$k$ models for classification and ranking.
\newblock In {\em NIPS}, pages 3050--3058, 2012.

\bibitem{TewBar2007}
A.~Tewari and P.~Bartlett.
\newblock On the consistency of multiclass classification methods.
\newblock {\em JMLR}, 8:1007--1025, 2007.

\bibitem{tsochantaridis2005large}
I.~Tsochantaridis, T.~Joachims, T.~Hofmann, and Y.~Altun.
\newblock Large margin methods for structured and interdependent output
  variables.
\newblock {\em JMLR}, pages 1453--1484, 2005.

\bibitem{usunier2009ranking}
N.~Usunier, D.~Buffoni, and P.~Gallinari.
\newblock Ranking with ordered weighted pairwise classification.
\newblock In {\em ICML}, pages 1057--1064, 2009.

\bibitem{Veberic20122622}
D.~Veberi{\v c}.
\newblock Lambert {W} function for applications in physics.
\newblock {\em Computer Physics Communications}, 183(12):2622--2628, 2012.

\bibitem{vedaldi15matconvnet}
A.~Vedaldi and K.~Lenc.
\newblock Matconvnet -- convolutional neural networks for matlab.
\newblock In {\em Proceeding of the {ACM} Int. Conf. on Multimedia}, 2015.

\bibitem{wah2011caltech}
C.~Wah, S.~Branson, P.~Welinder, P.~Perona, and S.~Belongie.
\newblock The {Caltech-UCSD} {Birds-200-2011} dataset.
\newblock Technical report, California Institute of Technology, 2011.

\bibitem{wang2015places}
L.~Wang, S.~Guo, W.~Huang, and Y.~Qiao.
\newblock Places205-vggnet models for scene recognition.
\newblock {\em CoRR}, abs/1508.01667, 2015.

\bibitem{Weston2011}
J.~Weston, S.~Bengio, and N.~Usunier.
\newblock Wsabie: scaling up to large vocabulary image annotation.
\newblock {\em IJCAI}, pages 2764--2770, 2011.

\bibitem{xiao2010sun}
J.~Xiao, J.~Hays, K.~A. Ehinger, A.~Oliva, and A.~Torralba.
\newblock {SUN} database: Large-scale scene recognition from abbey to zoo.
\newblock In {\em CVPR}, 2010.

\bibitem{zhang2014part}
N.~Zhang, J.~Donahue, R.~Girshick, and T.~Darrell.
\newblock Part-based rcnn for fine-grained detection.
\newblock In {\em ECCV}, 2014.

\bibitem{zhou2014learning}
B.~Zhou, A.~Lapedriza, J.~Xiao, A.~Torralba, and A.~Oliva.
\newblock Learning deep features for scene recognition using places database.
\newblock In {\em NIPS}, 2014.

\end{thebibliography}
}

\iflongversion
\begin{table*}[ht]\scriptsize\centering\setlength{\tabcolsep}{.4em}
\begin{tabular}{l|cccccccccc}
\multicolumn{11}{c}{\small\textbf{ALOI}} \\\toprule
Method & Top-$1$ & Top-$2$ & Top-$3$ & Top-$4$ & Top-$5$ & Top-$6$ & Top-$7$ & Top-$8$ & Top-$9$ & Top-$10$ \\
\midrule
\midrule
\SvmOva & $82.4$ & $87.4$ & $89.5$ & $90.7$ & $91.5$ & $92.1$ & $92.6$ & $93.1$ & $93.4$ & $93.7$ \\
\midrule
\LrOva & $86.1$ & $91.1$ & $93.0$ & $94.1$ & $94.8$ & $95.4$ & $95.8$ & $96.1$ & $96.4$ & $96.6$ \\
\midrule
\SvmTopKa{1} / \SvmMulti & $90.0$ & $93.4$ & $95.1$ & $96.0$ & $96.7$ & $97.1$ & $97.5$ & $97.7$ & $97.9$ & $98.1$ \\
\SvmTopKa{2} & $90.0$ & $94.0$ & $95.5$ & $96.4$ & $97.0$ & $97.4$ & $97.7$ & $97.9$ & $98.1$ & $98.3$ \\
\SvmTopKa{3} & $89.2$ & $94.2$ & $95.5$ & $96.7$ & $97.2$ & $97.6$ & $97.8$ & $98.1$ & $98.2$ & $98.4$ \\
\SvmTopKa{4} & $88.4$ & $94.3$ & $95.6$ & $96.8$ & $97.4$ & $97.8$ & $98.0$ & $98.2$ & $98.4$ & $98.5$ \\
\bottomrule
\end{tabular}
\caption{Comparison of different methods in top-$k$ accuracy (\%).}
\end{table*}
\clearpage
\begin{table*}[ht]\scriptsize\centering\setlength{\tabcolsep}{.4em}
\begin{tabular}{l|cccccccccc}
\multicolumn{11}{c}{\small\textbf{Letter}} \\\toprule
Method & Top-$1$ & Top-$2$ & Top-$3$ & Top-$4$ & Top-$5$ & Top-$6$ & Top-$7$ & Top-$8$ & Top-$9$ & Top-$10$ \\
\midrule
\midrule
\SvmOva & $63.0$ & $75.7$ & $82.0$ & $85.7$ & $88.1$ & $89.9$ & $91.4$ & $92.8$ & $93.9$ & $94.6$ \\
\midrule
\LrOva & $68.1$ & $81.1$ & $86.1$ & $88.4$ & $90.6$ & $92.2$ & $93.4$ & $94.6$ & $95.3$ & $96.2$ \\
\midrule
\SvmTopKa{1} / \SvmMulti & $76.5$ & $85.5$ & $89.2$ & $91.5$ & $93.1$ & $94.3$ & $95.5$ & $96.5$ & $97.0$ & $97.7$ \\
\SvmTopKa{2} & $76.1$ & $86.9$ & $90.1$ & $92.2$ & $93.3$ & $94.8$ & $96.0$ & $96.5$ & $97.2$ & $97.7$ \\
\SvmTopKa{3} & $74.0$ & $87.0$ & $91.0$ & $93.0$ & $94.4$ & $95.4$ & $96.2$ & $96.7$ & $97.3$ & $97.8$ \\
\SvmTopKa{4} & $71.4$ & $86.4$ & $91.4$ & $93.5$ & $94.8$ & $95.7$ & $96.6$ & $97.2$ & $97.6$ & $98.2$ \\
\bottomrule
\end{tabular}
\caption{Comparison of different methods in top-$k$ accuracy (\%).}
\end{table*}
\clearpage
\begin{table*}[ht]\scriptsize\centering\setlength{\tabcolsep}{.4em}
\begin{tabular}{l|cccccccccc}
\multicolumn{11}{c}{\small\textbf{News 20}} \\\toprule
Method & Top-$1$ & Top-$2$ & Top-$3$ & Top-$4$ & Top-$5$ & Top-$6$ & Top-$7$ & Top-$8$ & Top-$9$ & Top-$10$ \\
\midrule
\midrule
\SvmOva & $84.3$ & $93.0$ & $95.4$ & $97.0$ & $97.9$ & $98.5$ & $98.8$ & $99.0$ & $99.3$ & $99.5$ \\
\midrule
\LrOva & $84.9$ & $93.1$ & $96.3$ & $97.2$ & $97.8$ & $98.3$ & $98.7$ & $98.8$ & $99.0$ & $99.3$ \\
\midrule
\SvmTopKa{1} / \SvmMulti & $85.4$ & $92.7$ & $94.9$ & $96.4$ & $97.2$ & $97.6$ & $98.2$ & $98.5$ & $98.7$ & $99.1$ \\
\SvmTopKa{2} & $85.3$ & $94.3$ & $96.4$ & $97.2$ & $97.9$ & $98.4$ & $98.5$ & $98.9$ & $99.0$ & $99.1$ \\
\SvmTopKa{3} & $85.1$ & $94.8$ & $96.6$ & $97.7$ & $98.2$ & $98.6$ & $99.0$ & $99.2$ & $99.3$ & $99.3$ \\
\SvmTopKa{4} & $85.0$ & $94.5$ & $96.7$ & $97.7$ & $98.3$ & $98.6$ & $98.9$ & $99.1$ & $99.3$ & $99.4$ \\
\bottomrule
\end{tabular}
\caption{Comparison of different methods in top-$k$ accuracy (\%).}
\end{table*}
\clearpage
\begin{table*}[ht]\scriptsize\centering\setlength{\tabcolsep}{.4em}
\begin{tabular}{l|cccccccccc}
\multicolumn{11}{c}{\small\textbf{Caltech 101 Silhouettes}} \\\toprule
Method & Top-$1$ & Top-$2$ & Top-$3$ & Top-$4$ & Top-$5$ & Top-$6$ & Top-$7$ & Top-$8$ & Top-$9$ & Top-$10$ \\
\midrule
\midrule
\SvmOva & $61.8$ & $73.1$ & $76.5$ & $78.5$ & $80.8$ & $82.5$ & $83.6$ & $84.6$ & $85.6$ & $86.6$ \\
\midrule
\LrOva & $63.2$ & $77.1$ & $80.4$ & $82.6$ & $84.4$ & $85.7$ & $87.0$ & $87.9$ & $88.6$ & $89.4$ \\
\midrule
\SvmTopKa{1} / \SvmMulti & $62.8$ & $74.6$ & $77.8$ & $80.0$ & $82.0$ & $83.3$ & $84.4$ & $85.0$ & $85.9$ & $86.9$ \\
\SvmTopKa{2} & $63.1$ & $76.2$ & $79.0$ & $81.0$ & $82.7$ & $84.4$ & $85.5$ & $86.2$ & $87.0$ & $87.6$ \\
\SvmTopKa{3} & $63.4$ & $76.7$ & $79.7$ & $81.5$ & $83.6$ & $85.1$ & $85.8$ & $86.6$ & $87.6$ & $88.3$ \\
\SvmTopKa{4} & $63.2$ & $76.6$ & $79.8$ & $82.4$ & $84.0$ & $85.3$ & $86.0$ & $86.9$ & $87.8$ & $88.6$ \\
\bottomrule
\end{tabular}
\caption{Comparison of different methods in top-$k$ accuracy (\%).}
\end{table*}
\clearpage
\begin{table*}[ht]\scriptsize\centering\setlength{\tabcolsep}{.4em}
\begin{tabular}{l|cccccccccc}
\multicolumn{11}{c}{\small\textbf{Caltech 101}
(average of $39$ kernels from \cite{gehler2009feature}, $5$ splits)} \\\toprule
Method & Top-$1$ & Top-$2$ & Top-$3$ & Top-$4$ & Top-$5$ & Top-$6$ & Top-$7$ & Top-$8$ & Top-$9$ & Top-$10$ \\
\midrule
\midrule
\SvmTopKa{1} / \SvmMulti & $73.2 \pm 1.0$ & $81.6 \pm 0.8$ & $85.5 \pm 0.8$ & $87.8 \pm 0.7$ & $89.5 \pm 0.6$ & $90.8 \pm 0.5$ & $91.8 \pm 0.4$ & $92.7 \pm 0.4$ & $93.3 \pm 0.4$ & $93.8 \pm 0.4$ \\
\SvmTopKa{2} & $73.2 \pm 0.9$ & $81.7 \pm 0.7$ & $85.5 \pm 0.9$ & $87.8 \pm 0.7$ & $89.6 \pm 0.6$ & $90.9 \pm 0.5$ & $91.9 \pm 0.4$ & $92.7 \pm 0.3$ & $93.3 \pm 0.4$ & $93.9 \pm 0.5$ \\
\SvmTopKa{3} & $73.0 \pm 1.0$ & $81.6 \pm 0.7$ & $85.4 \pm 0.9$ & $87.8 \pm 0.6$ & $89.6 \pm 0.6$ & $90.9 \pm 0.4$ & $91.8 \pm 0.3$ & $92.7 \pm 0.4$ & $93.3 \pm 0.4$ & $93.8 \pm 0.5$ \\
\SvmTopKa{4} & $72.8 \pm 1.0$ & $81.4 \pm 0.7$ & $85.4 \pm 0.9$ & $87.9 \pm 0.6$ & $89.6 \pm 0.5$ & $90.9 \pm 0.4$ & $91.8 \pm 0.3$ & $92.7 \pm 0.4$ & $93.3 \pm 0.4$ & $93.8 \pm 0.4$ \\
\SvmTopKa{5} & $72.6 \pm 1.0$ & $81.2 \pm 0.7$ & $85.2 \pm 0.8$ & $87.7 \pm 0.6$ & $89.5 \pm 0.6$ & $90.8 \pm 0.5$ & $91.7 \pm 0.4$ & $92.6 \pm 0.4$ & $93.2 \pm 0.5$ & $93.7 \pm 0.5$ \\
\SvmTopKa{10} & $71.0 \pm 0.8$ & $80.2 \pm 1.0$ & $84.2 \pm 0.7$ & $87.0 \pm 0.8$ & $88.8 \pm 0.7$ & $90.1 \pm 0.5$ & $91.3 \pm 0.4$ & $92.2 \pm 0.5$ & $92.9 \pm 0.4$ & $93.5 \pm 0.4$ \\
\bottomrule
\end{tabular}
\caption{Comparison of different methods in top-$k$ accuracy (\%).}
\end{table*}
\clearpage
\begin{table*}[ht]\scriptsize\centering\setlength{\tabcolsep}{.4em}
\begin{tabular}{l|cccccccccc}
\multicolumn{11}{c}{\small\textbf{Caltech 256}
(average of $39$ kernels from \cite{gehler2009feature})} \\\toprule
Method & Top-$1$ & Top-$2$ & Top-$3$ & Top-$4$ & Top-$5$ & Top-$6$ & Top-$7$ & Top-$8$ & Top-$9$ & Top-$10$ \\
\midrule
\midrule
\SvmTopKa{1} / \SvmMulti & $47.1$ & $56.3$ & $61.4$ & $64.7$ & $67.3$ & $69.3$ & $71.5$ & $72.6$ & $74.1$ & $75.2$ \\
\SvmTopKa{2} & $47.1$ & $56.3$ & $61.4$ & $64.6$ & $67.2$ & $69.4$ & $71.3$ & $72.6$ & $74.1$ & $75.2$ \\
\SvmTopKa{3} & $47.0$ & $56.3$ & $61.5$ & $64.6$ & $67.3$ & $69.4$ & $71.3$ & $72.6$ & $74.1$ & $75.1$ \\
\SvmTopKa{4} & $46.9$ & $56.2$ & $61.4$ & $64.5$ & $67.3$ & $69.3$ & $71.3$ & $72.8$ & $74.1$ & $75.2$ \\
\SvmTopKa{5} & $46.8$ & $56.1$ & $61.2$ & $64.4$ & $67.1$ & $69.3$ & $71.2$ & $72.8$ & $74.2$ & $75.1$ \\
\SvmTopKa{10} & $45.4$ & $55.3$ & $60.7$ & $64.1$ & $66.7$ & $69.0$ & $71.0$ & $72.6$ & $73.9$ & $75.2$ \\
\bottomrule
\end{tabular}
\caption{Comparison of different methods in top-$k$ accuracy (\%).}
\end{table*}
\clearpage
\begin{table*}[ht]\scriptsize\centering\setlength{\tabcolsep}{.4em}
\begin{tabular}{l|cccccccccc}
\multicolumn{11}{c}{\small\textbf{CUB}} \\\toprule
Method & Top-$1$ & Top-$2$ & Top-$3$ & Top-$4$ & Top-$5$ & Top-$6$ & Top-$7$ & Top-$8$ & Top-$9$ & Top-$10$ \\
\midrule
\midrule
\SvmOva & $60.6$ & $71.4$ & $77.1$ & $80.7$ & $83.4$ & $85.2$ & $86.6$ & $87.9$ & $89.0$ & $89.9$ \\
\midrule
\LrOva & $62.3$ & $74.8$ & $80.5$ & $84.6$ & $87.4$ & $89.4$ & $90.7$ & $91.9$ & $92.8$ & $93.5$ \\
\midrule
\SvmTopKa{1} / \SvmMulti & $61.0$ & $73.3$ & $79.2$ & $82.8$ & $85.7$ & $87.8$ & $89.5$ & $90.7$ & $91.6$ & $92.3$ \\
\SvmTopKa{2} & $61.2$ & $73.7$ & $79.9$ & $83.5$ & $85.9$ & $88.2$ & $89.9$ & $91.0$ & $91.9$ & $92.6$ \\
\SvmTopKa{3} & $61.3$ & $74.9$ & $80.4$ & $83.9$ & $86.3$ & $88.1$ & $89.9$ & $91.3$ & $91.9$ & $92.5$ \\
\SvmTopKa{4} & $61.1$ & $75.1$ & $81.1$ & $84.6$ & $86.7$ & $88.7$ & $89.8$ & $90.9$ & $91.9$ & $92.8$ \\
\bottomrule
\end{tabular}
\caption{Comparison of different methods in top-$k$ accuracy (\%).}
\end{table*}
\clearpage
\begin{table*}[ht]\scriptsize\centering\setlength{\tabcolsep}{.4em}
\begin{tabular}{l|cccccccccc}
\multicolumn{11}{c}{\small\textbf{Flowers}} \\\toprule
Method & Top-$1$ & Top-$2$ & Top-$3$ & Top-$4$ & Top-$5$ & Top-$6$ & Top-$7$ & Top-$8$ & Top-$9$ & Top-$10$ \\
\midrule
\midrule
\SvmOva & $82.0$ & $89.2$ & $91.7$ & $93.2$ & $94.3$ & $95.2$ & $95.9$ & $95.8$ & $96.2$ & $96.8$ \\
\midrule
\LrOva & $82.6$ & $89.4$ & $92.2$ & $93.9$ & $94.8$ & $95.8$ & $96.4$ & $96.9$ & $97.3$ & $97.6$ \\
\midrule
\SvmTopKa{1} / \SvmMulti & $82.5$ & $89.5$ & $92.2$ & $93.8$ & $94.8$ & $95.6$ & $96.2$ & $95.5$ & $96.0$ & $96.4$ \\
\SvmTopKa{2} & $82.3$ & $89.5$ & $92.3$ & $93.9$ & $95.0$ & $95.7$ & $96.5$ & $95.6$ & $95.9$ & $96.2$ \\
\SvmTopKa{3} & $81.9$ & $89.3$ & $92.2$ & $93.8$ & $95.0$ & $95.8$ & $96.4$ & $97.0$ & $96.0$ & $96.1$ \\
\SvmTopKa{4} & $81.8$ & $89.3$ & $92.3$ & $94.0$ & $95.0$ & $95.9$ & $96.6$ & $97.0$ & $95.6$ & $97.8$ \\
\bottomrule
\end{tabular}
\caption{Comparison of different methods in top-$k$ accuracy (\%).}
\end{table*}
\clearpage
\begin{table*}[ht]\scriptsize\centering\setlength{\tabcolsep}{.4em}
\begin{tabular}{l|cccccccccc}
\multicolumn{11}{c}{\small\textbf{FMD}} \\\toprule
Method & Top-$1$ & Top-$2$ & Top-$3$ & Top-$4$ & Top-$5$ & Top-$6$ & Top-$7$ & Top-$8$ & Top-$9$ & Top-$10$ \\
\midrule
\midrule
\SvmOva & $77.4$ & $87.2$ & $92.4$ & $94.4$ & $96.4$ & $97.8$ & $99.0$ & $99.8$ & $100.0$ & $100.0$ \\
\midrule
\LrOva & $79.6$ & $90.2$ & $94.2$ & $96.6$ & $98.2$ & $98.8$ & $99.2$ & $100.0$ & $99.8$ & $100.0$ \\
\midrule
\SvmTopKa{1} / \SvmMulti & $77.6$ & $88.8$ & $93.8$ & $95.4$ & $97.2$ & $98.4$ & $100.0$ & $100.0$ & $100.0$ & $100.0$ \\
\SvmTopKa{2} & $78.2$ & $89.2$ & $94.0$ & $95.8$ & $97.4$ & $98.6$ & $99.8$ & $100.0$ & $100.0$ & $100.0$ \\
\SvmTopKa{3} & $78.8$ & $89.2$ & $94.6$ & $96.4$ & $97.8$ & $98.8$ & $99.4$ & $99.8$ & $100.0$ & $100.0$ \\
\SvmTopKa{4} & $78.2$ & $89.4$ & $94.6$ & $96.8$ & $98.0$ & $98.8$ & $99.4$ & $99.6$ & $100.0$ & $100.0$ \\
\bottomrule
\end{tabular}
\caption{Comparison of different methods in top-$k$ accuracy (\%).}
\end{table*}
\clearpage
\begin{table*}[ht]\scriptsize\centering\setlength{\tabcolsep}{.4em}
\begin{tabular}{l|cccccccccc}
\multicolumn{11}{c}{\small\textbf{Indoor 67}
(AlexNet trained on Places 205, FC7 output,
provided by \cite{lapin2015topk})} \\\toprule
Method & Top-$1$ & Top-$2$ & Top-$3$ & Top-$4$ & Top-$5$ & Top-$6$ & Top-$7$ & Top-$8$ & Top-$9$ & Top-$10$ \\
\midrule
\midrule
\SvmOva & $71.7$ & $80.1$ & $85.9$ & $90.2$ & $92.6$ & $94.3$ & $94.9$ & $95.7$ & $96.4$ & $96.9$ \\
\midrule
\LrOva & $73.1$ & $84.5$ & $89.6$ & $91.8$ & $93.3$ & $94.3$ & $95.3$ & $96.1$ & $96.6$ & $97.0$ \\
\midrule
\SvmTopKa{1} / \SvmMulti & $74.0$ & $85.2$ & $89.3$ & $91.9$ & $93.4$ & $94.9$ & $95.6$ & $95.8$ & $96.4$ & $96.9$ \\
\SvmTopKa{2} & $73.1$ & $85.7$ & $90.4$ & $92.2$ & $94.5$ & $95.1$ & $96.2$ & $96.6$ & $97.0$ & $97.3$ \\
\SvmTopKa{3} & $71.6$ & $86.3$ & $91.1$ & $93.2$ & $94.7$ & $95.7$ & $96.4$ & $96.6$ & $97.1$ & $97.2$ \\
\SvmTopKa{4} & $71.4$ & $85.7$ & $90.7$ & $93.3$ & $94.8$ & $95.7$ & $96.2$ & $96.6$ & $97.2$ & $97.8$ \\
\bottomrule
\end{tabular}
\caption{Comparison of different methods in top-$k$ accuracy (\%).}
\end{table*}
\clearpage
\begin{table*}[ht]\scriptsize\centering\setlength{\tabcolsep}{.4em}
\begin{tabular}{l|cccccccccc}
\multicolumn{11}{c}{\small\textbf{Indoor 67}
(VGG-16 trained on Places 205, FC6 output)} \\\toprule
Method & Top-$1$ & Top-$2$ & Top-$3$ & Top-$4$ & Top-$5$ & Top-$6$ & Top-$7$ & Top-$8$ & Top-$9$ & Top-$10$ \\
\midrule
\midrule
\SvmOva & $80.4$ & $89.9$ & $94.3$ & $95.3$ & $96.0$ & $96.9$ & $97.3$ & $97.7$ & $97.9$ & $98.3$ \\
\midrule
\LrOva & $82.5$ & $92.6$ & $95.1$ & $96.3$ & $97.5$ & $97.5$ & $98.3$ & $98.6$ & $98.4$ & $98.7$ \\
\midrule
\SvmTopKa{1} / \SvmMulti & $82.6$ & $91.4$ & $95.0$ & $96.1$ & $96.0$ & $97.9$ & $97.5$ & $97.8$ & $98.3$ & $98.9$ \\
\SvmTopKa{2} & $83.1$ & $92.3$ & $95.4$ & $96.8$ & $97.7$ & $98.0$ & $98.3$ & $98.7$ & $98.7$ & $98.5$ \\
\SvmTopKa{3} & $83.0$ & $92.5$ & $95.4$ & $97.5$ & $97.8$ & $98.3$ & $98.5$ & $98.7$ & $98.7$ & $98.9$ \\
\SvmTopKa{4} & $82.9$ & $92.5$ & $95.6$ & $97.0$ & $97.8$ & $98.2$ & $98.7$ & $98.7$ & $98.8$ & $99.0$ \\
\bottomrule
\end{tabular}
\caption{Comparison of different methods in top-$k$ accuracy (\%).}
\end{table*}
\clearpage
\begin{table*}[ht]\scriptsize\centering\setlength{\tabcolsep}{.4em}
\begin{tabular}{l|cccccccccc}
\multicolumn{11}{c}{\small\textbf{Indoor 67}
(VGG-16 trained on Places 205, FC7 output)} \\\toprule
Method & Top-$1$ & Top-$2$ & Top-$3$ & Top-$4$ & Top-$5$ & Top-$6$ & Top-$7$ & Top-$8$ & Top-$9$ & Top-$10$ \\
\midrule
\midrule
\SvmOva & $81.9$ & $91.0$ & $94.3$ & $95.7$ & $96.5$ & $97.2$ & $97.4$ & $97.7$ & $97.8$ & $98.0$ \\
\midrule
\LrOva & $82.0$ & $91.6$ & $94.9$ & $96.3$ & $97.2$ & $97.8$ & $98.1$ & $98.3$ & $98.6$ & $98.7$ \\
\midrule
\SvmTopKa{1} / \SvmMulti & $82.5$ & $91.7$ & $95.4$ & $96.9$ & $97.3$ & $97.8$ & $97.6$ & $98.5$ & $98.8$ & $99.1$ \\
\SvmTopKa{2} & $82.0$ & $91.3$ & $95.1$ & $96.6$ & $97.6$ & $97.8$ & $98.0$ & $98.6$ & $98.7$ & $99.0$ \\
\SvmTopKa{3} & $81.6$ & $91.4$ & $95.1$ & $96.8$ & $97.7$ & $98.2$ & $98.6$ & $98.7$ & $98.8$ & $99.0$ \\
\SvmTopKa{4} & $80.8$ & $91.3$ & $95.2$ & $96.9$ & $97.7$ & $98.6$ & $98.6$ & $98.7$ & $98.9$ & $99.1$ \\
\bottomrule
\end{tabular}
\caption{Comparison of different methods in top-$k$ accuracy (\%).}
\end{table*}
\clearpage
\begin{table*}[ht]\scriptsize\centering\setlength{\tabcolsep}{.4em}
\begin{tabular}{l|cccccccccc}
\multicolumn{11}{c}{\small\textbf{SUN 397}
(Fisher Vector kernels provided by \cite{lapin2014scalable})} \\\toprule
Method & Top-$1$ & Top-$2$ & Top-$3$ & Top-$4$ & Top-$5$ & Top-$6$ & Top-$7$ & Top-$8$ & Top-$9$ & Top-$10$ \\
\midrule
\midrule
\SvmTopKa{1} / \SvmMulti & $48.9 \pm 0.3$ & $60.6 \pm 0.3$ & $66.8 \pm 0.2$ & $70.8 \pm 0.3$ & $73.8 \pm 0.2$ & $76.1 \pm 0.2$ & $78.0 \pm 0.2$ & $79.5 \pm 0.2$ & $80.9 \pm 0.2$ & $82.0 \pm 0.2$ \\
\SvmTopKa{2} & $48.9 \pm 0.3$ & $60.5 \pm 0.3$ & $66.8 \pm 0.2$ & $70.8 \pm 0.3$ & $73.8 \pm 0.2$ & $76.1 \pm 0.2$ & $78.0 \pm 0.2$ & $79.5 \pm 0.2$ & $80.9 \pm 0.2$ & $82.0 \pm 0.2$ \\
\SvmTopKa{3} & $48.9 \pm 0.3$ & $60.5 \pm 0.3$ & $66.8 \pm 0.2$ & $70.8 \pm 0.3$ & $73.8 \pm 0.2$ & $76.1 \pm 0.2$ & $78.0 \pm 0.3$ & $79.5 \pm 0.2$ & $80.9 \pm 0.2$ & $82.0 \pm 0.2$ \\
\SvmTopKa{4} & $48.8 \pm 0.3$ & $60.5 \pm 0.3$ & $66.8 \pm 0.2$ & $70.8 \pm 0.3$ & $73.8 \pm 0.2$ & $76.1 \pm 0.2$ & $78.0 \pm 0.2$ & $79.6 \pm 0.2$ & $80.9 \pm 0.2$ & $82.0 \pm 0.2$ \\
\SvmTopKa{5} & $48.7 \pm 0.3$ & $60.5 \pm 0.3$ & $66.8 \pm 0.2$ & $70.9 \pm 0.3$ & $73.9 \pm 0.3$ & $76.2 \pm 0.2$ & $78.1 \pm 0.2$ & $79.6 \pm 0.2$ & $80.9 \pm 0.2$ & $82.1 \pm 0.2$ \\
\SvmTopKa{10} & $48.3 \pm 0.4$ & $60.4 \pm 0.3$ & $66.9 \pm 0.2$ & $71.1 \pm 0.3$ & $74.1 \pm 0.2$ & $76.5 \pm 0.2$ & $78.4 \pm 0.2$ & $80.0 \pm 0.2$ & $81.3 \pm 0.2$ & $82.5 \pm 0.2$ \\
\bottomrule
\end{tabular}
\caption{Comparison of different methods in top-$k$ accuracy (\%).}
\end{table*}
\clearpage
\begin{table*}[ht]\scriptsize\centering\setlength{\tabcolsep}{.4em}
\begin{tabular}{l|cccccccccc}
\multicolumn{11}{c}{\small\textbf{SUN 397}
(AlexNet trained on Places 205, FC7 output,
provided by \cite{lapin2015topk})} \\\toprule
Method & Top-$1$ & Top-$2$ & Top-$3$ & Top-$4$ & Top-$5$ & Top-$6$ & Top-$7$ & Top-$8$ & Top-$9$ & Top-$10$ \\
\midrule
\midrule
\SvmOva & $44.1 \pm 0.4$ & $60.8 \pm 0.3$ & $69.9 \pm 0.2$ & $75.8 \pm 0.1$ & $79.8 \pm 0.1$ & $82.8 \pm 0.1$ & $85.1 \pm 0.2$ & $86.9 \pm 0.2$ & $88.3 \pm 0.2$ & $89.6 \pm 0.2$ \\
\midrule
\LrOva & $53.9 \pm 0.2$ & $69.2 \pm 0.2$ & $76.5 \pm 0.2$ & $80.9 \pm 0.2$ & $84.0 \pm 0.3$ & $86.2 \pm 0.2$ & $87.9 \pm 0.2$ & $89.2 \pm 0.2$ & $90.3 \pm 0.2$ & $91.2 \pm 0.2$ \\
\midrule
\SvmTopKa{1} / \SvmMulti & $58.2 \pm 0.2$ & $71.7 \pm 0.2$ & $78.2 \pm 0.1$ & $82.3 \pm 0.2$ & $85.0 \pm 0.2$ & $87.1 \pm 0.2$ & $88.8 \pm 0.2$ & $90.0 \pm 0.2$ & $91.0 \pm 0.2$ & $91.9 \pm 0.2$ \\
\SvmTopKa{2} & $58.8 \pm 0.2$ & $72.7 \pm 0.2$ & $79.3 \pm 0.2$ & $83.3 \pm 0.2$ & $85.9 \pm 0.2$ & $87.8 \pm 0.2$ & $89.2 \pm 0.2$ & $90.3 \pm 0.2$ & $91.3 \pm 0.2$ & $92.2 \pm 0.2$ \\
\SvmTopKa{3} & $59.0 \pm 0.1$ & $73.2 \pm 0.2$ & $79.9 \pm 0.2$ & $83.8 \pm 0.2$ & $86.5 \pm 0.2$ & $88.3 \pm 0.2$ & $89.7 \pm 0.2$ & $90.9 \pm 0.2$ & $91.8 \pm 0.2$ & $92.6 \pm 0.2$ \\
\SvmTopKa{4} & $58.9 \pm 0.1$ & $73.5 \pm 0.2$ & $80.3 \pm 0.2$ & $84.2 \pm 0.2$ & $86.8 \pm 0.2$ & $88.6 \pm 0.2$ & $90.0 \pm 0.2$ & $91.1 \pm 0.2$ & $92.0 \pm 0.2$ & $92.8 \pm 0.2$ \\
\bottomrule
\end{tabular}
\caption{Comparison of different methods in top-$k$ accuracy (\%).}
\end{table*}
\clearpage
\begin{table*}[ht]\scriptsize\centering\setlength{\tabcolsep}{.4em}
\begin{tabular}{l|cccccccccc}
\multicolumn{11}{c}{\small\textbf{SUN 397}
(VGG-16 trained on Places 205, FC7 output)} \\\toprule
Method & Top-$1$ & Top-$2$ & Top-$3$ & Top-$4$ & Top-$5$ & Top-$6$ & Top-$7$ & Top-$8$ & Top-$9$ & Top-$10$ \\
\midrule
\midrule
\SvmOva & $65.4 \pm 0.2$ & $77.6 \pm 0.1$ & $83.8 \pm 0.2$ & $87.2 \pm 0.1$ & $89.6 \pm 0.1$ & $91.3 \pm 0.1$ & $92.6 \pm 0.2$ & $93.6 \pm 0.2$ & $94.3 \pm 0.2$ & $94.9 \pm 0.1$ \\
\midrule
\LrOva & $67.6 \pm 0.1$ & $81.5 \pm 0.2$ & $87.2 \pm 0.2$ & $90.4 \pm 0.2$ & $92.4 \pm 0.1$ & $93.7 \pm 0.1$ & $94.7 \pm 0.1$ & $95.4 \pm 0.1$ & $96.0 \pm 0.1$ & $96.4 \pm 0.1$ \\
\midrule
\SvmTopKa{1} / \SvmMulti & $65.8 \pm 0.1$ & $79.0 \pm 0.2$ & $85.1 \pm 0.2$ & $88.4 \pm 0.2$ & $90.8 \pm 0.1$ & $92.3 \pm 0.1$ & $93.3 \pm 0.1$ & $94.2 \pm 0.1$ & $94.8 \pm 0.1$ & $95.3 \pm 0.1$ \\
\SvmTopKa{2} & $66.4 \pm 0.2$ & $80.2 \pm 0.2$ & $86.1 \pm 0.1$ & $89.4 \pm 0.1$ & $91.5 \pm 0.1$ & $92.9 \pm 0.2$ & $93.8 \pm 0.2$ & $94.6 \pm 0.2$ & $95.3 \pm 0.1$ & $95.7 \pm 0.1$ \\
\SvmTopKa{3} & $66.5 \pm 0.2$ & $80.6 \pm 0.2$ & $86.5 \pm 0.1$ & $89.7 \pm 0.2$ & $91.8 \pm 0.1$ & $93.2 \pm 0.1$ & $94.2 \pm 0.1$ & $95.0 \pm 0.1$ & $95.3 \pm 0.1$ & $95.9 \pm 0.1$ \\
\SvmTopKa{4} & $66.4 \pm 0.2$ & $80.8 \pm 0.2$ & $86.8 \pm 0.1$ & $90.0 \pm 0.2$ & $92.1 \pm 0.2$ & $93.4 \pm 0.1$ & $94.4 \pm 0.1$ & $95.1 \pm 0.1$ & $95.7 \pm 0.1$ & $96.2 \pm 0.1$ \\
\bottomrule
\end{tabular}
\caption{Comparison of different methods in top-$k$ accuracy (\%).}
\end{table*}
\clearpage
\begin{table*}[ht]\scriptsize\centering\setlength{\tabcolsep}{.4em}
\begin{tabular}{l|cccccccccc}
\multicolumn{11}{c}{\small\textbf{Places 205}
(AlexNet trained on Places 205, FC7 output,
provided by \cite{lapin2015topk})} \\\toprule
Method & Top-$1$ & Top-$2$ & Top-$3$ & Top-$4$ & Top-$5$ & Top-$6$ & Top-$7$ & Top-$8$ & Top-$9$ & Top-$10$ \\
\midrule
\midrule
\SvmTopKa{1} / \SvmMulti & $50.6$ & $64.5$ & $71.4$ & $75.5$ & $78.5$ & $80.7$ & $82.5$ & $84.0$ & $85.1$ & $86.2$ \\
\SvmTopKa{2} & $51.1$ & $65.7$ & $73.1$ & $77.5$ & $80.7$ & $83.1$ & $84.9$ & $86.3$ & $87.5$ & $88.4$ \\
\SvmTopKa{3} & $51.3$ & $66.2$ & $73.2$ & $77.9$ & $81.3$ & $83.6$ & $85.6$ & $87.1$ & $88.3$ & $89.4$ \\
\SvmTopKa{4} & $51.2$ & $66.3$ & $73.5$ & $78.1$ & $81.4$ & $83.7$ & $85.7$ & $87.3$ & $88.7$ & $89.7$ \\
\SvmTopKa{5} & $50.8$ & $66.2$ & $73.7$ & $78.2$ & $81.4$ & $83.9$ & $85.8$ & $87.5$ & $88.9$ & $90.0$ \\
\SvmTopKa{10} & $50.1$ & $65.8$ & $73.4$ & $78.3$ & $81.6$ & $84.0$ & $86.0$ & $87.6$ & $89.0$ & $90.1$ \\
\bottomrule
\end{tabular}
\caption{Comparison of different methods in top-$k$ accuracy (\%).}
\end{table*}
\clearpage
\begin{table*}[ht]\scriptsize\centering\setlength{\tabcolsep}{.4em}
\begin{tabular}{l|cccccccccc}
\multicolumn{11}{c}{\small\textbf{Places 205}
(VGG-16 trained on Places 205, FC7 output)} \\\toprule
Method & Top-$1$ & Top-$2$ & Top-$3$ & Top-$4$ & Top-$5$ & Top-$6$ & Top-$7$ & Top-$8$ & Top-$9$ & Top-$10$ \\
\midrule
\midrule
\SvmTopKa{1} / \SvmMulti & $58.4$ & $72.5$ & $78.7$ & $82.3$ & $84.7$ & $86.4$ & $87.5$ & $88.4$ & $89.2$ & $89.9$ \\
\SvmTopKa{2} & $58.6$ & $73.4$ & $80.1$ & $84.1$ & $86.6$ & $88.5$ & $89.9$ & $90.8$ & $91.6$ & $92.2$ \\
\SvmTopKa{3} & $58.6$ & $73.7$ & $80.3$ & $84.5$ & $87.3$ & $89.3$ & $90.8$ & $91.8$ & $92.6$ & $93.3$ \\
\SvmTopKa{4} & $58.6$ & $73.8$ & $80.5$ & $84.6$ & $87.4$ & $89.5$ & $91.0$ & $92.1$ & $93.0$ & $93.8$ \\
\SvmTopKa{5} & $58.4$ & $73.8$ & $80.5$ & $84.5$ & $87.4$ & $89.5$ & $91.1$ & $92.3$ & $93.2$ & $94.0$ \\
\SvmTopKa{10} & $58.0$ & $73.2$ & $80.4$ & $84.6$ & $87.4$ & $89.6$ & $91.2$ & $92.5$ & $93.5$ & $94.3$ \\
\bottomrule
\end{tabular}
\caption{Comparison of different methods in top-$k$ accuracy (\%).}
\end{table*}
\clearpage
\begin{table*}[ht]\scriptsize\centering\setlength{\tabcolsep}{.4em}
\begin{tabular}{l|cccccccccc}
\multicolumn{11}{c}{\small\textbf{ILSVRC 2012}
(AlexNet trained on ImageNet, FC7 output,
provided by \cite{lapin2015topk})} \\\toprule
Method & Top-$1$ & Top-$2$ & Top-$3$ & Top-$4$ & Top-$5$ & Top-$6$ & Top-$7$ & Top-$8$ & Top-$9$ & Top-$10$ \\
\midrule
\midrule
\SvmTopKa{1} / \SvmMulti & $56.6$ & $67.3$ & $72.4$ & $75.4$ & $77.7$ & $79.3$ & $80.8$ & $82.0$ & $82.9$ & $83.7$ \\
\SvmTopKa{2} & $56.6$ & $68.1$ & $73.2$ & $76.4$ & $78.6$ & $80.3$ & $81.7$ & $82.8$ & $83.7$ & $84.6$ \\
\SvmTopKa{3} & $56.6$ & $68.3$ & $73.6$ & $76.8$ & $79.0$ & $80.7$ & $82.1$ & $83.2$ & $84.2$ & $85.0$ \\
\SvmTopKa{4} & $56.5$ & $68.4$ & $73.8$ & $77.1$ & $79.3$ & $81.1$ & $82.4$ & $83.5$ & $84.5$ & $85.3$ \\
\SvmTopKa{5} & $56.5$ & $68.4$ & $73.8$ & $77.2$ & $79.4$ & $81.1$ & $82.5$ & $83.7$ & $84.6$ & $85.4$ \\
\SvmTopKa{10} & $55.9$ & $68.2$ & $73.8$ & $77.3$ & $79.8$ & $81.4$ & $82.8$ & $84.0$ & $85.0$ & $85.8$ \\
\bottomrule
\end{tabular}
\caption{Comparison of different methods in top-$k$ accuracy (\%).}
\end{table*}
\clearpage
\begin{table*}[ht]\scriptsize\centering\setlength{\tabcolsep}{.4em}
\begin{tabular}{l|cccccccccc}
\multicolumn{11}{c}{\small\textbf{ILSVRC 2012}
(VGG-16 trained on ImageNet, FC7 output)} \\\toprule
Method & Top-$1$ & Top-$2$ & Top-$3$ & Top-$4$ & Top-$5$ & Top-$6$ & Top-$7$ & Top-$8$ & Top-$9$ & Top-$10$ \\
\midrule
\midrule
\SvmTopKa{1} / \SvmMulti & $68.3$ & $78.6$ & $82.9$ & $85.4$ & $87.0$ & $88.2$ & $89.2$ & $89.9$ & $90.6$ & $91.1$ \\
\SvmTopKa{2} & $68.3$ & $79.3$ & $83.7$ & $86.3$ & $87.8$ & $89.0$ & $89.9$ & $90.6$ & $91.3$ & $91.8$ \\
\SvmTopKa{3} & $68.2$ & $79.5$ & $84.0$ & $86.5$ & $88.1$ & $89.3$ & $90.2$ & $91.0$ & $91.6$ & $92.1$ \\
\SvmTopKa{4} & $68.0$ & $79.6$ & $84.1$ & $86.6$ & $88.3$ & $89.5$ & $90.4$ & $91.2$ & $91.8$ & $92.3$ \\
\SvmTopKa{5} & $67.8$ & $79.5$ & $84.1$ & $86.6$ & $88.2$ & $89.5$ & $90.5$ & $91.2$ & $91.9$ & $92.4$ \\
\SvmTopKa{10} & $67.0$ & $79.0$ & $83.8$ & $86.5$ & $88.3$ & $89.6$ & $90.6$ & $91.4$ & $92.1$ & $92.6$ \\
\bottomrule
\end{tabular}
\caption{Comparison of different methods in top-$k$ accuracy (\%).}
\end{table*}
\clearpage

\fi

\end{document}